\documentclass[3p]{elsarticle}

\usepackage{amssymb}
\usepackage{amsthm}

\usepackage{natbib}

\biboptions{sort&compress}

\newtheorem{definition}{Definition}
\newtheorem{theorem}{Theorem}
\newtheorem{proposition}[theorem]{Proposition}

\newtheorem{corollary}[theorem]{Corollary}
\newtheorem{example}{Example}

\usepackage[utf8]{inputenc}
\inputencoding{utf8} 

\usepackage{hyperref}
\usepackage{color}
\usepackage{mathtools}
\usepackage{enumerate}
\usepackage{enumitem,multicol}

\usepackage{nicefrac}

\usepackage{tikz}
\usetikzlibrary{calc}
\usetikzlibrary{decorations.pathmorphing}
\usetikzlibrary{decorations.markings}
\usetikzlibrary{fit}
\usetikzlibrary{backgrounds}
\usetikzlibrary{trees}
\usetikzlibrary{matrix}
\usetikzlibrary{shapes,arrows}
\usetikzlibrary{positioning}

\newcommand{\simplex}[1]{\Sigma_{\values{#1}}}

\newcommand{\newmid}{\hspace{-1pt}\mid\hspace{-1pt}}
\newcommand{\credal}{\M}
\newcommand{\powerG}{\mathcal{P}(\values{G})}
\newcommand{\eventsG}{\mathcal{P}_\emptyset(\values{G})}
\newcommand*{\M}{\mathcal{M}}
\newcommand{\reals}{\mathbb{R}}
\newcommand{\nodes}{G}
\newcommand{\parents}[1]{\mathit{\Pi}(#1)}
\newcommand{\children}[2][]{C_{#1}(#2)}
\newcommand{\descendants}[1]{D(#1)}
\newcommand{\precedes}{\sqsubseteq}
\newcommand{\sprecedes}{\sqsubset}
\newcommand{\nonparnondes}[1]{N(#1)}
\newcommand{\nondes}[1]{\mathit{\Pi}{N}(#1)}
\newcommand{\var}[1]{X_{#1}}
\newcommand{\values}[1]{\mathcal{X}_{#1}}
\newcommand{\gambles}[1]{\mathcal{G}(\values{#1})}
\newcommand{\xval}[1]{x_{#1}}
\newcommand{\zval}[1]{z_{#1}}
\newcommand{\ind}[1]{\mathbb{I}_{#1}}
\newcommand{\HIR}[3]{\Hirrelevant(#1,#2\mid#3)}
\newcommand{\Hirrelevant}{\mathrm{HIR}}
\newcommand{\AD}[3]{\asymmetricalD(#1,#2\newmid#3)}
\newcommand{\asymmetricalD}{\mathrm{AD}}
\newcommand{\irr}{\mathrm{irr}}

\newcommand{\mass}{p}
\newcommand{\element}{s}

\newcommand{\varel}{\var{\element}}
\newcommand{\valuesel}{\values{\element}}
\newcommand{\heads}{h}
\newcommand{\headsel}{\heads_\element}
\newcommand{\headslowel}{\underline{p}(\headsel)}
\newcommand{\headsupel}{\overline{p}(\headsel)}
\newcommand{\headsone}{\heads_{1}}
\newcommand{\headslowone}{\underline{p}(\headsone)}
\newcommand{\headsupone}{\overline{p}(\headsone)}
\newcommand{\headstwo}{\heads_{2}}
\newcommand{\headslowtwo}{\underline{p}(\headstwo)}
\newcommand{\headsuptwo}{\overline{p}(\headstwo)}
\newcommand{\tails}{t}
\newcommand{\tailsel}{\tails_\element}
\newcommand{\tailslowel}{\underline{p}(\tailsel)}
\newcommand{\tailsupel}{\overline{p}(\tailsel)}
\newcommand{\tailsone}{\tails_1}
\newcommand{\tailslowone}{\underline{p}(\tailsone)}
\newcommand{\tailsupone}{\overline{p}(\tailsone)}
\newcommand{\tailstwo}{\tails_2}
\newcommand{\tailslowtwo}{\underline{p}(\tailstwo)}
\newcommand{\tailsuptwo}{\overline{p}(\tailstwo)}

\journal{IJAR}

\begin{document}

\begin{frontmatter}



\title{Credal Networks under Epistemic Irrelevance}


\author{Jasper De Bock
}
\ead{jasper.debock@ugent.be}
\ead[url]{http://users.ugent.be/~jdbock}


\address{Ghent University, Department of Electronics and Information Systems\\ Technologiepark - Zwijnaarde 914, 9052 Zwijnaarde, Belgium}

\begin{abstract}
A credal network under epistemic irrelevance is a generalised type of Bayesian network that relaxes its two main building blocks. On the one hand, the local probabilities are allowed to be partially specified. On the other hand, the assessments of independence do not have to hold exactly. 
Conceptually, these two features turn credal networks under epistemic irrelevance into a powerful alternative to Bayesian networks, offering a more flexible approach to graph-based multivariate uncertainty modelling. However, in practice, they have long been perceived as very hard to work with, both theoretically and computationally.

The aim of this paper is to demonstrate that this perception is no longer justified. We provide a general introduction to credal networks under epistemic irrelevance, give an overview of the state of the art, and present several new theoretical results. 
Most importantly, we explain how these results can be combined to allow for the design of recursive inference methods. We provide numerous concrete examples of how this can be achieved, and use these to demonstrate that computing with credal networks under epistemic irrelevance is most definitely feasible, and in some cases even highly efficient.
We also discuss several philosophical aspects, including the lack of symmetry, how to deal with probability zero, the interpretation of lower expectations, the axiomatic status of graphoid properties, and the difference between updating and conditioning. 
\end{abstract}


\end{frontmatter}






\section{Introduction}\label{sec:introduction}

Bayesian networks \citep{Pearl:1988wk} 
owe their succes to the main feature that all probabilistic graphical models have in common: they are able to model the uncertainty that is associated with large multivariate problems in a manageable way, by combining local uncertainty models with intuitive graph-based independence assumptions. For a Bayesian network, the independence assumptions are derived from a directed acyclic graph and the local uncertainty models are probability distributions.

Credal networks~\citep{Cozman:2000ug,Cozman:2005ct,Antonucci:2014ty} generalize this concept by replacing the local probability distributions with closed convex sets of probability distributions, also called credal sets. In this way, they do not require the exact specification of all the local probabilities, but allow the user to provide partial constraints on them, such as intervals or inequalities. Depending on the type of credal network that is being considered, the independence assumptions that are derived from the graph are also generalised, by replacing them with weaker types of independence assessments.

This paper focusses on credal networks that adopt epistemic irrelevance as their notion of independence, called credal networks under epistemic irrelevance. This type of credal network does not impose independence in the usual sense, but instead replaces it with assessments of epistemic irrelevance. From a probabilistic point of view, these assessments of epistemic irrelevance can be regarded as relaxed assumptions of independence that allow for mild forms of correlation.

To the best of our knowledgde, the concept of a credal network under epistemic irrelevance was first introduced by Cozman~\cite{Cozman:1998:IIR:2074094.2074105}, who then referred to it as a locally defined Quasi-Bayesian network. Now, almost twenty years later, it is firmly established as one of the two main types of credal networks. However, compared with the other main type, which adopts strong independence as its notion of independence, credal networks under epistemic irrelevance have received considerably less attention. In particular, only very few algorithms for credal networks under epistemic irrelevance have so far been developed.

The reason for this shortage of algorithms can be explained as follows. In a Bayesian network, as a result of the independence assumptions that are imposed, the global probabilities of the network are a product of the local ones. This factorisation formula is very well known, and lies at the heart of all Bayesian network algorithms. Since credal networks under strong independence are essentially just sets of Bayesian networks, they inherit this factorisation formula. For this reason, algorithms for credal networks under strong independence can developed by starting from an existing algorithm for Bayesian networks and then adapting it to the context of sets of probabilities by adding an extra optimisation step; see Reference~\cite{Antonucci:2014ty} for a recent overview of related literature. 
Although this optimisation step is highly combinatoric and therefore often inefficient, this approach does allow for the development of efficient approximate algorithms.
For credal networks under epistemic irrelevance, the situation is more complicated. Basically, the main issue is that the factorisation formula of Bayesian networks no longer holds. Therefore, it is not possible to start from existing algorithms for Bayesian networks and adapt them. Instead, new algorithmic techniques need to be developed from scratch.

The good news though is that once these theoretical challenges are overcome, the resulting algorithms are often surprisingly efficient. Most notably, for credal networks under epistemic irrelevance of which the graphical structure is a tree, there are several types of inference problems that can be solved in polynomial time~\cite{deCooman:2010gd,DeBock:2014ts,Benavoli:2011ei}. This is quite remarkable. For example, it is worth noting that for the algorithm in Reference~\cite{deCooman:2010gd}, the corresponding inference problem in credal networks under strong independence is known to be NP-hard~\cite{Maua:2014ti}. However, despite these promising results, the scope of the algorithms that have so far been developed is rather limited. On the one hand, they require the graph of the network to be a tree. On the other hand, they only consider particular types of inference problems.

The first important aim of this paper is to allow for this scope to be expanded. In order to achieve this, we present a complete theoretical framework for the study of credal networks under epistemic irrelevance. Furthermore, and perhaps most importantly, we demonstrate how this framework can be used to develop new algorithms, for networks of which the graph is not required to be a tree, and for various types of inference problems. Trough the use of numerous examples, we explain what are the crucial techniques that lie at the heart of existing algorithms, and we show how these techniques can be extended to allow for the design of new ones.

A crucial feature of our framework is that we express everything in terms of probabilities and expectations, thereby remaining close to the approach that was taken in the first papers on credal networks under epistemic irrelevance~\citep{Cozman:1998:IIR:2074094.2074105,Cozman:2000ug,deCampos:2007kg}. In contrast, most of the recent work on credal networks under epistemic irrelevance---including some of our own---does not consider probabilities and expectations, but instead considers other, closely related uncertainty models, such as lower previsions and sets of desirable gambles~\citep{deCooman:2010gd,Benavoli:2011ei,DeBock:2014ts,DeBock:2014bv}. On the one hand, this change of perspective has been fruitful, because it has lead to the development of various new theoretical properties.
On the other hand, unfortunately, these alternative uncertainty models are not well known within the Bayesian network community. As a direct result, recent work on credal networks under epistemic irrelevance is rather inaccessible to this community.

The second aim of this paper, therefore, is to reintroduce credal networks under epistemic irrelevance to the Bayesian network community, and to present the most recent theoretical developments in this field in a way that is easily accessible, using the language of probabilities and expectations; lower previsions and sets of desirable gambles are mentioned only in passing. 
Proofs are therefore not provided, because these do require extensive use of these other frameworks for modelling uncertainty. For a more technical exposition that does include proofs, the interested reader is referred to~\citep{DeBock2015:thesis}. 

The remainder of this paper is organised as follows.

 We start in Section~\ref{sec:credal} by providing a brief introduction to the basics of credal networks, discussing in particular their graphical structure---a Directed Acyclic Graph (DAG)---and their local models---sets of probability mass functions, called credal sets. Next, in Section~\ref{sec:complete} and~\ref{sec:irrelevance}, we introduce three particular types of credal networks. Credal networks under complete independence are defined in Section~\ref{sec:complete}, together with another very close related type of credal network, called credal networks under strong independence. 
 Section~\ref{sec:irrelevance} then goes on to introduce our main topic of interest, which are credal networks under epistemic irrelevance. We explain and illustrate some crucial differences with the other two types of credal networks, and provide a formal definition for the corresponding joint uncertainty model, called the irrelevant natural extension. Throughout these two sections, we also pay special attention to the treatment of probability zero, by using full conditional probability measures as our basic uncertainty model.

 From a practical point of view, regardless of the type of credal network that is being considered, the main computational challenge is to obtain tight lower and upper bounds on expectations and probabilities. Conveniently, as we explain in Section~\ref{sec:lowerexpectations}, it suffices to consider lower expectations, because upper expectations and lower and upper probabilities can be obtained as special cases. 
 Furthermore, as we will also discuss, working directly with these bounds offers philosophical advantages, because they are compatible with multiple interpretations.
In the rest of the paper, we focus on the problem of how to compute lower expectations, for the particular case of credal networks under epistemic irrelevance.

 We start in Section~\ref{sec:linearprogramming} by showing how these lower expectations can be obtained as the solution of a linear program, thereby providing a general-purpose method for computing them. Basically, we show that the linear programming methods of Cozman remain valid even if the positivity conditions in Reference~\cite{Cozman:2000ug} are dropped.
 Unfortunately, the size of the required linear programs is exponential in the size of the network, and therefore, this direct approach only works for small networks. 

Therefore, in Section~\ref{sec:theory}, we develop alternative methods that consist in decomposing large inference problems into several smaller sub-problems. In some cases, the remaining sub-problems are trivial local optimisations. In other cases, they can be solved by applying the linear programming techniques of Section~\ref{sec:linearprogramming} to so-called sub-networks. This section also contains several examples, which we use to demonstrate how the algorithms of De Cooman et al.~\cite{deCooman:2010gd} can be extended beyond the case of credal trees. Most importantly, we present theoretical properties that can be used to generalise the methods in this section even further, including marginalisation, factorisation and additivity results, and a law of iterated lower expectation.

In Section~\ref{sec:separation}, we take a step back from all these computational techniques and study the separation properties of credal networks under epistemic irrelevance. In particular, we show that for the irrelevant natural extension of a credal network under epistemic irrelevance, similarly to how d-separation implies independence in a Bayesian network, AD-separation now implies epistemic irrelevance. We also use this result to start a discussion on the relevance of graphoid properties, and on whether or not they should be regarded as axioms.

Finally, in Section~\ref{sec:conditioningversusupdating}, we explain the subtle difference between updating and conditioning and argue how, in the context of updating, it makes sense to replace the conditional lower expectations that we have so far considered by those that correspond to what is called the irrelevant regular extension. We also explain how Lavine's bracketing algorithm can be used to compute updated and conditional lower expectations, and we provide several examples to demonstrate the feasibility of such an approach in the context of credal networks under epistemic irrelevance.

\section{The Basics of Credal Networks}\label{sec:credal}

Basically, a credal network is just a special type of multivariate uncertainty model for a finite set of variables $\{\var{s}\}_{s\in G}$, with $G$ some finite index set. Each of the variables $\var{s}$ takes values $\xval{s}$ in a finite set $\values{s}$ and, for any $S\subseteq G$, we use $\var{S}$ to denote the vector that consists of the variables $\{\var{s}\}_{s\in S}$, which takes values $\xval{S}$ in $\values{S}\coloneqq\times_{s\in S}\values{s}$.

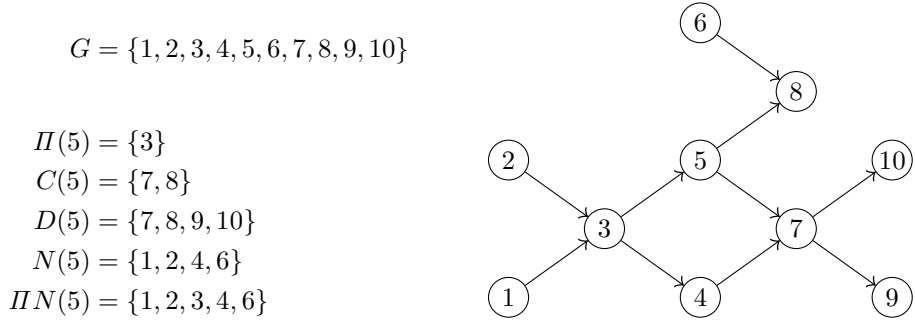
\begin{figure}
\begin{center}
\begin{tikzpicture}[node distance=15pt and 25pt
,>=triangle 45
,decoration={
      markings,
      mark=at position 1 with {\arrow[scale=1.5]{to}};
    }
]
\tikzstyle{punt} = [circle, draw, inner sep=0pt, minimum size=15pt]
\tikzstyle{punt2} = [circle, draw, line width=1.3pt, inner sep=0pt, minimum size=15pt]
\tikzstyle{stip}=[circle, fill, inner sep=0pt, minimum size=4pt]

\node[punt] (s1) {$1$};

\node[punt] (s3) [above right = of s1] {$3$};
\node[punt] (s2) [above left = of s3] {$2$};
\node[punt] (s4) [below right = of s3] {$4$}; 
\node[punt] (s5) [above right = of s3] {$5$};
\node[punt] (s7) [below right = of s5] {$7$};
\node[punt] (s8) [above right = of s5] {$8$};
\node[punt] (s6) [above left = of s8] {$6$};
\node[punt] (s9) [below right = of s7] {$9$};
\node[punt] (s10) [above right = of s7] {$10$};

\node (G) [right = -9.5cm of s6,yshift=-1.8cm] {
\begin{minipage}{2cm}
\begin{align*}
G&=\{1,2,3,4,5,6,7,8,9,10\}\\[20pt]
\parents{5}&=\{3\}\\
\children{5}&=\{7,8\}\\
\descendants{5}&=\{7,8,9,10\}\\
\nonparnondes{5}&=\{1,2,4,6\}\\
\nondes{5}&=\{1,2,3,4,6\}
\end{align*}
\end{minipage}
};

\draw[postaction={decorate}] (s1) -- (s3);
\draw[postaction={decorate}] (s2) -- (s3);
\draw[postaction={decorate}] (s3) -- (s4);
\draw[postaction={decorate}] (s3) -- (s5);
\draw[postaction={decorate}] (s6) -- (s8);
\draw[postaction={decorate}] (s5) -- (s8);
\draw[postaction={decorate}] (s4) -- (s7);
\draw[postaction={decorate}] (s7) -- (s10);

\draw[postaction={decorate}] (s5) -- (s7);
\draw[postaction={decorate}] (s7) -- (s9);

\end{tikzpicture}
\vspace{15pt}
\caption{Example of a directed acyclic graph (DAG).}
\label{fig:DAG}
\end{center}
\end{figure}

In a credal network, just like in a Bayesian network, the variables $\{\var{s}\}_{s\in G}$ are identified with their indices $s\in G$. These indices are then interpreted as the nodes of a directed acyclic graph (DAG)---see Figure~\ref{fig:DAG} for an example---and the arrows of this graph are taken to represent (in)dependencies among the individual variables. Finally, these assessments of independence are combined with local uncertainty models, and in this way, a global uncertainty model for $\var{G}$ is defined. The main difference with a Bayesian network is that the local and global uncertainty models are now \emph{sets} of probability distributions, and that the interpretation of independence may no longer coincide with the traditional one.

In order to formalize this idea, we need some basic graph-theoretic concepts, which are illustrated in Figure~\ref{fig:DAG}. 
For two nodes $s$ and $u$ in $G$, if there is a directed edge from $s$
to $u$, we denote this as $s\rightarrow u$ and say that $s$ is a \emph{parent} of $u$ and $u$ is a \emph{child} of $s$. 
For any node $s$, its set of parents is denoted by
$\parents{s}$ and its set of children by $\children{s}$. 
A node $s$ is said to \emph{precede} a node $v$, denoted by $s\precedes v$, if it is possible to start from $s$ and follow the edges of the graph along their
direction to reach $v$.  
If $s\precedes v$ and $s\neq v$, we say that $s$ \emph{strictly
  precedes} $v$ and write $s\sprecedes v$. For any node $s$, we call
\mbox{$\descendants{s}\coloneqq\{v\in\nodes\colon s\sprecedes v\}$} its set of \emph{descendants} and \mbox{$\nonparnondes{s}\coloneqq\nodes\setminus(\parents{s}\cup\{s\}\cup\descendants{s})$} its set of \emph{non-parent non-descendants}. 
We also use the shorthand notation
\mbox{$\nondes{s}\coloneqq\parents{s}\cup\nonparnondes{s}=\nodes\setminus(\{s\}\cup\descendants{s})$} to refer to what we call the \emph{non-descendants} of $s$. 
By definition, the graph is assumed to be acyclic. Informally, this means that it is not possible to start in a node, follow the edges along their direction, and end up in the same node one started out from. More technically, this means that there are no two nodes $s,v\in G$ such that $s\sprecedes v$ and $v\sprecedes s$. 
With this terminology in place, we can now formally introduce the two main building blocks of a credal network, which are local uncertainty models and assessments of independence. 

The basic premise of a credal network is that it is sometimes unrealistic to provide exact values for the local probabilities $P(\xval{s}\newmid\xval{\parents{s}})$ that are required to specify a Bayesian network. Therefore, in those cases, the local uncertainty models of a credal network are taken to be sets of probability distributions. For every variable $\var{s}$ and every instantiation $\xval{\parents{s}}$ of its parent variables $\var{\parents{s}}$, the associated local uncertainty model is a set $\credal_{s\mid\xval{\parents{s}}}$ of probability mass functions on $\values{s}$, the elements of which are regarded as candidates for some ideal---but unknown---conditional probability mass function $P(\var{s}\newmid\xval{\parents{s}})$, in the sense that
\begin{equation}\label{eq:local:assessments}
P(\var{s}\newmid\xval{\parents{s}})
\in
\credal_{s\mid\xval{\parents{s}}}
\text{~for all~}
s\in G
\text{ and }
\xval{\parents{s}}\in\values{\parents{s}}.
\end{equation}
Most authors require the sets $\credal_{s\mid\xval{\parents{s}}}$ to be closed and convex, and then call them \emph{credal sets}; we will follow this convention here as well. In practice, these local credal sets can be obtained in various ways~\cite{Walley:1991vk}; they can for example be elicited from experts~\cite{Piatti:2010vd}, learned from data~\cite{Bernard:2005hd,Walley:1996vt}, or constructed as some type of neighbourhood model~\cite{berger1985}. 

Mathematically, since the local credal sets are closed convex subsets of the---bounded---set $\Sigma_{\values{s}}$ of all probability mass functions on $\values{s}$, they can be represented using either constraints or vertices. A vertex-based representation consists in characterising $\credal_{s\mid\xval{\parents{s}}}$ by means of its set of extreme points
\begin{equation*}
\mathrm{Ext}\,\credal_{s\mid\xval{\parents{s}}}
\coloneqq
\big\{
p\in\credal_{s\mid\xval{\parents{s}}}
\colon
p\notin\mathrm{Conv}\big(\credal_{s\mid\xval{\parents{s}}}\setminus\{p\}\big)
\big\},
\vspace{3pt}
\end{equation*}
where $\mathrm{Conv}$ denotes the convex hull. This set of extreme points $\mathrm{Ext}\,\credal_{s\mid\xval{\parents{s}}}$ is the smallest subset of $\credal_{s\mid\xval{\parents{s}}}$ that has $\credal_{s\mid\xval{\parents{s}}}$ as its convex hull. If a credal set has a finite number of extreme points, then we say that it is \emph{finitely generated}. Such credal sets are polytopes of which the vertices are the extreme points. 
A constraint-based representation of a credal set consists in characterising it as the largest set of probability mass functions on $\values{s}$ that satisfies a set of linear constraints:
\begin{equation}\label{eq:originallocalconstraints}
p\in\credal_{s\mid\xval{\parents{s}}}
\Leftrightarrow~~
p\in\Sigma_{\values{s}}
\text{~~and~~}
(\forall i\in\mathcal{I})
\sum_{x_s\in\values{s}}p(x_s)\alpha_i(x_s)\geq\beta_i,
\end{equation}
where $\mathcal{I}$ is a---possibly infinite---index set and, for all $i\in\mathcal{I}$ and $x_s\in\values{s}$, $\alpha_i(x_s)$ and $\beta_i$ are real numbers. A credal set is finitely generated if and only if it admits a constrained-based representation that requires only a finite number of constraints.\footnote{This equivalence follows from the fact that a compact convex set can be specified as the
intersection of a finite number of closed half spaces if and only if
it is the convex hull of a finite number of vertices \cite[Theorem~3.1.3]{Grunbaum:2003vq}.} In this finitely generated case, the conversion between vertex-based and constrained-based representations can be achieved using various algorithms.\footnote{Constructing a vertex-based representation is called the vertex enumeration problem, whereas constructing a constrained-based representation is called the facet enumeration problem.} Unfortunately, the complexity of these algorithms will in general be exponential in the number of vertices and/or constraints. However, in practice, for the local credal sets of a credal network, these conversions tend to be feasible.

In the remainder of this paper, we assume that the local credal sets $\credal_{s\mid\xval{\parents{s}}}$ are given; they constitute the first main building block of the credal networks that we will consider. Our theoretical developments do not require these local credal sets to be finitely generated. However, in order to apply our results in practice, such an assumption will nevertheless sometimes be required, in which case we will mention it explicitly.

The second main building block of the credal networks that we consider is a collection of independence assessments. As in Bayesian networks, these independence assessments are inferred from the graph of the network in the following way: every variable $\var{s}$ is assumed to be conditionally independent of its non-parent non-descendants $\var{\nonparnondes{s}}$ given its parents $\var{\parents{s}}$. However, in the context of sets of probability distributions, there is no consensus on what is meant here by independence. Depending on the notion of independence that is chosen, a different type of credal network is obtained.

\section{Credal Networks under Complete Independence}\label{sec:complete}

The most straightforward way to define independence for a set of distributions, is to simply impose the usual notion of independence, which we will henceforth call \emph{stochastic independence}, to each of its elements $P(\var{G})$. This type of independence---element-wise stochastic independence---is called \emph{complete independence}~\cite{Cozman:2012fc,seidenfeld2007}. In the case of credal networks, this results in the following assessment:
\begin{equation}\label{eq:stochastic}
P(\var{s}\newmid\xval{\nondes{s}})
=
P(\var{s}\newmid\xval{\parents{s}})
\text{~for all~}
s\in G
\text{ and }
\xval{\nondes{s}}\in\values{\nondes{s}}.
\end{equation}
Conventionally, the conditional probabilities in this expression are taken to be derived from $P(\var{G})$ through Bayes's rule. However, this creates issues in the case of probability zero; for example, if $P(\xval{\nondes{s}})=0$, then $P(\var{s}\newmid\xval{\nondes{s}})$ is ill-defined.

In order to avoid these issues in an elegant yet rigorous way, we will not regard conditional probabilities as derived concepts that are obtained through Bayes's rule, but rather as primitive notions that are part of a (full) conditional probability measure~\citep{Dubins:1975ej}.

\begin{definition}\label{def:cond_prob}
A full conditional probability measure $P$ on a finite set $\Omega$ is a map
\begin{equation*}
P \colon \mathcal{P}(\Omega)\times\mathcal{P}_{\emptyset}(\Omega)\to \reals \colon (A,B)\to P(A\newmid B),
\end{equation*}
with $\mathcal{P}_{\emptyset}(\Omega)\coloneqq\mathcal{P}(\Omega)\setminus\{\emptyset\}$,
such that for any $A,C\in\mathcal{P}(\Omega)$ and $B\in\mathcal{P}_\emptyset(\Omega)$:
\begin{enumerate}[label=\emph{F\arabic*:},
ref=F\arabic*]
\item
$P(\,\cdot\,\vert B)$ is a probability measure on $\mathcal{P}(\Omega)$ with $P(B\vert B)=1$;\label{def:cond_prob:measure}
\item
$P(A\cap C\vert B)=P(A\vert C\cap B)P(C\vert B)$ if $C\cap B\neq\emptyset$.\label{def:cond_prob:Bayes}\\[-11pt]
\end{enumerate}
\end{definition}
The axioms \ref{def:cond_prob:measure} and~\ref{def:cond_prob:Bayes} correspond to the usual rules of probability. The only difference is that Bayes's rule---\ref{def:cond_prob:Bayes}---does not define conditional probabilities by means of division, but instead regards them as primitive notions and requires them to satisfy a product rule.

In the case of a credal network, we model the uncertainty about $\var{G}$ by means of a---possibly partially specified---full conditional probability measure $P$ on $\values{G}$. For any event $A\in\powerG$ and any non-empty event $B\in\eventsG$, $P(A\vert B)$ is the probability of $A$ conditional on~$B$ and $P(A)\coloneqq P(A\vert\values{G})$ is the (unconditional) probability of $A$. We will mostly focus on events of the form $\{\zval{G}\in\values{G}\colon \zval{S}=\xval{S}\}$, with $S\subseteq G$; for ease of notation, we denote these events as $x_S$. The events $x_s$, $x_{\parents{s}}$, $x_{\nondes{s}}$, $x_G$ and $x_{\emptyset}=\values{G}$ correspond to special cases.

Within this framework, Equation~\eqref{eq:stochastic} is now simply a constraint on the full conditional probability measure $P$ and does not require---nor suffers from---divisions by zero. As is very well known from the theory of Bayesian networks, this constraint implies that the unconditional global probabilities $P(\xval{G})$ are completely determined by the local probabilities $P(\xval{s}\newmid\xval{\parents{s}})$: 
\begin{equation}\label{eq:factorisation}
P(\xval{G})=\prod_{s\in G}P(\xval{s}\newmid\xval{\parents{s}})
\text{~~for all~~}
\xval{G}\in\values{G}.
\end{equation}
This is not necessarily true for conditional probabilities. For example, there may be some $S,T\subseteq G$, $\xval{S}\in\values{S}$ and $\xval{T}\in\values{T}$ such that $P(\xval{S}\newmid\xval{T})$ is not uniquely determined by Equation~\eqref{eq:stochastic} and the local probabilities. However, this is usually ignored; the theory of Bayesian networks focusses on cases where $P(\xval{T})>0$, which guarantees that $P(\xval{S}\newmid\xval{T})$ can be computed by means of Bayes's rule---\ref{def:cond_prob:Bayes}. This restricted focus is unfortunate because, even if $\xval{T}$ has probability zero, $P(\xval{S}\newmid\xval{T})$ is often still uniquely determined by the local probabilities and Equation~\eqref{eq:stochastic}.

In any case, for credal networks, probability zero is not the only source of non-uniqueness. Indeed, the local probabilities may themselves not be unique, because Equation~\eqref{eq:local:assessments} will in general---unless all the local credal sets are singletons---only impose partial constraints on the local probabilities. Due to this inherent non-uniqueness, a credal network does not correspond to a single full conditional probability measure, but rather to a set of them. We will denote such a set of full conditional probability measures on $\values{G}$ by $\mathcal{F}_G$, and for any $B\in\mathcal{P}_{\emptyset}(\values{G})$ and any $S\subseteq G$, we will then use $\mathcal{F}_G(\var{S}\newmid B)$ to refer to the set of probability mass functions $\{P(\var{S}\newmid B)\colon P\in\mathcal{F}_G\}$. The sets $\mathcal{F}_G(X_s\newmid\xval{\parents{s}})$, $\mathcal{F}_G(X_s\newmid\xval{\nondes{s}})$ and $\mathcal{F}_G(X_G)\coloneqq\mathcal{F}_G(X_G\newmid\values{G})$ correspond to important special cases.

The largest set of (full conditional) probability measures that is compatible with the defining constraints of a credal network is called its \emph{extension}. For a credal network under complete independence, these defining constraints are Equations~\eqref{eq:local:assessments} and~\eqref{eq:stochastic}, and the corresponding extension is called the \emph{complete extension}. We will denote this complete extension by $\mathcal{F}^{\mathrm{com}}_G$. Clearly, if we let $\mathcal{F}^*_G$ be the set of all full conditional probability measures on $\values{G}$, then $\mathcal{F}^{\mathrm{com}}_G$ is given by
\begin{equation}\label{eq:complete:explicit}
\mathcal{F}^{\mathrm{com}}_G
=
\Big\{
P\in\mathcal{F}^*_G
\colon
\big(\forall s\in G\big)\,
\big(\forall \xval{\nondes{s}}\in\values{\nondes{s}}\big)~
P(\var{s}\newmid\xval{\nondes{s}})
=
P(\var{s}\newmid\xval{\parents{s}})
\in
\credal_{s\mid\xval{\parents{s}}}
\Big\}.
\end{equation}
If we make abstraction of the `full conditional' aspects that we have added, and focus on the unconditional part $\mathcal{F}^{\mathrm{com}}_G(\var{G})$, then this complete extension is simply the set of all Bayesian networks whose local probability mass functions $P(\var{s}\newmid\xval{\parents{s}})$ take values in the local credal sets $\credal_{s\mid\xval{\parents{s}}}$. This approach is highly intuitive if one is convinced that the uncertainty about $\var{G}$ can be modelled by means of a single Bayesian network and, for some reason, the required local probability mass functions are not exactly known, but are only partially specified. It seems reasonable to model this type of situation by means of a set of Bayesian models, and it should therefore not be surprising that credal networks under complete independence were the first type of credal network to be considered; see~\citep{Fagiuoli:1998ft}. In fact, at that point in time, since there were no other types yet, credal networks under complete independence were simply called credal networks.

Nevertheless, today, rather surprisingly, credal networks under complete independence are almost never considered. Instead, the majority of work on credal networks considers what are called credal networks under \emph{strong independence}. We will not go into details here; for our present purposes, it suffices to know that the unconditional part $\mathcal{F}^{\mathrm{str}}_G(\var{G})$ of the extension of a credal network under strong independence, which is called the strong extension of the network, is equal to the convex hull of the unconditional part $\mathcal{F}^{\mathrm{com}}_G(\var{G})$ of the complete extension. 
Given the choice between these two extensions, I favour the complete extension, because of its clear and intuitive sensitivity analysis interpretation, which the strong extension does not have. I fail to understand why most authors prefer the strong extension instead. In any case, the choice is mainly a philosophical one, because in practice, there is little difference between the two approaches. Essentially,~Fagiuoli and Zaffalon~\cite[Theorem~5]{Fagiuoli:1998ft} already showed that for the vast majority of commonly considered parameters of interest---such as (conditional) lower and upper probabilities and expectations---it makes no difference whether we compute them with respect to the complete extension or its convex hull---the strong extension. Therefore, existing algorithms that have been developed for strong extensions can be immediately applied to complete extensions as well.

\section{Credal Networks under Epistemic Irrelevance}\label{sec:irrelevance}

In this paper, we do not impose complete or strong independence, but instead impose the following assessments of epistemic irrelevance:
\begin{equation}\label{eq:irrelevance:assessment}
\mathcal{F}_G(X_s\newmid \xval{\parents{s}})=\mathcal{F}_G(X_s\newmid \xval{\nondes{s}})
\text{~for all~}
s\in G
\text{ and }
\xval{\nondes{s}}\in\values{\nondes{s}}.
\end{equation}
For every $s\in G$, this assessment requires that, conditional on $\var{\parents{s}}$, $\var{\nonparnondes{s}}$ should be epistemically irrelevant to $\var{s}$. 
The idea here is that since we are modelling uncertainty by means of sets of (full conditional) probability measures, independence should be a statement about such sets, not about the individual elements of these sets. Equation~\eqref{eq:irrelevance:assessment} imposes that given the value $\xval{\parents{s}}$ of its parents, our beliefs about the variable $\var{s}$ remain identical if we are also given the value $\xval{\nonparnondes{s}}$ of its non-parent non-descendants. The only difference with the more conventional notions of---stochastic, complete or strong---independence lies in the fact that beliefs are now no longer identified with individual probability distributions, but rather with the information that is available about these distributions, that is, with sets of probabilities---or, equivalently, with constraints on probabilities.

\begin{figure}[t]
\begin{center}
\begin{tikzpicture}[node distance=15pt and 15pt
,>=triangle 45
,decoration={
      markings,
      mark=at position 1 with {\arrow[scale=1.5]{to}};
    }
]
\tikzstyle{punt} = [circle, draw, inner sep=0pt, minimum size=15pt]
\tikzstyle{punt2} = [circle, draw, line width=1.3pt, inner sep=0pt, minimum size=15pt]
\tikzstyle{stip}=[circle, fill, inner sep=0pt, minimum size=4pt]

\node[punt] (s1) {$1$};

\node[punt] (s2) [right = of s1] {$2$};
\node[punt] (s3) [right = of s2] {$3$};

\node (G) [right = -0.6cm of s1,yshift=-0.8cm] {
\begin{minipage}{2cm}
\begin{align*}
G=\{1,2,3\}
\end{align*}
\end{minipage}
};

\node (ps1) [right = 1cm of s3,yshift=-0.3cm] {
\begin{minipage}{2cm}
\begin{align*}
\parents{1}&=\emptyset\\
\nonparnondes{1}&=\emptyset\\
\nondes{1}&=\emptyset
\end{align*}
\end{minipage}
};
\node (ps2) [right = 4cm of s3,yshift=-0.3cm] {
\begin{minipage}{2cm}
\begin{align*}
\parents{2}&=\{1\}\\
\nonparnondes{2}&=\emptyset\\
\nondes{2}&=\{1\}
\end{align*}
\end{minipage}
};
\node (ps3) [right = 7cm of s3,yshift=-0.3cm] {
\begin{minipage}{2cm}
\begin{align*}
\parents{3}&=\{2\}\\
\nonparnondes{3}&=\{1\}\\
\nondes{3}&=\{1,2\}
\end{align*}
\end{minipage}
};

\draw[postaction={decorate}] (s1) -- (s2);
\draw[postaction={decorate}] (s2) -- (s3);
\end{tikzpicture}
\vspace{4pt}
\caption{The DAG of an imprecise Markov chain with three nodes. 
}\label{fig:chainwiththreenodes}
\vspace{-11pt}
\end{center}
\end{figure}
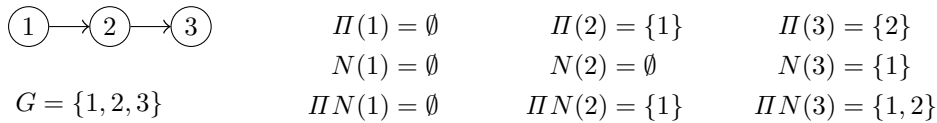

\begin{example}\label{exmp:chainwiththreenodes}
Consider a credal network whose DAG is depicted in Figure~\ref{fig:chainwiththreenodes}, which is an imprecise Markov chain~\cite{deCooman:2009jz} of length three. In this case, the epistemic irrelevance constraints in Equation~\eqref{eq:irrelevance:assessment} are trivially satisfied for $s=1$ and $s=2$, because for those two nodes, $\parents{s}=\nondes{s}$. For $s=3$, we get the following assessment:
\begin{equation}\label{eq:exmp:chainwiththreenodes}
\mathcal{F}_G(X_3\newmid \xval{2})=\mathcal{F}_G(X_3\newmid \xval{2},\xval{1})
\text{~for all~}
\xval{2}\in\values{2}
\text{~and~}
\xval{1}\in\values{1}.
\vspace{7pt}
\end{equation}
What this assessment says is that the probability mass functions $P(X_3\newmid \xval{2})$ and $P(X_3\newmid \xval{2},\xval{1})$ take values in the same \emph{set} of probability mass functions---or, equivalently, that they satisfy the same constraints. It does not say however that they are equal: $P(X_3\newmid \xval{2})$ and $P(X_3\newmid \xval{2},\xval{1})$ are not forced to be identical.\hfill$\lozenge$
\end{example}

As illustrated in this example, an assessment of epistemic irrelevance does not imply stochastic independence: $P(\var{s}\newmid\xval{\nondes{s}})$ does not need to be equal to $P(\var{s}\newmid\xval{\parents{s}})$, and can furthermore depend on $\xval{\nonparnondes{s}}$ in arbitrary ways. However, epistemic irrelevance does constrain the extent to which $P(\var{s}\newmid\xval{\nondes{s}})$ can differ from $P(\var{s}\newmid\xval{\parents{s}})$, because it requires that they should both belong to the same set. In this sense, epistemic irrelevance imposes a notion of `almost' stochastic independence, which, in those applications where stochastic independence is an approximation that is imposed out of mathematical convenience, provides a more realistic alternative.

Together, the assessments of epistemic irrelevance in Equation~\eqref{eq:irrelevance:assessment} and the constraints that are imposed by the local credal sets in Equation~\eqref{eq:local:assessments} are the two main building blocks of what is called a \emph{credal network under epistemic irrelevance}.
In the context of sets of full conditional probability measures, the latter---that is, Equation~\eqref{eq:local:assessments}---can be reformulated as follows:
\begin{equation}\label{eq:local:assessments:full}
\mathcal{F}_G(X_s\newmid \xval{\parents{s}})\subseteq\credal_{s\vert\xval{\parents{s}}}
\text{~for all~}
s\in G
\text{ and }
\xval{\parents{s}}\in\values{\parents{s}}.
\end{equation}
The largest set $\mathcal{F}_G$ of full conditional probability measures on $\values{G}$ that is compatible with a credal network under epistemic irrelevance---that satisfies Equations~\eqref{eq:irrelevance:assessment} and~\eqref{eq:local:assessments:full}---is called the \emph{irrelevant natural extension} of the credal network. 
We will denote this extension by $\mathcal{F}^{\mathrm{irr}}_G$. As shown in~\citep{Cozman:2000ug} under strict positivity conditions, and more generally in~\citep{DeBock2015:thesis}, $\mathcal{F}^{\mathrm{irr}}_G$ is given by
\begin{equation}\label{eq:epistemic:explicit}
\mathcal{F}^{\mathrm{irr}}_G
=
\Big\{
P\in\mathcal{F}^*_G
\colon
\big(\forall s\in G\big)\,
\big(\forall \xval{\nondes{s}}\in\values{\nondes{s}}\big)~
P(\var{s}\newmid\xval{\nondes{s}})
\in
\credal_{s\mid\xval{\parents{s}}}
\Big\}.
\end{equation}

If all the local credal sets $\credal_{s\vert\xval{\parents{s}}}$ are singletons, Equation~\eqref{eq:local:assessments:full} implies that $\mathcal{F}_G(X_s\newmid \xval{\parents{s}})$ contains only a single probability mass function. Therefore, in this particular case, Equations~\eqref{eq:irrelevance:assessment} and~\eqref{eq:stochastic} will be equivalent, and $\mathcal{F}^{\mathrm{irr}}_G$ and $\mathcal{F}^{\mathrm{com}}_G$ will coincide. However, in most other cases, as can be seen by comparing Equations~\eqref{eq:complete:explicit} and~\eqref{eq:epistemic:explicit}, the complete extension $\mathcal{F}^{\mathrm{com}}_G$ will be a strict subset of the irrelevant natural extension $\mathcal{F}^{\mathrm{irr}}_G$. In fact, it can easily be shown that $\mathcal{F}^{\mathrm{com}}_G$ satisfies all the epistemic irrelevance assessments in Equation~\eqref{eq:irrelevance:assessment}. However, as we already illustrated in Example~\ref{exmp:chainwiththreenodes}, the converse is usually not true because the full conditional probability measures in $\mathcal{F}^{\irr}_G$ are not forced to satisfy Equation~\eqref{eq:stochastic}. The full conditional probability measures that do not satisfy Equation~\eqref{eq:stochastic} are exactly the ones that belong to $\mathcal{F}^{\irr}_G\setminus\mathcal{F}^{\mathrm{com}}_G$. The following example provides a more concrete illustration of this crucial difference between $\mathcal{F}^{\irr}_G$ and $\mathcal{F}^{\mathrm{com}}_G$.

\begin{figure}[t]
\begin{center}
\begin{tikzpicture}[node distance=15pt and 15pt
,>=triangle 45
,decoration={
      markings,
      mark=at position 1 with {\arrow[scale=1.5]{to}};
    }
]
\tikzstyle{punt} = [circle, draw, inner sep=0pt, minimum size=15pt]
\tikzstyle{punt2} = [circle, draw, line width=1.3pt, inner sep=0pt, minimum size=15pt]
\tikzstyle{stip}=[circle, fill, inner sep=0pt, minimum size=4pt]

\node[punt] (s1) {$1$};

\node[punt] (s2) [right = of s1] {$2$};

\node (G) [right = -0.8cm of s1,yshift=-0.8cm] {
\begin{minipage}{2cm}
\begin{align*}
G=\{1,2\}
\end{align*}
\end{minipage}
};

\node (ps1) [right = 0.2cm of s3,yshift=-0.3cm] {
\begin{minipage}{2cm}
\begin{align*}
\parents{1}&=\emptyset\\
\nonparnondes{1}&=\{2\}\\
\nondes{1}&=\{2\}
\end{align*}
\end{minipage}
};
\node (ps2) [right = 3.2cm of s3,yshift=-0.3cm] {
\begin{minipage}{2cm}
\begin{align*}
\parents{2}&=\emptyset\\
\nonparnondes{2}&=\{1\}\\
\nondes{2}&=\{1\}
\end{align*}
\end{minipage}
};
\end{tikzpicture}
\vspace{4pt}
\caption{The DAG of a credal network with two disconnected nodes. 
}\label{fig:twodisconnectednodes}
\vspace{-11pt}
\end{center}
\end{figure}
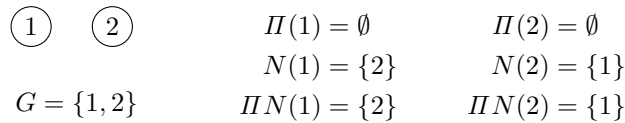

\begin{example}\label{exmp:extensionoftwodisconnectednodes}
Consider a credal network whose DAG consists of two disconnected nodes, as depicted in Figure~\ref{fig:twodisconnectednodes}. For each of the two nodes 
$s\in G=\{1,2\}$, the corresponding variable $\varel$ assumes values
in its state space $\valuesel$ and has an associated local uncertainty
model in the form of a credal set $\credal_s$, which is a closed and convex set of probability mass functions on $\valuesel$.\footnote{Since in this particular case, $\parents{s}=\emptyset$, we can omit $\xval{\parents{s}}$ from the notation, by defining $\credal_s\coloneqq\credal_{s\mid\xval{\parents{s}}}$.} 
By applying Equation~\eqref{eq:epistemic:explicit}, we find that the irrelevant natural extension of such a network is given by
\begin{equation}\label{eq1:exmp:extensionoftwodisconnectednodes}
\mathcal{F}^{\mathrm{irr}}_{\{1,2\}}
=
\Big\{
P\in\mathcal{F}^*_{\{1,2\}}
\colon
\big(\forall \xval{2}\in\values{2}\big)~
P(\var{1}\newmid\xval{2})
\in
\credal_{1}
\text{~and~}
\big(\forall \xval{1}\in\values{1}\big)~
P(\var{2}\newmid\xval{1})
\in
\credal_{2}
\Big\}.
\end{equation}
In order to better understand this expression, we will in the remainder of this example focus on the case where $X_1$ and $X_2$ are both binary, with $\values{1}=\{h_1,t_1\}$ and $\values{2}=\{h_2,t_2\}$.

What makes a credal set $\credal_s$ on such a binary state space $\values{s}=\{h_s,t_s\}$ particularly elegant, is that it can be uniquely characterised by the lower and upper probability of one of its two states. For example, if we consider the lower and upper probability of $h_s$, as defined by
\begin{equation*}
\underline{p}(h_s)\coloneqq\min\{p(h_s)\colon p\in\credal_s\}
\text{~~~and~~~}\overline{p}(h_s)\coloneqq\max\{p(h_s)\colon p\in\credal_s\},
\end{equation*}
then $\credal_s$ is the largest set of probability mass functions on $\values{s}$ such that $p(h_s)$ complies with these bounds, in the sense that
\begin{equation}\label{eq2:exmp:extensionoftwodisconnectednodes}
p\in\credal_s
~~\Leftrightarrow~~\Big(~
p\in\simplex{s}
\text{~and~}
p(h_s)\in\big[\,\underline{p}(h_s),\overline{p}(h_s)\big]
~\Big).
\vspace{4pt}
\end{equation}
Of course, by symmetry, this is also true if we replace $h_s$ by $t_s$. The reason why these two characterisations are equivalent is the unitary constraint $p(h_s)+p(t_s)=1$: as a direct consequence of this unitary constraint, it follows that $\underline{p}(t_s)=1-\overline{p}(h_s)$ and $\overline{p}(t_s)=1-\underline{p}(h_s)$, which in turn implies that $\underline{p}(h_s)\leq p(h_s)\leq\overline{p}(h_s)$ is equivalent to $\underline{p}(t_s)\leq p(t_s)\leq\overline{p}(t_s)$.

By combining Equations~\eqref{eq1:exmp:extensionoftwodisconnectednodes} and~\eqref{eq2:exmp:extensionoftwodisconnectednodes}, we find that in this binary case, the irrelevant natural extension $\mathcal{F}^{\mathrm{irr}}_{\{1,2\}}$ is the set of all full conditional probability measures $P\in\mathcal{F}^*_{\{1,2\}}$ on $\values{G}=\values{1}\times\values{2}$ such that
\begin{equation*}
P(h_1\newmid h_2)
\in
\big[\,\underline{p}(h_1),\overline{p}(h_1)\big]
\text{~~and~~}
P(h_1\newmid t_2)
\in
\big[\,\underline{p}(h_1),\overline{p}(h_1)\big]\vspace{-5pt}
\end{equation*}
and
\begin{equation*}
P(h_2\newmid h_1)
\in
\big[\,\underline{p}(h_2),\overline{p}(h_2)\big]
\text{~~and~~}
P(h_2\newmid t_1)
\in
\big[\,\underline{p}(h_2),\overline{p}(h_2)\big],\vspace{8pt}
\end{equation*}
which is completely determined by four numbers: $\underline{p}(h_1)$, $\overline{p}(h_1)$, $\underline{p}(h_2)$ and $\overline{p}(h_2)$. 

For the sake of this example, let us assume that these four numbers are equal to $\nicefrac{1}{4}$, $\nicefrac{3}{4}$, $\nicefrac{1}{4}$ and $\nicefrac{3}{4}$, respectively. The corresponding irrelevant natural extension $\mathcal{F}^{\mathrm{irr}}_{\{1,2\}}$ is then the set of all full conditional probability measures $P\in\mathcal{F}^*_{\{1,2\}}$ on $\values{G}=\values{1}\times\values{2}$ such that
\begin{equation}\label{eq3:exmp:extensionoftwodisconnectednodes}
P(h_1\newmid h_2)
\in
\big[\,\nicefrac{1}{4},\nicefrac{3}{4}\big],~
P(h_1\newmid t_2)
\in
\big[\,\nicefrac{1}{4},\nicefrac{3}{4}\big],~
P(h_2\newmid h_1)
\in
\big[\,\nicefrac{1}{4},\nicefrac{3}{4}\big]
\text{~~and~~}
P(h_2\newmid t_1)
\in
\big[\,\nicefrac{1}{4},\nicefrac{3}{4}\big].
\end{equation}
One example of such a full conditional probability measure is the one that has the following joint probability values:
\begin{equation}\label{eq4:exmp:extensionoftwodisconnectednodes}
P(h_1,h_2)=\nicefrac{1}{8},~
P(h_1,t_2)=\nicefrac{3}{8},~
P(t_1,h_2)=\nicefrac{3}{8}
\text{~and~}
P(t_1,t_2)=\nicefrac{1}{8}.\vspace{6pt}
\end{equation}
Since all these probabilities are strictly positive, they determine a unique full conditional probability measure on $\values{G}=\values{1}\times\values{2}$ of which, for all $A\in\mathcal{P}(\values{G})$ and $B\in\mathcal{P}_\emptyset(\values{G})$, $P(A\vert B)$ can be derived from~\ref{def:cond_prob:measure},~\ref{def:cond_prob:Bayes} and Equation~\eqref{eq4:exmp:extensionoftwodisconnectednodes}.\footnote{This would not be true if, for example, $P(h_1,h_2)=P(h_1,t_2)=0$ and $P(t_1,h_2)=P(t_1,t_2)=\nicefrac{1}{2}$. In that case, since $P(h_1)=P(h_1,h_2)+P(h_1,t_2)=0$, $P(h_2\vert h_1)$ would not be uniquely determined by Bayes' rule---\ref{def:cond_prob:Bayes}.} In particular, we find that $P(h_1)=P(t_1)=P(h_2)=P(t_2)=\nicefrac{1}{2}$ and therefore, that
\begin{equation*}
P(h_1\newmid h_2)=\nicefrac{1}{4},~
P(h_1\newmid t_2)=\nicefrac{3}{4},~
P(h_2\newmid h_1)=\nicefrac{1}{4}
\text{~~and~~}
P(h_2\newmid t_1)=\nicefrac{3}{4}.
\end{equation*}
On the one hand, this implies that $P$ satisfies Equation~\eqref{eq3:exmp:extensionoftwodisconnectednodes}, and therefore, that it is indeed an element of $\mathcal{F}^{\mathrm{irr}}_{\{1,2\}}$. On the other hand, it also illustrates that $P$ fails Equation~\eqref{eq:stochastic}, and therefore, that it does not belong to $\mathcal{F}^{\mathrm{com}}_{\{1,2\}}$. For example, we find that $P(h_1\vert h_2)\neq P(h_1)$, or equivalently---since $P(h_2)>0$---that $P(h_1,h_2)\neq P(h_1)P(h_2)$, which implies that, according to $P$, $X_1$ and $X_2$ are not independent.
\hfill$\lozenge$
\end{example}

In the example above, the epistemic irrelevance assessments that are imposed by the graph are symmetric: $X_1$ is epistemically irrelevant to $X_2$, and vice versa. However, this is not always the case. For example, in the imprecise Markov chain in Example~\ref{exmp:chainwiththreenodes}, conditional on $X_2$, the graph of the network imposes that $X_1$ should be epistemically irrelevant to $X_3$, but not the other way around. For a standard---precise---Markov chain, where epistemic irrelevance is replaced by stochastic independence, these two assessments would be equivalent, because stochastic independence is symmetric.\footnote{Provided that all the relevant conditional events have a strictly positive probability. Otherwise, depending on the specific definition of (stochastic) independence that is adopted, symmetry may still fail. In fact, in the presence of conditioning events with probability zero, our definition of stochastic independence---see Equation~\eqref{eq:stochastic}---is asymmetric.} However, this is not the case for epistemic irrelevance, because it is an intrinsicly asymmetric assessment~\cite{Couso:1999wh}.
At first sight, this asymmetry  may seem very peculiar, and perhaps even undesired. 
However, as we are about to argue, this aspect is not strange at all and is in fact very natural, for the following reasons. 

First of all, there simply seems to be no fundamental reason why an assessment of epistemic irrelevance should lead us to adopt the reverse version. As Dawid~\cite{Dawid:1997je} put it: ``The desirability of the symmetry property is not so obvious: if learning $Y$ is irrelevant to $X$, must it follow that learning $X$ is irrelevant to $Y$?''. We agree with Dawid that this is indeed not obvious, and that therefore, there is no reason why it should be imposed by definition. Of course, this does not exclude that there might be instances where it makes sense to impose mutual irrelevance. In fact, as we have seen in Example~\ref{exmp:extensionoftwodisconnectednodes}, we will sometimes do so.

Secondly, the assessments of epistemic irrelevance that we make---in the specific direction that we make them---have an intuitive meaning in terms of the local models that we introduced in the previous section. For example, Equation~\eqref{eq:irrelevance:assessment} is not just an assessment of epistemic irrelevance. In combination with Equation~\eqref{eq:local:assessments:full}, it can also be interpreted as an additional collection of direct assessments: for any $\xval{\nondes{s}}$, it implies hat $\mathcal{F}_G(X_s\newmid \xval{\nondes{s}})$ should be a subset of $\credal_{s\vert\xval{\parents{s}}}$. 
Basically, the effect of our epistemic irrelevance assessments is therefore that the local assessments are duplicated and applied to a larger class of conditional distributions.

 Thirdly, to a practitioner who constructs a credal---or Bayesian---network, the direction of the arrows matters intuitively. Suppose for example that in Example~\ref{exmp:chainwiththreenodes}, $\var{1}$ represents whether or not someone is a smoker, $\var{2}$ represents whether or not someone has lung cancer and $\var{3}$ represents whether or not some medical test indicates that the person has lung cancer. Then most people would put the edges in Figure~\ref{fig:chainwiththreenodes} as we have put them, in that particular direction, and would feel that changing the direction would alter the meaning of the assessment. Nevertheless, in a Bayesian network, because stochastic independence is symmetric, these two graphs---ours and the one where the arrows are reversed---correspond to the same assessments of independence. In technical parlance: they are Markov-equivalent. 
For credal networks under epistemic irrelevance, no such equivalence is present: since epistemic irrelevance is an asymmetric notion of independence, graphs that are Markov-equivalent in the Bayesian network sense can lead to different assessments of epistemic irrelevance.

The intuitive philosophical difference between Markov-equivalent graphs that is perceived by users of Bayesian networks is often associated with causality. As Shafer~\cite{Shafer:1997uk} put it: ``we need a way to give mathematical and philosophical content to the differences between Markov-equivalent graphs, differences that are none the less real to practitioners for the fact that they are not expressed by conditional independence''. The theory of causal networks~\cite[Chapter~21]{Koller:2009ty} does exactly that: although it is based on the theory of Bayesian networks, it differentiates between Markov equivalent graphs, both mathematically and philosophically. However, as we have just seen, causal networks are not the only type of networks that are capable of doing so. They share these features with credal networks under epistemic irrelevance.

\section{Lower expectations and their interpretations}\label{sec:lowerexpectations}

From a practical point of view, the main object of interest is usually not the irrelevant natural extension $\mathcal{F}^{\mathrm{irr}}_G$---or the complete extension $\mathcal{F}^{\mathrm{com}}_G$---itself, but rather the corresponding bounds on some parameters of interest, such as probabilities and expected values. 
The most important such bounds are tight lower and upper bounds on conditional expectations $E(f(\var{S})\newmid B)
\coloneqq
\sum_{\zval{S}\in\values{S}}
f(\zval{S})P(\zval{S}\newmid B)$, where~$f$ belongs to the set $\gambles{S}$ of all real-valued functions on $\values{S}$, with $S$ a subset of $G$, and where $B\in\mathcal{P}_{\emptyset}(\values{G})$ is a non-empty event. The tightest lower bound on this conditional expectation is called the \emph{lower expectation} of $f$. For any non-empty set $\mathcal{F}_G$ of full conditional probability measures $P$ on $\values{G}$, it is defined by\footnote{\label{footnote:minimumforepistemic}If $\mathcal{F}_G$ is equal to the irrelevant natural extension $\mathcal{F}_G^\irr$, the infimum in this expression can be replaced by a minimum. This follows from the fact that (a) $\underline{E}_G^\irr(\cdot\vert\cdot)$ is a coherent conditional lower prevision whose dominating conditional linear previsions are in one-to-one correspondence with the full conditional probability measures in $\mathcal{F}_G^\irr$---see Reference~\cite[Chapter~5]{DeBock2015:thesis}---and (b)  for any element of its domain, the value of such a coherent conditional lower prevision is always attained by at least one of its dominating conditional linear previsions---see the (end of the) proof of~\cite[Theorem~2]{Williams:2007eu}.}
\begin{equation}\label{eq:lowerexpectation}
\underline{E}_G(f(\var{S})\newmid B)\coloneqq\inf\left\{E(f(\var{S})\newmid B)\colon P\in\mathcal{F}_G\right\}.
\vspace{3pt}
\end{equation}
The \emph{upper expectation} $\smash{\overline{E}_G(f(\var{G})\newmid B)}$ can be defined analogously, simply by replacing the infimum with a supremum. However, because $\smash{\overline{E}_G(f(\var{S})\newmid B)=-\underline{E}_G(-f(\var{S})\newmid B)}$, it suffices to focus on lower expectations. Lower and upper probabilities are defined similarly, but these too will not be our main focus, because they correspond to the special case where $f$ is the indicator $\ind{A}$ of an event $A\in\mathcal{P}(\values{G})$, defined by $\ind{A}(\xval{G})\coloneqq 1$ if $\xval{G}\in A$ and $\ind{A}(\xval{G})\coloneqq 0$ otherwise. Indeed, since $P(A\newmid B)$ is clearly equal to $E(\ind{A}(\var{G})\newmid B)$, it follows that the conditional lower probability $\underline{P}_G(A\newmid B)
\coloneqq
\inf\left\{P(A\newmid B)\colon P\in\mathcal{F}_G\right\}$ is equal to $\underline{E}_G(\ind{A}(\var{G})\newmid B)$, and similarly for the conditional upper probability $\smash{\overline{P}_G(A\newmid B)}$. Unconditional lower and upper expectations and probabilities are obtained by conditioning on the certain event $B=\values{G}$, in which case we drop this trivial conditioning event $\values{G}$ from the notation, for example by writing $\underline{E}_G(f(\var{G}))$ instead of $\underline{E}_G(f(\var{G})\newmid\values{G})$, or writing $\underline{P}_G(A)$ instead of $\underline{P}_G(A\newmid\values{G})$.

Since, as we have just seen, upper expectations and lower and upper probabilities can all be regarded as special cases of lower expectations, we can focus on the latter without loss of generality. Therefore, in the remainder of this paper, we will focus on studying and computing lower expectations of the form $\underline{E}_G(f(\var{S})\newmid B)$. Usually, the conditional event $B$ will be of the form $x_T$, with $T\subseteq G$ and $x_T\in\values{T}$. In that case, we have the following simple yet very useful property:
\begin{equation}\label{eq:plugin}
\underline{E}_G(f(\var{S})\newmid x_T)
=
\underline{E}_G(f(\var{S\setminus T},x_{S\cap T})\newmid x_T),
\end{equation}
which follows directly from the definition of $E_G(f(\var{S})\newmid x_T)$ and the fact that, for every $\zval{S}\in\values{S}$ such that $\zval{S\cap T}\neq\xval{S\cap T}$, $P(\zval{S}\newmid\xval{T})=0$ because of~\ref{def:cond_prob:measure}. If $x_T$ is not fixed, we write $\underline{E}_G(f(\var{S})\newmid X_T)$ and interpret this as a real-valued function on $\values{T}$, defined for all $\xval{T}\in\values{T}$ by Equation~\eqref{eq:plugin}.

All of the definitions, results and conventions that we have so far introduced in this section can be applied to any non-empty set $\mathcal{F}_G$ of full conditional probability measures on $\values{G}$. Hence, in particular, they can also be applied to the irrelevant natural extension $\mathcal{F}^{\mathrm{irr}}_G$ of a credal network, as well as to the complete extension $\mathcal{F}^{\mathrm{com}}_G$. In the former case, we will denote the corresponding lower expectations as $\underline{E}^{\mathrm{irr}}_G(f(\var{S})\newmid B)$, whereas in the latter, we will use $\underline{E}^{\mathrm{com}}_G(f(\var{S})\newmid B)$ instead.

In this context of credal networks, in addition to the global lower expectations that we have just introduced, it is natural to consider local lower expectations as well, which are the lower expectations that correspond to the local credal sets. For any $s\in G$ and $\xval{\parents{s}}\in\values{\parents{s}}$, this local lower expectation $\underline{E}_{s\vert\xval{\parents{s}}}$ is defined by\footnote{The minimum in this expression is well defined because $\credal_{s\vert\xval{\parents{s}}}$ is assumed to be closed.}
\vspace{-0pt}
\begin{equation}\label{eq:locallowerexpectation}
\underline{E}_{s\vert\xval{\parents{s}}}(f)
\coloneqq
\min
\left\{
\sum_{x\in\values{s}}
f(x)p(x)
\colon
p\in\credal_{s\vert\xval{\parents{s}}}	
\right\}
~~\text{ for all $f\in\gambles{s}$.}\vspace{7pt}
\end{equation}
The local upper expectation $\smash{\overline{E}_{s\vert\xval{\parents{s}}}}$ can be defined analogously, by replacing the minimum with a maximum. However, as before, it suffices to focus on lower expectations because $\smash{\overline{E}_{s\vert\xval{\parents{s}}}(f)=-\underline{E}_{s\vert\xval{\parents{s}}}(-f)}$. Similarly, local lower and upper probabilities also correspond to special cases: for all $A\subseteq\values{s}$, we have that $\underline{P}_{s\mid\xval{\parents{s}}}(A)\coloneqq\underline{E}_{s\mid\xval{\parents{s}}}(\ind{A})$ and $\overline{P}_{s\mid\xval{\parents{s}}}(A)\coloneqq\overline{E}_{s\mid\xval{\parents{s}}}(\ind{A})$. Computing local lower expectations------and therefore also local upper expectations and local lower and upper probabilities---can be done using either of the following two approaches, which, in practice, are only feasible if $\credal_{s\vert\xval{\parents{s}}}$ is finitely generated. The first approach consists in altering Equation~\eqref{eq:locallowerexpectation}, by replacing $\credal_{s\vert\xval{\parents{s}}}$ with its set of extreme points $\mathrm{Ext}\,\credal_{s\mid\xval{\parents{s}}}$. Since $\credal_{s\vert\xval{\parents{s}}}$ is a compact set, this replacement will not change the result of the computation. Furthermore, if $\credal_{s\vert\xval{\parents{s}}}$ is finitely generated, we are now left with the trivial problem of finding the minimum of a finite number of values. The second approach is to express the inclusion in Equation~\eqref{eq:locallowerexpectation} in terms of the constraints in Equation~\eqref{eq:originallocalconstraints}. If $\credal_{s\vert\xval{\parents{s}}}$ is finitely generated, this reduces the problem to a linear optimisation problem, which can then be solved by means of standard linear programming methods. 
In the remainder of this paper, we will assume that the local lower expectation operators $\smash{\underline{E}_{s\vert\xval{\parents{s}}}}$---and therefore also the corresponding upper expectations and lower and upper probabilities---can be evaluated efficiently. In practice, for finitely generated models, this will usually be the case.

The interesting question though is not how to compute these local lower expectations, but rather how to compute the global lower expecations $\underline{E}^{\mathrm{irr}}_G(f(\var{S})\newmid B)$ and $\underline{E}^{\mathrm{com}}_G(f(\var{S})\newmid B)$ that correspond to the irrelevant natural extension or complete extension of a credal network. For most inferences, that is, for most choices of $f(X_S)$ and $B$, $\underline{E}^{\mathrm{irr}}_G(f(\var{S})\newmid B)$ and $\underline{E}^{\mathrm{com}}_G(f(\var{S})\newmid B)$ will not be equal and---since $\mathcal{F}_G^{\mathrm{com}}$ is a subset of $\mathcal{F}_G^{\mathrm{irr}}$---the former will then be strictly lower than the latter. However, there is one important simple special case for which they do coincide, which is when $S=\{s\}$ and $B=(\xval{\parents{S}},\xval{T})$, with $T\subseteq\nonparnondes{s}$, $\xval{\parents{S}}\in\values{\parents{S}}$ and $\xval{T}\in\values{T}$. In that case, the two types of global lower expectations that we consider both reduce to their common local lower expectation, in the sense that
\begin{equation}\label{eq:theybothreducetolocallowerexp}
\underline{E}^{\mathrm{irr}}_G(f(X_s)\newmid\xval{\parents{s}},\xval{T})
=\underline{E}^{\mathrm{com}}_G(f(X_s)\newmid\xval{\parents{s}},\xval{T})
=\underline{E}_{s\vert\xval{\parents{s}}}(f)
~~\text{for all $f\in\gambles{s}.$}
\end{equation}
There are also other---less trivial---types of inferences for which $\underline{E}_G^\irr$ and $\underline{E}_G^\mathrm{com}$ coincide; see for example Examples~\ref{exmp:forwardinferenceinMCs},~\ref{exmp:twodisconnectedwithoutlinearprogramming} and~\ref{exmp:HMMforward}. However, as explained above, in general, $\underline{E}^{\mathrm{irr}}_G(f(\var{S})\newmid B)$ provides a lower bound on $\underline{E}^{\mathrm{com}}_G(f(\var{S})\newmid B)$ that is usually strict.

For credal networks under epistemic irrelevance, Equation~\eqref{eq:theybothreducetolocallowerexp} is particularly important from a philosophical point of view, because it implies the following two equalities:
\begin{equation}\label{eq:assessmentsusinglowerexps}
\underline{E}^{\mathrm{irr}}_G(f(X_s)\newmid\xval{\nondes{s}})
=
\underline{E}^{\mathrm{irr}}_G(f(X_s)\newmid\xval{\parents{s}})
=\underline{E}_{s\vert\xval{\parents{s}}}(f)
~~\text{for all $f\in\gambles{s},$}
\end{equation}
which are obtained by applying Equation~\eqref{eq:theybothreducetolocallowerexp} twice, once for $T=\nonparnondes{S}$ and once for $T=\emptyset$. The first equality in Equation~\eqref{eq:assessmentsusinglowerexps} is an alternative formulation of the epistemic irrelevance assessments in Equation~\eqref{eq:irrelevance:assessment}, whereas the second equality can be regarded as a strengthened version of the local assessments that were imposed by Equation~\eqref{eq:local:assessments:full}, but now expressed in terms of lower expectations.

The philosophical importance of Equation~\eqref{eq:assessmentsusinglowerexps} is that it leads to an alternative characterisation for $\underline{E}_G^\irr$, because every conditional lower expectation operator $\underline{E}_G$ that satisfies Equation~\eqref{eq:assessmentsusinglowerexps} dominates $\underline{E}_G^\irr$, in the sense that
\begin{equation*}
\underline{E}_G(f(X_S)\vert B)\geq\underline{E}_G^\irr(f(X_S)\vert B)
~~\text{for all $S\subseteq G$, $f\in\gambles{S}$ and $B\in\mathcal{P}_{\emptyset}(\values{G})$.}\vspace{3pt}
\end{equation*}
Indeed, as an immediate consequence, it follows that $\underline{E}_G^\irr$ is the unique smallest---least committal, most conservative---lower expectation operator that satisfies Equation~\eqref{eq:assessmentsusinglowerexps}~\cite{DeBock2015:thesis}. In other words, if we impose that our lower expectation operator $\underline{E}_G$ should satisfy Equation~\eqref{eq:assessmentsusinglowerexps}, but make no further assumptions, then $\underline{E}_G$ should be equal to $\underline{E}_G^\irr$.  This result has profound philosophical implications, because it is expressed solely in terms of lower expectations, without any reference to probabilities. Therefore, this characterisation of $\underline{E}_G^\irr$---in contrast to our definition in terms of Equations~\eqref{eq:irrelevance:assessment} and~\eqref{eq:local:assessments:full}---broadens the scope of credal networks under epistemic irrelevance to other theories of uncertainty that also define a notion of lower expectation, but which do not necessarily interpret it in terms of probabilities.

An important example of such a theory of uncertainty is that of lower previsions~\cite{Troffaes:2014tl,Miranda:2008fo,Walley:1991vk}. Basically, a lower prevision is the same as a lower expectation. However, there is one imporant difference, which is that a lower prevision is not interpreted as a lower bound on some unknown expectation. In fact, it does not even assume the existence of some `true' expectation, nor does it assume the existence of probabilities. Instead, in this theory, the lower expectation---which they call lower prevision---of a function $f\in\gambles{G}$ is interpreted as the highest price that you are willing to pay to receive a lottery ticket that yields the---possibly negative---reward $f(x_G)$ if the uncertain variable $X_G$ ends up having the value $x_G$. For the special case of lower probabilities, this interpretation simplifies, and the lower probability of an event becomes the highest betting rate at which you would be willing to bet on that event. Given this difference in interpretation, Equation~\eqref{eq:lowerexpectation} no longer makes sense, and it is therefore replaced by a rationality criterion that is called coherence, which basically requires that it should not be possible for other people to combine multiple bets against you in such a way that they are guaranteed to win. Rather remarkably, this theory of lower previsions is mathematically equivalent to that of lower expectations. On the one hand, from a statistical point of view, the latter might seem simpler and more appealing. On the other hand, from the perspective of fields such as mathematical finance, lower previsions are more sensible, because of their direct interpretation in terms of gambling. 

In any case, the main message here is that Equation~\eqref{eq:assessmentsusinglowerexps} guarantees that $\underline{E}_G^\irr$ can be sensibly defined in terms of lower previsions, thereby extending the applicability of our results to other---say, financial---contexts, where lower expectations are no longer bounds on expectations, but are instead given a direct interpretation in terms of prices and gambling. This is not possible for the lower expectation $\underline{E}_G^{\mathrm{com}}$ of the complete extension, because its definition relies crucially on the assessements of stocastic independence in Equation~\eqref{eq:stochastic}, which---unlike epistemic irrelevance---cannot be expressed directly in terms of lower expectations, but only in terms of probabilities.\footnote{In all fairness, there are some very recent results that indicate that complete independence could be axiomatised in terms of choice functions, a framework for modelling uncertainty that falls beyond the scope of the present discussion~\cite[Section~4]{Cozman:2012fc}. We believe this to be promising. In cases where these kinds of axioms can be defended, they could lead to a justification for complete independence---and hence an interpretation for $\underline{E}_G^{\mathrm{com}}$---that does not rely on probabilities.} In that sense, credal networks under epistemic irrelevance have a broader scope.

In the rest of this paper, we present theoretical results and algorithmic tools that can be used to compute the global lower expectations $\underline{E}^{\mathrm{irr}}_G(f(\var{S})\newmid B)$ that correspond to a credal network under epistemic irrelevance. We will occasionally consider lower expectations of the form $\smash{\underline{E}^{\mathrm{com}}_G(f(\var{S})\newmid B)}$ as well, but only for the purposes of comparing them to $\smash{\underline{E}^{\mathrm{irr}}_G(f(\var{S})\newmid B)}$. All of our results for $\underline{E}_G^{\mathrm{irr}}$ are applicable regardless of the reader's preferred interpretation for lower expectations---such as lower bounds on expectations, supremum buying prices, etcetera. Nevertheless, for the purposes of this introductory paper, in order to stay closer to the typical interpretation of Bayesian networks, we will assume that it is possible to represent the uncertainty about $\var{G}$ by means of a single (full conditional) probability measure that may however not be known exactly, and will interpret lower expectations as lower bounds on the corresponding unkown expectations. In that context, as explained in Section~\ref{sec:irrelevance}, epistemic irrelevance can be regarded as a notion of `almost' stochastic independence, which does not require that $P(x_s\vert x_{\nondes{s}})$ and $P(x_s\vert x_{\parents{s}})$ should be equal, but instead restricts the distance between them by requiring that they belong to the same set.

\section{A Linear Programming Approach}\label{sec:linearprogramming}

The first and most straightforward method for computing $\underline{E}^{\mathrm{irr}}_G(f(\var{S})\newmid B)$ is to express it as the solution of a linear program, and to then solve this linear program by means of standard software tools. The advantage of this method is that it is always applicable. However, unfortunately, it is often intractable. Nevertheless, as we will see further on, this linear programming approach can still be an important algorithmic tool, provided that it is combined with other techniques.

For now, we focus on global unconditional lower expectations of the form $\underline{E}^{\mathrm{irr}}_G(f(\var{G}))$; extensions to conditional and more localized lower expectations will be discussed in Sections~\ref{sec:theory} and~\ref{sec:conditioningversusupdating}. In order to express such a global unconditional lower expectation $\underline{E}^{\mathrm{irr}}_G(f(\var{G}))$ as the solution to a linear program, the first step is to realize that
\begin{equation}\label{eq:minimisation}
\underline{E}^{\mathrm{irr}}_G(f(\var{G}))
\coloneqq
\inf\left\{E(f(\var{S}))\colon P\in\mathcal{F}_G^\irr\right\}
=\min\left\{\sum_{x_G\in\values{G}}f(\xval{G})P(x_G)\colon P(X_G)\in\mathcal{F}_G^\irr(X_G)\right\},
\end{equation}
where the use of the minimum is warranted by Footnote~\ref{footnote:minimumforepistemic}, and where $\mathcal{F}_G^\irr(X_G)$ consists of the global probability mass functions that correspond to $\mathcal{F}_G^\irr$---that is, the restrictions of the full conditional probability measures $P\in\mathcal{F}_G^\irr$ to events of the form $\xval{G}$. Because of this equation, computing $\underline{E}^{\mathrm{irr}}_G(f(\var{G}))$ is equivalent to minimizing a linear function of $P(\var{G})$, where $P(X_G)$ takes values in the set of probability mass functions $\mathcal{F}_G^\irr(X_G)$. Therefore, in order to express $\underline{E}^{\mathrm{irr}}_G(f(\var{G}))$ as the solution of a linear program, it suffices to characterise $\mathcal{F}_G^\irr(X_G)$ in terms of linear constraints.

In order to obtain such a characterisation for $\mathcal{F}_G^{\mathrm{irr}}(\var{\nodes})$, we start from similar characterisations for the local models $\credal_{s\mid\xval{\parents{s}}}$. In particular, for all $s\in\nodes$ and $\xval{\parents{s}}\in\values{\parents{s}}$, we describe the local credal set $\credal_{s\mid\xval{\parents{s}}}$ as the set of all real-valued functions $p$ on $\values{s}$ that satisfy the following constraints:
\begin{equation}\label{eq:localhomogeneous}
\sum_{\zval{s}\in\values{s}}p(\zval{s})=1
~~\text{and}~~
(\forall\gamma\in\Gamma(s,\xval{\parents{s}}))
~
\sum_{\zval{s}\in\values{s}}p(\zval{s})\gamma(\zval{s})\geq
0,
\end{equation}
where $\Gamma(s,\xval{\parents{s}})$ is a---possibly infinite---set of real-valued functions on $\values{s}$---a subset of $\gambles{s}$. Such a characterisation is always possible, because it can be derived from the constraint-based representation in Equation~\eqref{eq:originallocalconstraints}. First, $p\in\Sigma_{\values{s}}$ implies the equality in Equation~\eqref{eq:localhomogeneous}, which is the normalization condition of the probability mass function $p$. By multiplying this equality by $\beta$, and substracting it from the inequalities in Equation~\eqref{eq:originallocalconstraints}, we obtain homogeneous inequalities as in Equation~\eqref{eq:localhomogeneous}---with $\gamma=\alpha-\beta$. Finally, for every $x_s\in\values{s}$, $p\in\Sigma_{\values{s}}$ also requires that $p(x_s)$ should be non-negative. If this non-negativity constraint is not yet implied by the former inequalities, it should be added explicitly by choosing $\gamma=\ind{x_s}\coloneqq\ind{\{x_s\}}$. If $\credal_{s\mid\xval{\parents{s}}}$ is finitely generated, the total number of inequalities---that is, the size of $\Gamma(s,\xval{\parents{s}})$---will be finite.

The importance of these local representations in terms of homogeneous
linear constraints---regardless of whether $\Gamma(s,\xval{\parents{s}})$ is finite or not---is that we can use the local constraints to derive global ones, thereby obtaining the following characterisation of $\mathcal{F}_G^{\irr}(X_G)$.

\begin{proposition}\label{prop:linearprogram}
For all $s\in\nodes$
and \mbox{$\xval{\parents{s}}\in\values{\parents{s}}$}, let $\smash{\credal_{s\mid\xval{\parents{s}}}}$ be fully characterised
by means of Equation~\eqref{eq:localhomogeneous}.
Then $\mathcal{F}^{\mathrm{irr}}_G(\var{G})$ consists of those $P(\var{\nodes})\in\simplex{\nodes}$ for which, for all
  $s\in\nodes$,
 $\xval{\nondes{s}}\in\values{\nondes{s}}$
  and $\gamma\in\Gamma(s,\xval{\parents{s}})$:
\begin{equation}\label{eq:prop:linearprogram}
\sum_{\zval{s}\in\values{s}}\sum_{\zval{\descendants{s}}\in\values{\descendants{s}}}
P(\zval{s},\zval{\descendants{s}},\xval{\nondes{s}})\gamma(\zval{s})\geq0.
\end{equation}
\end{proposition}

When all lower probabilities are strictly positive, this result is
fairly straightforward. The global inequalities can then be obtained
by imposing all irrelevancies through element-wise Bayes's rule and clearing the denominators, as follows. For all
  $s\in\nodes$,
 $\xval{\nondes{s}}\in\values{\nondes{s}}$
  and $\gamma\in\Gamma(s,\xval{\parents{s}})$, we infer from Equations~\eqref{eq:epistemic:explicit} and~\eqref{eq:localhomogeneous} that
\begin{align*}
\sum_{\zval{s}\in\values{s}}P(\zval{s}\newmid\xval{\nondes{s}})\gamma(\zval{s})\geq
0
\Leftrightarrow
\sum_{\zval{s}\in\values{s}}
\frac{P(\zval{s},\xval{\nondes{s}})}{P(\xval{\nondes{s}})}\gamma(\zval{s})\geq
0
&\Leftrightarrow
\sum_{\zval{s}\in\values{s}}
P(\zval{s},\xval{\nondes{s}})\gamma(\zval{s})\geq
0,
\end{align*}
where the first equivalence follows from~\ref{def:cond_prob:Bayes} and the assumption that $P(\xval{\nondes{s}})$ is strictly positive. Because of~\ref{def:cond_prob:measure}, the last inequality in this series of equivalences is clearly identical to Equation~\eqref{eq:prop:linearprogram}. Therefore, this line of reasoning leads to a proof of Proposition~\ref{prop:linearprogram}, under an additional assumption of strict positivity. This argument is well known, and was first presented in Ref.~\cite[Section 8.3]{Cozman:2000ug}. The importance of our present result
is that it shows that these inequalities also remain valid---and still fully characterise $\mathcal{F}^{\mathrm{irr}}_G(\var{G})$---if we drop the assumption that $P(\xval{\nondes{s}})$ should be strictly positive.

Ref.~\cite{Cozman:2000ug} does not explicitly impose
$P(\var{\nodes})\in\simplex{\nodes}$ as a constraint. It seems to
assume that it suffices to impose only the unitary constraint
$\sum_{\zval{\nodes}\in\values{\nodes}}P(\zval{\nodes})=1$, and takes the non-negativity of $P(\zval{\nodes})$ for granted. Although we agree with this
statement, we do not believe it to be trivial. The following result makes this explicit.

\begin{theorem}\label{theo:linearprogramenkelunitary}
For all $s\in\nodes$
and \mbox{$\xval{\parents{s}}\in\values{\parents{s}}$}, let $\smash{\credal_{s\mid\xval{\parents{s}}}}$ be fully characterised
by means of Equation~\eqref{eq:localhomogeneous}.
Then $\mathcal{F}^{\mathrm{irr}}_G(\var{G})$ consists of those real-valued functions \mbox{$P(\var{\nodes})\in\reals^{\values{\nodes}}$} for which
\mbox{$\sum_{\zval{\nodes}\in\values{\nodes}}P(\zval{\nodes})=1$} and, for all
  $s\in\nodes$,
 $\xval{\nondes{s}}\in\values{\nondes{s}}$
  and \mbox{$\gamma\in\Gamma(s,\xval{\parents{s}})$}:
\begin{equation*}
\sum_{\zval{s}\in\values{s}}\sum_{\zval{\descendants{s}}\in\values{\descendants{s}}}P(\zval{s},\zval{\descendants{s}},\xval{\nondes{s}})\gamma(\zval{s})\geq0.
\end{equation*}
\end{theorem}

Proposition~\ref{prop:linearprogram} and Theorem~\ref{theo:linearprogramenkelunitary} are valid regardless of the cardinality of $\Gamma(s,\xval{\parents{s}})$, but if the cardinality is infinite, then the value of these results is mainly of a theoretical nature. They can only be used in practice---at least in an exact way---if $\Gamma(s,\xval{\parents{s}})$ is finite for all $s\in\nodes$ and \mbox{$\xval{\parents{s}}\in\values{\parents{s}}$}, or
equivalently, if all local credal sets are finitely generated.\footnote{If we allow for non-linear constraints, then local credal sets that are not finitely generated could be practical as well, provided that they can be described by means of a finite set of non-linear constraints. We believe that Proposition~\ref{prop:linearprogram} and Theorem~\ref{theo:linearprogramenkelunitary} could easily be adapted to allow for such non-linear (homogeneous) constraints, thereby expanding their practical use when combined with non-linear solvers.}
 In that case, Equation~\eqref{eq:minimisation} and Theorem~\ref{theo:linearprogramenkelunitary} provide global linear programs with a finite number of constraints, which can then be solved by standard linear programming software. Alternatively, in cases where $\underline{E}^{\mathrm{irr}}_G(f(\var{G}))$ needs to be evaluated for many different functions $f\in\gambles{G}$, it may also be beneficial to first compute the extreme points of $\mathcal{F}_G^\irr$ by applying a vertex enumeration program to the linear constraints that are provided by Theorem~\ref{theo:linearprogramenkelunitary}; in that case, $\underline{E}^{\mathrm{irr}}_G(f(\var{G}))$ is easily computed as the minimum of $E(f(\var{G}))$ over these extreme points.

 Unfortunately, none of these approaches is tractable for large credal networks, because the size of the obtained linear programs (as well as the corresponding number of extreme points) is exponential in the number of variables that define the network. Nevertheless, these methods do allow for inference problems in small networks to be solved exactly.

\begin{example}\label{exmp:twobinarylinearprogram}
In order to illustrate Theorem~\ref{theo:linearprogramenkelunitary}, we apply it to a very simple network that consists of two
unconnected binary variables $\var{1}$ and $\var{2}$---see Figure~\ref{fig:twodisconnectednodes}---where, for all
$\element\in\{1,2\}=G$, the variable $\varel$ assumes values
in a binary state space $\valuesel=\{\headsel,\tailsel\}$.
As explained in Reference~\cite{DeBock:2013wf}, $\mathcal{F}^{\mathrm{irr}}_G(\var{G})$ is then the so-called independent natural extension~\cite{deCooman:2011ey} of $\credal_1$ and $\credal_2$. Our goal here is to characterise this independent natural extension by means of linear constraints.

We start by constructing a suitable characterisation for the local models, in the form of Equation~\eqref{eq:localhomogeneous}. As explained in Example~\ref{exmp:extensionoftwodisconnectednodes}, for each of the binary
variables $\varel$---with $\element\in\{1,2\}$---the local credal set $\credal_s$
is uniquely characterised by the lower
 and upper probability of $\headsel$, in the sense that $p\in\credal_s$ if and only if $p\in\simplex{s}$ and $p(h_s)\in\big[\,\underline{p}(h_s),\overline{p}(h_s)\big]$. This immediately implies that $\credal_s$ is the set of all real-valued functions $p$ on $\values{s}$ that satisfy the following five constraints:
 \begin{equation*}
p(h_s)+p(t_s)=1,~
p(h_s)\geq0,~
p(t_s)\geq0,~
p(h_s)\geq\underline{p}(h_s)
\text{~and~}
p(h_s)\leq\overline{p}(h_s),
 \end{equation*}
 of which the first three constraints impose that $p\in\simplex{s}$ and the last two that $p(h_s)\in\big[\,\underline{p}(h_s),\overline{p}(h_s)\big]$. Of these five constraints, since $\underline{p}(h_s)\geq0$, the second constraint is clearly implied by the fourth. Similarly, since $\overline{p}(h_s)\leq 1$, the third constraint is implied by the first and last constraint. Hence, the second and third constraint are redundant, which implies that $\credal_s$ is the set of all real-valued functions $p$ on $\values{s}$ such that
 \begin{equation*}
p(h_s)+p(t_s)=1,~
p(h_s)\geq\underline{p}(h_s)
\text{~and~}
p(h_s)\leq\overline{p}(h_s).
 \end{equation*}
In order to obtain a representation in the form of Equation~\eqref{eq:localhomogeneous}, we need to make the two remaining inequalities homogeneous. This can be achieved by multiplying the unitary constraint by a suitable constant and substracting it from the inequalities. For example, for the first inequality, $p(h_s)+p(t_s)=1$ implies that
\begin{equation*}
p(h_s)\geq\underline{p}(h_s)
\Leftrightarrow
p(h_s)-\underline{p}(h_s)\big[p(h_s)+p(t_s)\big]
\geq\underline{p}(h_s)-\underline{p}(h_s)
\Leftrightarrow
\big[1-\underline{p}(h_s)\big]p(h_s)-\underline{p}(h_s)p(t_s)
\geq0,
\end{equation*}
which, since we know from Example~\ref{exmp:extensionoftwodisconnectednodes} that $\overline{p}(t_s)=1-\underline{p}(h_s)$, is equivalent to $\overline{p}(t_s)p(h_s)-\underline{p}(h_s)p(t_s)
\geq0$. A similar argument leads to a homogeneous version of $p(h_s)\leq\overline{p}(h_s)$, and we then find that
$\credal_s$ is the set of all real-valued functions $p$ on $\values{s}$ such that
 \begin{equation*}
p(h_s)+p(t_s)=1,~
\tailsupel\mass(\headsel)-\headslowel\mass(\tailsel)\geq0
\text{~and~}
-\tailslowel\mass(\headsel)+\headsupel\mass(\tailsel)\geq0.
 \end{equation*}
 This characterisation is clearly of the form in Equation~\eqref{eq:localhomogeneous}. Therefore, we can now apply Theorem~\ref{theo:linearprogramenkelunitary} to find that in this particular binary case, $\mathcal{F}^{\mathrm{irr}}_{G}(X_G)=\mathcal{F}^{\mathrm{irr}}_{\{1,2\}}(X_1,X_2)$ is the largest set of real-valued functions $P(X_G)=P(X_1,X_2)$ on $\values{G}=\values{1}\times\values{2}$ such that
\begin{align}
\tailsupone P(\headsone,\headstwo)-\headslowone P(\tailsone,\headstwo)&\geq0,~~
-\tailslowone P(\headsone,\headstwo)+\headsupone P(\tailsone,\headstwo)\geq0,\label{eq:constraints1}\\
\tailsupone P(\headsone,\tailstwo)-\headslowone P(\tailsone,\tailstwo) &\geq0,~~
-\tailslowone P(\headsone,\tailstwo)+\headsupone P(\tailsone,\tailstwo)\geq0,\label{eq:constraints2}\\
\tailsuptwo P(\headsone,\headstwo)-\headslowtwo P(\headsone,\tailstwo) &\geq0,~~
-\tailslowtwo P(\headsone,\headstwo)+\headsuptwo P(\headsone,\tailstwo)\geq0,\label{eq:constraints3}\\
\tailsuptwo P(\tailsone,\headstwo)-\headslowtwo P(\tailsone,\tailstwo) &\geq0,~~
-\tailslowtwo\mass(\tailsone,\headstwo)+\headsuptwo\mass(\tailsone,\tailstwo)\geq0\label{eq:constraints4}
\end{align}
and
\begin{equation}\label{eq:unitaryinexample}
\mass(\headsone,\headstwo)+
\mass(\headsone,\tailstwo)+
\mass(\tailsone,\headstwo)+
\mass(\tailsone,\tailstwo)=1.\vspace{6pt}
\end{equation}
Proposition~\ref{prop:linearprogram} leads to a similar characterisation. However, it additionally imposes that $P(X_1,X_2)$ should be non-negative, in the sense that $P(x_1,x_2)\geq0$ for all $x_1\in\{h_1,t_1\}$ and $x_2\in\{h_2,t_2\}$. Theorem~\ref{theo:linearprogramenkelunitary} guarantees that these non-negativity constraints are redundant.

For any $f\in\gambles{G}$, because of Equation~\eqref{eq:minimisation}, computing $\underline{E}^{\mathrm{irr}}_G(f(\var{G}))$ is now a matter of minimising the linear function
\begin{equation}\label{eq:linearfunctionexample}
f(h_1,h_2)P(h_1,h_2)+
f(h_1,t_2)P(h_1,t_2)+
f(t_1,h_2)P(t_1,h_2)+
f(t_1,t_2)P(t_1,t_2)
\end{equation}
with respect to the linear constraints in Equations~\eqref{eq:constraints1}--\eqref{eq:unitaryinexample}, which is easily solved using standard linear programming software.
For example, say that we are interested in the lower probability $\underline{P}_G^\irr(A)$ of the event $A$ that $X_1$ and $X_2$ are equal, in the sense that $(X_1,X_2)=(h_1,h_2)$ or $(X_1,X_2)=(t_1,t_2)$. Then as explained in Section~\eqref{sec:lowerexpectations}, this lower probability is equal to the lower expectation $\underline{E}_G^\irr(f(X_G))$ of the function $f=\ind{A}$, with $\ind{A}$ the indicator of $A$. Hence, in this case we have that $f(h_1,h_2)=f(t_1,t_2)=1$ and $f(h_1,t_2)=f(t_1,h_2)=0$. If the local credal sets are specified as in Example~\ref{exmp:extensionoftwodisconnectednodes}---that is, if we let $\underline{p}(x_i)=\nicefrac{1}{4}$ and $\overline{p}(x_i)=\nicefrac{3}{4}$---then solving the linear program above yields $\smash{\underline{P}_G^\irr(A)=\underline{E}_G^\irr(f(X_G)=\nicefrac{1}{4}}$.

Alternatively, if we want to compute $\underline{E}_G^\irr(f(X_G)$ for many different functions $f\in\gambles{G}$, then it can be worthwhile to first compute the extreme points of $\smash{\mathcal{F}^{\mathrm{irr}}_{\{1,2\}}(X_1,X_2)}$, which are the vertices of the polytope that is described by Equations~\eqref{eq:constraints1}--\eqref{eq:unitaryinexample}. Computing the value of $\smash{\underline{E}_{\{1,2\}}^\irr(f(X_1,X_2))}$ is then a matter of minimising Equation~\eqref{eq:linearfunctionexample} over this finite set of extreme points. In general, computing the extreme points of $\smash{\mathcal{F}^{\mathrm{irr}}_{G}(X_G)}$ requires running a vertex enumeration program. However, in this very particular binary case, analytical expressions for these extreme points are available~\cite{DeBock:2013wf}. 
If the local credal sets are chosen as in Example~\ref{exmp:extensionoftwodisconnectednodes}, it follows from these expressions that $\smash{\mathcal{F}^{\mathrm{irr}}_{\{1,2\}}(X_1,X_2)}$ has six extreme points, one of which is given by Equation~\eqref{eq4:exmp:extensionoftwodisconnectednodes}. This particular extreme point is also the one that, for $f=\ind{A}$ with $A$ as above, achieves the minimal value of Equation~\eqref{eq:linearfunctionexample}. Again, we find that $\smash{\underline{P}_G^\irr(A)=\underline{E}_G^\irr(f(X_G)=\nicefrac{1}{4}}$.
\hfill$\lozenge$
\end{example}

\section{Decomposition Properties: Beyond the Special Case of Trees}\label{sec:theory}

For about ten years, starting with its introduction in Reference~\cite{Cozman:2000ug}, the linear programming approach that was outlined in the previous section was basically the only available method for computing inferences in credal networks under epistemic irrelevance. Since this approach is feasible only for very small networks, this has severly inhibited---if not made impossible---the application of credal networks under epistemic irrelevance to real problems. Credal networks under complete---or strong---independence, in contrast, did lead to such applications~\cite{Antonucci:2009jm,Antonucci:2007wv}. Therefore, it has long been thought that credal networks under epistemic irrelevance are impractical to work with, at least compared to other types of credal networks.

As a consequence of some recent promising new developments, this perception is now starting to change. A prime example of such developments is the design of algorithms that are expressed directly in terms of lower previsions, which---as explained in Section~\ref{sec:lowerexpectations}---are basically just lower expectations, but with a different interpretation attached to them. For credal networks under epistemic irrelevance of which the graph is a tree, there is now a polynomial-time updating algorithm that can compute $\underline{E}_G^\irr(f(\var{s})\newmid\xval{T})$ for $s\in G$ and $T\subseteq G\setminus\{s\}$ \citep{deCooman:2010gd}. 
This is rather remarkable, especially since the same inference problem is NP-hard for credal networks under strong (and complete) independence~\citep{Maua:2014ti}. Other recent algorithmic developments considered the case of imprecise hidden Markov models under epistemic irrelevance, the graph of which is again a---special type of---tree~\citep{Benavoli:2011ei,DeBock:2014ts}.

It is no coincidence that all these recent algorithmic successes with credal networks under epistemic irrelevance have been obtained for networks whose graph is a tree. Essentially, all of these algorithms are based on the fact that for trees, the irrelevant natural extension satisfies a number of convenient theoretical properties, which allow for large computational problems to be decomposed into smaller ones and, as such, enable the development of efficient recursive algorithms; see for example~\citep{deCooman:2010gd}. 

In order to develop efficient algorithms for networks that are more general than trees, an important first step is therefore to generalise these theoretical properties from trees to arbitrary directed acyclic graphs. Recently, it has been shown that this is indeed possible~\citep{DeBock:2013tt,DeBock:2014bv,DeBock2015:thesis}. However, these generalised properties have been obtained and stated using sets of desirable gambles, and are therefore rather inaccessible to the general Bayesian network community. In order to remedy this situation, we here present them in a more accessible format, in terms of lower expectations.

\subsection{Sub-networks and their Connection to the Original Network}\label{sec:subnetworks}

\begin{figure}
\begin{center}
\begin{tikzpicture}[node distance=15pt and 25pt
,>=triangle 45
,decoration={
      markings,
      mark=at position 1 with {\arrow[scale=1.5]{to}};
    }
]
\tikzstyle{punt} = [circle, draw, inner sep=0pt, minimum size=15pt]
\tikzstyle{punt2} = [circle, draw, line width=1.3pt, inner sep=0pt, minimum size=15pt]
\tikzstyle{punt4} = [circle, draw, inner sep=0pt, minimum size=15pt, fill=black!20]
\tikzstyle{stip}=[circle, fill, inner sep=0pt, minimum size=4pt]

\node[punt] (s1) {$1$};

\node[punt] (s3) [above right = of s1] {$3$};
\node[punt] (s2) [above left = of s3] {$2$};
\node[punt] (s4) [below right = of s3] {$4$};
\node[punt2] (s5) [above right = of s3] {$5$};
\node[punt2] (s7) [below right = of s5] {$7$};
\node[punt] (s8) [above right = of s5] {$8$};
\node[punt] (s6) [above left = of s8] {$6$};
\node[punt2] (s9) [below right = of s7] {$9$};
\node[punt] (s10) [above right = of s7] {$10$};

\node (G) [right = -7.9cm of s6,yshift=-1.35cm] {
\begin{minipage}{2cm}
\begin{align*}
G&=\{1,2,3,4,5,6,7,8,9,10\}\\[20pt]
K&=\{5,7,9\}\\
\parents{K}&=\{3,4\}\\
\descendants{K}&=\{8,{10}\}\\
\nondes{K}&=\{1,2,3,4,6\}\\
\nonparnondes{K}&=\{1,2,6\}
\end{align*}
\end{minipage}
};

\draw[postaction={decorate}] (s1) -- (s3);
\draw[postaction={decorate}] (s2) -- (s3);
\draw[postaction={decorate}] (s3) -- (s4);
\draw[postaction={decorate}] (s3) -- (s5);
\draw[postaction={decorate}] (s6) -- (s8);
\draw[postaction={decorate}] (s5) -- (s8);
\draw[postaction={decorate}] (s4) -- (s7);
\draw[postaction={decorate}] (s7) -- (s10);

\draw[postaction={decorate}, line width=1.2pt] (s5) -- (s7);
\draw[postaction={decorate}, line width=1.2pt] (s7) -- (s9);

\end{tikzpicture}
\vspace{15pt}
\caption{A sub-DAG of the DAG in Figure~\ref{fig:DAG}.}
\label{fig:DAGclosed}
\end{center}
\end{figure}
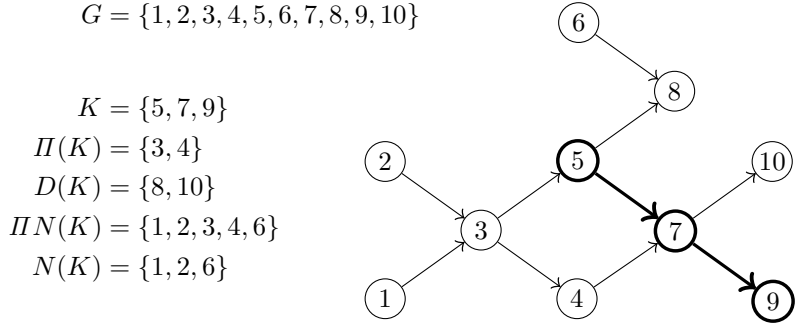

A first important property is Theorem~\ref{theo:subnetwork} further on, which relates the lower expectations of a credal network under epistemic irrelevance to the lower expectations of its so-called sub-networks, thereby allowing for a reduction of the size of the network. Loosely speaking, a sub-network is just a part of the original network, obtained by removing some of its nodes. A formal explanation requires some additional graph-theoretic concepts, which are illustrated in Figure~\ref{fig:DAGclosed}.

For any subset $K$ of $G$, we define its set of parents as $\parents{K}\coloneqq(\bigcup_{s\in K}\parents{s})\setminus K$ and its set of descendants as $\descendants{K}\coloneqq(\bigcup_{s\in K}\descendants{s})\setminus K$. The set of non-descendants of $K$ is given by $\nondes{K}\coloneqq\nodes\setminus(K\cup\descendants{K})$, and we also define the set $\nonparnondes{K}\coloneqq\nondes{K}\setminus\parents{K}$. If $K$ is a singleton $\{s\}$, these concepts reduce to the simple versions in Section~\ref{sec:credal}. Finally, we call a subset $K$ of $G$ \emph{closed} if, for all $s,t\in K$ and $k\in\nodes$, $s\precedes k\precedes t$ implies that $k\in K$.

With any subset $K$ of $G$ and any fixed value $\xval{\parents{K}}$ of $\var{\parents{K}}$, we now associate a new credal network, called a \emph{sub-network}. The graph of this sub-network depends only on $K$, and is obtained from the original graph by simply removing the nodes that do not belong to $K$ and the arrows that enter these nodes or depart from these nodes. For example, in Figure~\ref{fig:DAGclosed}, the sub-graph that corresponds to $K\coloneqq\{5,7,9\}$ is highlighted by means of thicker lines. The local credal sets of the sub-network are equal to the original ones. However, this is not immediate: a node $s\in K$ might have a parent $t$ that does not belong to $K$. For example, in Figure~\ref{fig:DAGclosed}, the node $7\in K$ has a parent $4$ that does not belong to $K$. For this reason, in order to obtain local models that only depend on parents that belong to the nodes $K$ of the sub-network, we fix the value $x_t$ of $X_t$ for every $t\in G\setminus K$ that has a child in $K$, or equivalently, we fix the value $\xval{\parents{K}}$ of $\var{\parents{K}}$. In this way, the local models of the sub-network depend on the chosen fixed value $\xval{\parents{K}}$ of $\var{\parents{K}}$. For example, in the sub-network of Figure~\ref{fig:DAGclosed}, the node $7$ has only one parent, which is node $5$, and the local credal sets for $7$ are therefore conditional on the value $z_{5}$ of $X_5$. For every such $z_{5}
\in\values{5}$, this local credal set is given by $\credal_{7\mid z_5}\coloneqq\credal_{7\mid(z_5,x_4)}$, where $\credal_{7\mid(z_5,x_4)}$ is the local credal set of the original network, with $x_4$ the value of $X_4$ that corresponds to the fixed choice of $\xval{\parents{K}}$.

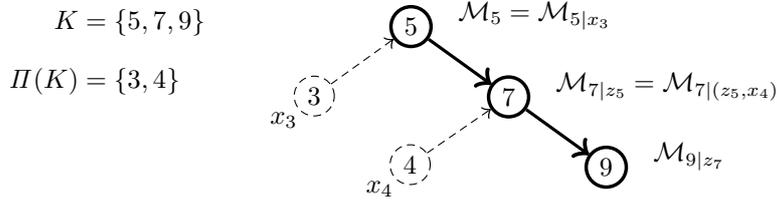
\begin{figure}
\begin{center}
\vspace{5pt}
\begin{tikzpicture}[node distance=15pt and 25pt
,>=triangle 45
,decoration={
      markings,
      mark=at position 1 with {\arrow[scale=1.5]{to}};
    }
]
\tikzstyle{punt} = [circle, draw, inner sep=0pt, minimum size=15pt]
\tikzstyle{punt2} = [circle, draw, line width=1.3pt, inner sep=0pt, minimum size=15pt]
\tikzstyle{punt3} = [circle, draw, densely dashed, inner sep=0pt, minimum size=15pt]
\tikzstyle{punt4} = [circle, draw, inner sep=0pt, minimum size=15pt, fill=black!20]
\tikzstyle{stip}=[circle, fill, inner sep=0pt, minimum size=4pt]

\node[punt3] (s3) {$3$};

\node[punt3] (s4) [below right = of s3] {$4$};
\node[punt2] (s5) [above right = of s3] {$5$};
\node[punt2] (s7) [below right = of s5] {$7$};
\node[punt2] (s9) [below right = of s7] {$9$};

\node (x3) [below=0pt of s3,yshift=5pt,xshift=-13pt] {$~\xval{3}$};

\node (x4) [below=0pt of s4,yshift=5pt,xshift=-13pt] {$~\xval{4}$};

\node (D5) [right=0pt of s5,xshift=6pt,yshift=4pt] {$\credal_{5}=\credal_{5\mid\xval{3}}$};

\node (D7) [right=0pt of s7,xshift=6pt,yshift=4pt] {$\credal_{7\mid\zval{5}}=\credal_{7\mid(\zval{5},\xval{4})}$};

\node (D9) [right=0pt of s9,xshift=6pt,yshift=4pt] {$\credal_{9\mid\zval{7}}$};

\node (G) [right = -5.7cm of s5,yshift=-0.1cm] {
\begin{minipage}{2cm}
\begin{align*}
K&=\{5,7,9\}\\[7pt]
\parents{K}&=\{3,4\}
\end{align*}
\end{minipage}
};

\draw[postaction={decorate},densely dashed] (s3) -- (s5);
\draw[postaction={decorate},densely dashed] (s4) -- (s7);

\draw[postaction={decorate}, line width=1.2pt] (s5) -- (s7);
\draw[postaction={decorate}, line width=1.2pt] (s7) -- (s9);

\end{tikzpicture}
\vspace{7pt}
\caption{A sub-network and its local models}
\label{fig:sub-DAG}
\end{center}
\end{figure}

For any choice of $K\subseteq G$ and $\xval{\parents{K}}$, the corresponding sub-network, like any credal network, has an irrelevant natural extension. We will denote this irrelevant natural extension by $\smash{\mathcal{F}_{K\vert\xval{\parents{K}}}^\irr}$ and will use $\smash{\underline{E}_{K\vert\xval{\parents{K}}}^\irr}$ to refer to the corresponding lower expectations. If $K$ is closed, these lower expectations satisfy the following important property.

\begin{theorem}[marginalisation]\label{theo:subnetwork}
Consider any closed subset $K$ of $G$ and any $\xval{\parents{K}}\in\values{\parents{K}}$. Then
\begin{align*}
\underline{E}_G^\irr\big(f(\var{K})\newmid B_K,\xval{\parents{K}},B_{\nonparnondes{K}}\big)
=
\underline{E}_{K\vert\xval{\parents{K}}}^\irr\big(f(\var{K})\newmid B_K\big)
=
\underline{E}_G^\irr\big(f(\var{K})\newmid B_K,\xval{\parents{K}}\big)
\end{align*}
for all $f\in\gambles{K}$, $B_K\in\mathcal{P}_\emptyset(\values{K})$ and $B_{\nonparnondes{K}}\in\mathcal{P}_\emptyset(\values{\nonparnondes{K}})$.
\end{theorem}

The crucial feature of this result is that it allows us to reduce an optimisation problem in a credal network into a similar but smaller-sized optimisation problem in one of its sub-networks. If $K$ is a singleton $\{s\}$, with $s\in K$, this sub-network contains only this single node $s$. The unconditional part of $\smash{\underline{E}^\irr_{s\mid\xval{\parents{s}}}}$ is then trivially equal to the local lower expectation $\smash{\underline{E}_{s\mid\xval{\parents{s}}}}$, and therefore, for all $f\in\gambles{s}$,
\begin{equation}\label{eq:trivialsubnetwork}
\underline{E}^\irr_{s\mid\xval{\parents{s}}}(f(X_s))=\underline{E}_{s\mid\xval{\parents{s}}}(f(X_s))
\end{equation}
 can easily be computed using the methods in Section~\ref{sec:lowerexpectations}. In less trivial cases, the obtained reduced optimisation problem can be solved by means of the linear programming techniques of Section~\ref{sec:linearprogramming}---provided that the sub-network is sufficiently small---or by means of other  techniques that will be presented further on.

\begin{example}\label{exmp:unconditionalsubnetwork}
In order to illustrate Theorem~\ref{theo:subnetwork}, we apply it to a credal network under epistemic irrelevance of which the graph is depicted in Figure~\ref{fig:DAG}. In particular, we consider some real-valued function $h\in\gambles{9}$ of $X_9$ and are interested in computing $\underline{E}_G^\irr(h(X_9)\newmid x_3,x_4,x_6)$, which is the  lower expectation of $h$ conditional on the event that $X_3=x_3$, $X_4=x_4$ and $X_6=x_6$, for some $x_3\in\values{3}$, $x_4\in\values{4}$ and $x_6\in\values{6}$.

If we choose $K=\{5,7,9\}$, $B_K=\values{K}$ and $B_{N(K)}=(X_6=x_6)$, then because $\parents{K}=\{3,4\}$, it follows from Theorem~\ref{theo:subnetwork} that
\begin{equation*}
\underline{E}_G^\irr(h(X_9)\newmid x_3,x_4,x_6)
=
\underline{E}_{K\mid\xval{\parents{K}}}^\irr(h(X_9)).
\vspace{6pt}
\end{equation*}
This equation clearly simplifies the problem considerably. The original inference problem at the left hand side of the equation is a conditional lower expectation in a network with ten nodes, whereas the right hand side of the equation is an unconditional lower expectation in a network with three nodes, the graph of which is depicted in Figure~\ref{fig:DAGclosed}. Since the size of this sub-network is very small, $\smash{\underline{E}_{K\mid\xval{\parents{K}}}^\irr(h(X_9))}$---and therefore also $\smash{\underline{E}_G^\irr(h(X_9)\newmid x_3,x_4,x_6)}$---can be computed using the linear programming methods of Section~\ref{sec:linearprogramming}. Alternatively, since the graph of this sub-network is a chain, we can also use the---more efficient---recursive method that will be described further on in Example~\ref{exmp:forwardinferenceinMCs}.\hfill$\lozenge$
\end{example}

Readers that are familiar with the Bayesian network techniques of removing barren nodes and d-separated evidence might have recognised some of these techniques in the above example. On the one hand, since $X_6$ is d-separated from $X_9$ given $X_3$ and $X_4$, one might for example expect that the evidence $x_6$ can be removed. Similarly, since $X_8$ and $X_{10}$ correspond to barren nodes---nodes that are not involved in the inference and that either have no children either have only barren children---one might expect that these nodes can be removed from the network. Caution is adviced however in applying this kind of intuition. Since these Bayesian network techniques are based on the factorisation property in Equation~\eqref{eq:factorisation}, which credal networks under epistemic irrelevance do not satisfy, it is far from obvious that these techniques apply to credal networks under epistemic irrelevance. Nevertheless, in some cases, it turns out that this is indeed possible.

First of all, since the removal of barren nodes yields a subnetwork of which the set of nodes is clearly closed, it follows from Theorem~\ref{theo:subnetwork} that these barren nodes can indeed be removed without affecting the inference, as is the case in Bayesian networks. In fact, under strict positivity conditions, this result can already be found in Reference~\cite[Theorem~15 and 16]{Cozman:2000ug}. The situation is more complicated for d-separation, because it needs to be replaced with an  alternative---asymmetrical---separation condition, called AD-separation. In the example above, $X_6$ is in fact AD-separated from $X_9$ given $X_3$ and $X_4$, and the aforementioned intuition is therefore correct in this case; see Example~\ref{ex:ADseparation} further on. However, this does not apply generally. A detailed discussion of AD-separation and its connection with credal networks under epistemic irrelevance will be provided in the next section. For now, it suffices to forget about this intuition, and to simply apply Theorem~\ref{theo:subnetwork} instead.

\subsection{The Law of Iterated Lower Expectation}

Whenever Theorem~\ref{theo:subnetwork} applies, it offers a very powerful tool, because it can turn our original inference problem into a single smaller one. However, clearly, it cannot always apply, because the converse would mean that every complicated inference in a credal network under epistemic irrelevance can be reduced to a simple one. Unfortunately, that's too good to be true. In order to move to more complicated types of inference, while still remaining efficient, the trick is not to try and reduce our inference problem to a \emph{single} simpler problem, but rather to try and decompose it into \emph{multiple} simpler ones. In the remainder of this section, we introduce a number of properties that will allow us to do exactly that.

The following property is perhaps the most important one, as it has been---and, no doubt, will remain to be---the backbone of recursive algorithms.
\begin{theorem}[law of iterated lower expectation]\label{theo:marginalextension}
Consider any set $S\subseteq G$, with $T\coloneqq G\setminus S$, such that $t\sprecedes s$ for all $t\in T$ and $s\in S$. Then
\begin{equation*}
\underline{E}^{\irr}_G(f(\var{G}))
=
\underline{E}^{\irr}_{T}(\underline{E}^{\irr}_{S\mid\var{\parents{S}}}(f(\var{G})))
\text{~~for all $f\in\gambles{G}$,}
\end{equation*}
where $\underline{E}^{\irr}_{S\mid\var{\parents{S}}}(f(\var{G}))$ is a function of $X_T$ whose value in $x_T\in\values{T}$ is equal to $\underline{E}^{\irr}_{S\mid\xval{\parents{S}}}(f(\var{S},\xval{T}))$.
\end{theorem}
Basically, this result is just an imprecise version of the law of iterated expectation, or equivalently, the law of total probability. For readers that are acquainted with imprecise-probabilistic jargon: this result establishes marginal extension. The following three examples should help to clarify this perhaps rather abstract property. The first example is a very simple illustration.

\begin{figure}
\begin{center}
\vspace{15pt}
\begin{tikzpicture}[node distance=15pt and 25pt
,>=triangle 45
,decoration={
      markings,
      mark=at position 1 with {\arrow[scale=1.5]{to}};
    }
]
\tikzstyle{punt} = [circle, draw, inner sep=0pt, minimum size=15pt]
\tikzstyle{punt2} = [circle, draw, line width=1.3pt, inner sep=0pt, minimum size=15pt]
\tikzstyle{punt4} = [circle, draw, inner sep=0pt, minimum size=15pt, fill=black!20]
\tikzstyle{stip}=[circle, fill, inner sep=0pt, minimum size=4pt]

\node[punt] (s1) {$1$};

\node[punt] (s3) [above right = of s1] {$3$};
\node[punt] (s2) [above left = of s3] {$2$};

\node (G) [right = -6.9cm of s6,yshift=-2.55cm] {
\begin{minipage}{2cm}
\begin{align*}
G&=\{1,2,3\}
\end{align*}
\end{minipage}
};

\node (G2) [right = 0.1cm of s6,yshift=-2.6cm] {
\begin{minipage}{2cm}
\begin{align*}
S&=\{3\}\\
T&=\{1,2\}\\
\parents{S}&=\{1,2\}
\end{align*}
\end{minipage}
};

\draw[postaction={decorate}] (s1) -- (s3);
\draw[postaction={decorate}] (s2) -- (s3);

\end{tikzpicture}
\vspace{10pt}
\caption{A simple illustration of Theorem~\ref{theo:marginalextension}.}
\label{fig:DAGreversetree}
\end{center}
\end{figure}
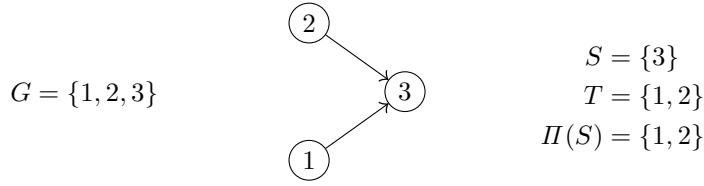

\begin{example}\label{exmp:twonodeswithcommonchild}
Let $G\coloneqq\{1,2,3\}$, consider a credal network of which the graph is depicted in Figure~\ref{fig:DAGreversetree} and let $f\in\gambles{G}$ be a function of $X_G=(X_1,X_2,X_3)$. The computation of the lower expectation $\underline{E}^\irr_{G}(f(X_G))$ can then be simplified as follows. Let $S\coloneqq\{3\}$ and $T\coloneqq G\setminus{S}=\{1,2\}$. Since $1\sprecedes3$ and $2\sprecedes3$, the conditions of Theorem~\ref{theo:marginalextension} are satisfied, and therefore, we find that
\vspace{3pt}
\begin{equation*}
\underline{E}^\irr_{G}(f(X_G))
=
\underline{E}^\irr_{\{1,2\}}(
\underline{E}^\irr_{3\mid(X_1,X_2)}(
f(X_1,X_2,X_3)
)
)
=
\underline{E}^\irr_{\{1,2\}}(
h(X_1,X_2)
),\vspace{-5pt}
\end{equation*}
with
\begin{equation*}
h(x_1,x_2)\coloneqq
\underline{E}^\irr_{3\mid(x_1,x_2)}(
f(x_1,x_2,X_3)
)
\text{~~for all $x_1\in\values{1}$ and $x_2\in\values{2}$.}
\vspace{7pt}
\end{equation*}
Computing the function $h$ is trivial because, for every $x_1\in\values{1}$ and $x_2\in\values{2}$, it follows from Equation~\eqref{eq:trivialsubnetwork} that computing $h(x_1,x_2)$ is a matter of evaluating the local lower expectation $\underline{E}_{3\mid(x_1,x_2)}$ for the univariate function $f(x_1,x_2,X_3)$, which is a local problem that can be solved by means of any of the methods that were discussed in Section~\ref{sec:lowerexpectations}. Instead of having to compute a lower expectation of a function $f$ that depends on three variables, in a credal network with three nodes, we are now left with the problem of computing  the lower expectation $\underline{E}^\irr_{\{1,2\}}(
h(X_1,X_2)$ of a function $h$ that depends on only two variables, in a credal network that consists of two disconnected nodes. This reduced problem can now be solved by means of the linear programming techniques in Section~\ref{sec:linearprogramming}. In the particular case where $X_1$ and $X_2$ are both binary, this procedure was explained in detail in Example~\ref{exmp:twobinarylinearprogram}.
\hfill$\lozenge$
\end{example}

Our second illustration of the law of iterated lower expectation is slightly more involved, and derives a general recursive inference method for imprecise Markov chains. 

\begin{figure}
\begin{center}
\begin{tikzpicture}[node distance=15pt and 15pt
,>=triangle 45
,decoration={
      markings,
      mark=at position 1 with {\arrow[scale=1.5]{to}};
    }
]
\tikzstyle{punt} = [circle, draw, inner sep=0pt, minimum size=15pt]
\tikzstyle{punt2} = [circle, draw, line width=1.3pt, inner sep=0pt, minimum size=15pt]
\tikzstyle{stip}=[circle, fill, inner sep=0pt, minimum size=4pt]

\node[punt] (s1) {$1$};

\node[punt] (s2) [right = of s1] {$2$};
\node[punt] (s3) [right = of s2] {$3$};
\node[punt] (s4) [right = of s3] {$4$};
\node[punt] (s5) [right = of s4] {$5$};
\node[punt] (s6) [right = of s5] {$6$};
\node[punt] (s7) [right = of s6] {$7$};
\node[punt] (s8) [right = of s7] {$8$};
\node[punt] (s9) [right = of s8] {$9$};
\node[punt] (s10) [right = of s9] {$10$};

\node (G) [right = -2.2cm of s1,yshift=-0.8cm] {
\begin{minipage}{2cm}
\begin{align*}
G=\{1,2,3,4,5,6,7,8,9,10\}
~~~~~
S=\{10\}
~~~~~
T=\{1,2,3,4,5,6,7,8,9\}
~~~~~
\parents{S}=\{9\}
\end{align*}
\end{minipage}
};

\draw[postaction={decorate}] (s1) -- (s2);
\draw[postaction={decorate}] (s2) -- (s3);
\draw[postaction={decorate}] (s3) -- (s4);
\draw[postaction={decorate}] (s4) -- (s5);
\draw[postaction={decorate}] (s5) -- (s6);
\draw[postaction={decorate}] (s6) -- (s7);
\draw[postaction={decorate}] (s7) -- (s8);
\draw[postaction={decorate}] (s8) -- (s9);
\draw[postaction={decorate}] (s9) -- (s10);
\end{tikzpicture}
\vspace{4pt}
\caption{The DAG of an imprecise Markov chain with $10$ nodes. 
}\label{fig:genericchain}
\end{center}
\end{figure}

\begin{example}\label{exmp:forwardinferenceinMCs}
Consider a credal network whose graph is a chain with ten nodes---see Figure~\ref{fig:genericchain}---and let $f\in\gambles{G}$ be a function of $X_G=(X_1,X_2,\dots,X_{10})$. We are interested in computing $\underline{E}^\irr_{G}(f(X_G))$. As we are about to show, Theorem~\ref{theo:marginalextension} enables us to do this recursively. In particular, if we choose $S=\{10\}$ and $T=\{1,\dots,9\}$, then
\vspace{2pt}
\begin{equation}\label{eq:MarkovGeneral}
\underline{E}^\irr_{G}(f(X_G))
=\underline{E}^\irr_T(
\underline{E}^\irr_{10\mid X_9}(
f(X_G)
)
)
=\underline{E}^\irr_T(
f_9(X_T)
),
\end{equation}
with
\begin{equation*}
f_9(x_T)\coloneqq
\underline{E}^\irr_{10\mid x_9}(
f(x_T,X_{10})
)
=
\underline{E}_{10\mid x_9}(
f(x_T,X_{10})
)
\text{~~for all $x_T\in\values{T}$,}
\vspace{7pt}
\end{equation*}
using Equation~\eqref{eq:trivialsubnetwork} for the last equality. In this way, the original problem---the computation of $\underline{E}^\irr_{G}(f(X_G))$---has been reduced to a new but smaller problem---the computation of $\underline{E}^\irr_{T}(f_9(X_T))$---in a chain that now has only nine nodes. By continuing in this way, removing the nodes of the chain one by one, we can eventually obtain an inference problem of the form $\underline{E}^\irr_1(f_1(X_1))$, which, as we know from Equation~\eqref{eq:trivialsubnetwork}, simply requires us to evaluate a local lower expectation.

For general functions $f$, the approach that is described above is rather inefficient because it has a complexity that is exponential in the length of the chain. For example, computing $f_9$ already has a complexity of the order $\prod_{i=1}^9\vert\values{i}\vert$, because it requires us to compute $f_9(x_T)$ for every $x_T\in\values{T}$. This is to be expected though, because for arbitrary $f\in\gambles{G}$, even simply printing $f$ has such a complexity.

Nevertheless, the approach that is described above is highly practical, because it can be adapted to specific types of functions, at which point the approach often does become very efficient. Several examples in support of this claim can be found in Reference~\cite{Lopatatzidis2015}. 

We here restrict ourselves to a simple prototypical case, where $f(X_G)=h(X_{10})$, with $h\in\gambles{10}$ a function of the last state in the chain. In this particular case, the equations above simplify as follows. First, since $f(X_G)=f(X_T,X_{10})=h(X_{10})$, we find that $f_9(X_T)=\underline{E}_{10\mid\xval{9}}(h(X_{10}))=\underline{T}_{10}(h)(X_9)$, where we let $\underline{T}_{10}$ be a map from $\gambles{10}$ to $\gambles{9}$, defined for all $g\in\gambles{10}$ by
\begin{equation*}
\underline{T}_{10}(g)(x_9)\coloneqq\underline{E}_{10\mid\xval{9}}(g(X_{10}))
\text{~~for all $x_9\in\values{9}$.}\vspace{-4pt}
\end{equation*}
Hence, it follows that
\begin{equation*}
\underline{E}^\irr_G(h(X_{10}))
=\underline{E}^\irr_T(\underline{T}_{10}(h)(X_9)).
\vspace{6pt}
\end{equation*}
The beauty of this equation is that the right hand side is of the same form as the left hand side. The only difference is that the last node of the chain has been removed, and that the role of $h$ has been replaced by $\underline{T}_{10}(h)$. Therefore, we can repeat the same argument over and over again. Eventually, we find that
\begin{equation}\label{eq:markovchain}
\underline{E}^\irr_G(h(X_{10}))
=\underline{E}^\irr_1(
\underline{T}_2(~\cdots~\underline{T}_9(\underline{T}_{10}(h))~\cdots~)
(X_1)
),
\end{equation}
with $\underline{E}^\irr_1$ equal to $\underline{E}_1$ because of Equation~\eqref{eq:trivialsubnetwork}. This computational procedure is linear in the length of the chain, and therefore highly efficient.
\hfill$\lozenge$
\end{example}

Readers that are familiar with the concept of an imprecise Markov chain~\cite{deCooman:2009jz,Skulj:2011db} should recognise Equation~\eqref{eq:markovchain}. Indeed, since imprecise Markov chains are credal networks whose graph is a chain, Equation~\eqref{eq:markovchain} provides a method for computing the lower expectation of a function on the last variable of an imprecise Markov chain. This method is well-known; see for example Reference~\cite{deCooman:2009jz}. As we have just seen, it can be easily derived from the properties in this section. Interestingly, Equation~\eqref{eq:markovchain} also remains valid if we consider a credal network under complete---or strong---independence; see for example Reference~\cite{Skulj:2011db}. In fact, in the context of imprecise Markov chains, the distinction between epistemic irrelevance, complete independence and strong independence is usually not made. Care should be taken though in dropping this distinction, because the equivalence does not extend to more complicated types of inferences. For example, for an imprecise Markov chain that adopts complete independence, Equation~\eqref{eq:MarkovGeneral} no longer holds.

As a third illustration of Theorem~\ref{theo:marginalextension}, in combination with Theorem~\ref{theo:subnetwork}, we consider an inference problem in a small dynamic credal network.

\begin{example}\label{exmp:cheesenatural}
Consider a network that consists of $3n$ nodes, with $n\in\mathbb{N}$. For every \mbox{$i\in\{1,\dots,n\}$}, the variables $\var{t_i}$, $\var{s_{ia}}$ and $\var{s_{ib}}$ represent domain-specific parameters at time $i$, whose local models depend on the value of these parameters in the previous time slot. This is a simple example of a dynamic network, which aims to model the evolution of parameters as time evolves. Figure~\ref{fig:dynamicnotdecomposable} depicts an example for $n=5$. For $n=14$, this network was used as an example in a recent paper about dynamic credal networks~\cite{Hourbracq:2013un}.  

As is explained that paper, this network is a simplified version of a network that models various aspects of the ripening process of Camembert cheese~\cite{Baudrit:2010cc}. In particular, it represents the coupled dynamics of a yeast behaviour---Kluyveromyces marxianus concentration $\var{s_{ia}}$---with their substrate consumptions---lactose concentration $\var{s_{ib}}$---influenced by temperature---$\var{t_i}$. For our present purposes, the local models that are attached to the variables in this network are not really important; see Reference~\cite{Hourbracq:2013un} for examples. What we intend to show here is that inference in this network can be performed efficiently.

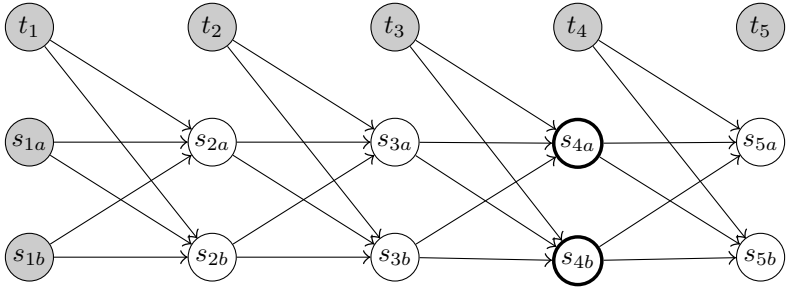
\begin{figure}
\begin{center}
\vspace{3pt}
\begin{tikzpicture}[node distance=15pt and 25pt
,>=triangle 45
,decoration={
      markings,
      mark=at position 0.999 with {\arrow[scale=1.5]{to}};
    }
]
\tikzstyle{punt} = [circle, draw, inner sep=0pt, minimum size=18pt]

\tikzstyle{punt2} = [circle, draw, line width=1.3pt, inner sep=0pt, minimum size=18pt]
\tikzstyle{punt3} = [circle, draw, densely dashed, inner sep=0pt, minimum size=18pt]
\tikzstyle{punt4} = [circle, draw, inner sep=0pt, minimum size=18pt, fill=black!20]

\node[punt4] (t1) {$t_1$};

\node[punt4] (s1a) [ below = of t1,yshift=-10pt] {$s_{1a}$};

\node[punt4] (s1b) [ below = of s1a,yshift=-10pt] {$s_{1b}$};

\node[punt4] (t2) [ right = of t1,xshift=25pt] {$t_2$};

\node[punt] (s2a) [ below = of t2,yshift=-10pt] {$s_{2a}$};

\node[punt] (s2b) [ below = of s2a,yshift=-10pt] {$s_{2b}$};

\node[punt4] (t3) [ right = of t2,xshift=25pt] {$t_3$};

\node[punt] (s3a) [ below = of t3,yshift=-10pt] {$s_{3a}$};

\node[punt] (s3b) [ below = of s3a,yshift=-10pt] {$s_{3b}$};

\node[punt4] (t4) [ right = of t3,xshift=25pt] {$t_4$};

\node[punt2] (s4a) [ below = of t4,yshift=-10pt] {$s_{4a}$};

\node[punt2] (s4b) [ below = of s4a,yshift=-10pt] {$s_{4b}$};

\node[punt4] (t5) [ right = of t4,xshift=25pt] {$t_5$};

\node[punt] (s5a) [ below = of t5,yshift=-10pt] {$s_{5a}$};

\node[punt] (s5b) [ below = of s5a,yshift=-10pt] {$s_{5b}$};

\draw[postaction={decorate}] (t1) -- (s2a);
\draw[postaction={decorate}] (t1) -- (s2b);
\draw[postaction={decorate}] (s1a) -- (s2a);
\draw[postaction={decorate}] (s1a) -- (s2b);
\draw[postaction={decorate}] (s1b) -- (s2a);
\draw[postaction={decorate}] (s1b) -- (s2b);

\draw[postaction={decorate}] (t2) -- (s3a);
\draw[postaction={decorate}] (t2) -- (s3b);
\draw[postaction={decorate}] (s2a) -- (s3a);
\draw[postaction={decorate}] (s2a) -- (s3b);
\draw[postaction={decorate}] (s2b) -- (s3a);
\draw[postaction={decorate}] (s2b) -- (s3b);

\draw[postaction={decorate}] (t3) -- (s4a);
\draw[postaction={decorate}] (t3) -- (s4b);
\draw[postaction={decorate}] (s3a) -- (s4a);
\draw[postaction={decorate}] (s3a) -- (s4b);
\draw[postaction={decorate}] (s3b) -- (s4a);
\draw[postaction={decorate}] (s3b) -- (s4b);

\draw[postaction={decorate}] (t4) -- (s5a);
\draw[postaction={decorate}] (t4) -- (s5b);
\draw[postaction={decorate}] (s4a) -- (s5a);
\draw[postaction={decorate}] (s4a) -- (s5b);
\draw[postaction={decorate}] (s4b) -- (s5a);
\draw[postaction={decorate}] (s4b) -- (s5b);

\end{tikzpicture}
\vspace{8pt}
\caption{Example of a very simple dynamic network}
\label{fig:dynamicnotdecomposable}
\end{center}
\vspace{-2pt}
\end{figure}

We focus on one particular inference problem, which is estimating the evolution of the variables $\var{s_{ia}}$ and $\var{s_{ib}}$ through time, based on their value at time $1$ and the evolution of the temperature---$\var{t_i}$---over time. In order to formalise this problem we define
\vspace{1pt}
\begin{equation*}
S_i\coloneqq\{s_{ia},s_{ib}\}
\text{~for all $i\in\{1,\dots,n\}$}
\vspace{-8pt}
\end{equation*}
and
\begin{equation*}
S_{k:\ell}\coloneqq\cup_{i=k}^\ell S_i
\text{ and }
T_{k:\ell}\coloneqq\{t_k,\dots,t_\ell\}
\text{~for all $k,\ell\in\{1,\dots,n\}$ such that $k\leq\ell$.}
\vspace{7pt}
\end{equation*}

\noindent
Let $m\in\{2,\dots,n\}$ be some arbitrary point in time and consider a gamble $h\in\gambles{S_m}$, a pair of initial states $\xval{S_1}\in\values{S_1}$ and a sequence of temperatures \mbox{$\xval{T_{1:n}}\in\values{T_{1:n}}$}. The generic inference problem that we intend to solve is the computation of $\underline{E}_{G}^\irr(h(X_{S_m})\newmid\xval{S_1\cup T_{1:n}})$, with $G=T_{1:n}\cup S_{1:n}$. The example in Figure~\ref{fig:dynamicnotdecomposable} corresponds to $m=4$. Reference~\cite{Hourbracq:2013un} considered a specific case of this inference problem,\footnote{Instead of general inferences about $\var{S_m}$, they computed lower and upper expected values of $\var{s_{ma}}$ and $\var{s_{mb}}$, which corresponds to choosing specific gambles $h\in\gambles{S_m}$.} for a credal network that has the same graphical structure but different independence assumptions---strong independence and a notion of independence which they call repetitive independence. Since no exact algorithm was available, the authors of Reference~\cite{Hourbracq:2013un} used an approximate Monte Carlo sampling algorithm to compute their inferences. As we are about to show, in our case---for a credal network under epistemic irrelevance---these inferences can be computed efficiently and exactly, with a recursive algorithm that has a computational complexity that is linear in $m$.

The first step consists in applying Theorem~\ref{theo:subnetwork} to simplify the inference problem. 
In particular, since $S_{2:m}$ is a closed subset of $G$, it follows from Theorem~\ref{theo:subnetwork}, with $K=S_{2:m}$ $B_K=\values{K}$, $f(X_K)=h(X_{S_m})$ and $B_{\nonparnondes{K}}=x_{T_{m:n}}$, that
\begin{equation*}
\underline{E}_{G}^\irr(h(X_{S_m})\newmid\xval{S_1\cup T_{1:n}})
=
\underline{E}_{S_{2:m}\mid\xval{\parents{S_{2:m}}}}^\irr(h(X_{S_m})),
\vspace{6pt}
\end{equation*}
where the right-hand side is an inference problem in the sub-network that corresponds to $S_{2:m}$ and $\xval{\parents{S_{2:m}}}$. Figure~\ref{fig:subnetworkdynamicnotdecomposable} depicts an example with $m=4$. The remaining task is to compute $\smash{\underline{E}_{S_{2:m}\mid\xval{\parents{S_{2:m}}}}^\irr(h(X_{S_m}))}$. 
In order to simplify this task further, the trick is to regard this sub-network as a new credal network, to which we can now again apply the results of this section and, in particular, Theorem~\ref{theo:marginalextension}. Indeed, since the only purpose of $\xval{\parents{S_{2:m}}}$ is to determine the local models of the sub-network, we can suppress them in our notation and can then think of our problem as that of computing $\smash{\underline{E}_{S_{2:m}}^\irr(h(X_{S_m}))}$ in a credal network whose graph is---for the case $m=4$---depicted in Figure~\ref{fig:subnetworkdynamicnotdecomposable:2}. 

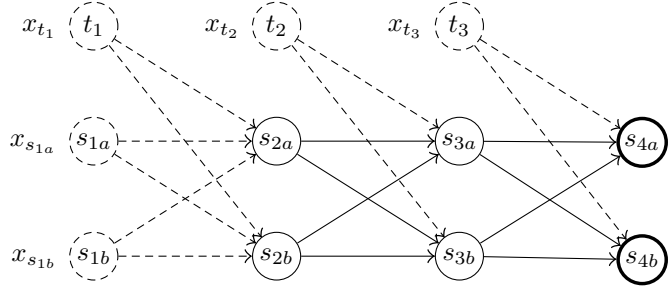
\begin{figure}
\begin{center}
\vspace{3pt}
\begin{tikzpicture}[node distance=15pt and 25pt
,>=triangle 45
,decoration={
      markings,
      mark=at position 0.999 with {\arrow[scale=1.5]{to}};
    }
]
\tikzstyle{punt} = [circle, draw, inner sep=0pt, minimum size=18pt]

\tikzstyle{punt2} = [circle, draw, line width=1.3pt, inner sep=0pt, minimum size=18pt]
\tikzstyle{punt3} = [circle, draw, densely dashed, inner sep=0pt, minimum size=18pt]
\tikzstyle{punt4} = [circle, draw, inner sep=0pt, minimum size=18pt, fill=black!10]

\node[punt3] (t1) {$t_1$};
\node (x1) [left=1pt of t1, yshift=-1pt] {$~\xval{t_1}$};

\node[punt3] (s1a) [ below = of t1,yshift=-10pt] {$s_{1a}$};
\node (x1a) [left=1pt of s1a, yshift=-1pt] {$~\xval{s_{1a}}$};

\node[punt3] (s1b) [ below = of s1a,yshift=-10pt] {$s_{1b}$};
\node (x1b) [left=1pt of s1b, yshift=-1pt] {$~\xval{s_{1b}}$};

\node[punt3] (t2) [ right = of t1,xshift=25pt] {$t_2$};
\node (x2) [left=1pt of t2, yshift=-1pt] {$~\xval{t_2}$};

\node[punt] (s2a) [ below = of t2,yshift=-10pt] {$s_{2a}$};

\node[punt] (s2b) [ below = of s2a,yshift=-10pt] {$s_{2b}$};

\node[punt3] (t3) [ right = of t2,xshift=25pt] {$t_3$};
\node (x3) [left=1pt of t3, yshift=-1pt] {$~\xval{t_3}$};

\node[punt] (s3a) [ below = of t3,yshift=-10pt] {$s_{3a}$};

\node[punt] (s3b) [ below = of s3a,yshift=-10pt] {$s_{3b}$};

\node[punt2] (s4a) [ below = of t4,yshift=-10pt] {$s_{4a}$};

\node[punt2] (s4b) [ below = of s4a,yshift=-10pt] {$s_{4b}$};

\draw[postaction={decorate}, densely dashed] (t1) -- (s2a);
\draw[postaction={decorate}, densely dashed] (t1) -- (s2b);
\draw[postaction={decorate}, densely dashed] (s1a) -- (s2a);
\draw[postaction={decorate}, densely dashed] (s1a) -- (s2b);
\draw[postaction={decorate}, densely dashed] (s1b) -- (s2a);
\draw[postaction={decorate}, densely dashed] (s1b) -- (s2b);

\draw[postaction={decorate}, densely dashed] (t2) -- (s3a);
\draw[postaction={decorate}, densely dashed] (t2) -- (s3b);
\draw[postaction={decorate}] (s2a) -- (s3a);
\draw[postaction={decorate}] (s2a) -- (s3b);
\draw[postaction={decorate}] (s2b) -- (s3a);
\draw[postaction={decorate}] (s2b) -- (s3b);

\draw[postaction={decorate}, densely dashed] (t3) -- (s4a);
\draw[postaction={decorate}, densely dashed] (t3) -- (s4b);
\draw[postaction={decorate}] (s3a) -- (s4a);
\draw[postaction={decorate}] (s3a) -- (s4b);
\draw[postaction={decorate}] (s3b) -- (s4a);
\draw[postaction={decorate}] (s3b) -- (s4b);

\end{tikzpicture}
\vspace{8pt}
\caption{Sub-network of the network in Figure~\ref{fig:dynamicnotdecomposable}}
\label{fig:subnetworkdynamicnotdecomposable}
\end{center}
\vspace{-2pt}
\end{figure}

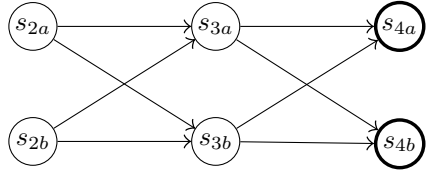
\begin{figure}
\begin{center}
\vspace{10pt}
\begin{tikzpicture}[node distance=15pt and 25pt
,>=triangle 45
,decoration={
      markings,
      mark=at position 0.999 with {\arrow[scale=1.5]{to}};
    }
]
\tikzstyle{punt} = [circle, draw, inner sep=0pt, minimum size=18pt]

\tikzstyle{punt2} = [circle, draw, line width=1.3pt, inner sep=0pt, minimum size=18pt]
\tikzstyle{punt3} = [circle, draw, densely dashed, inner sep=0pt, minimum size=18pt]
\tikzstyle{punt4} = [circle, draw, inner sep=0pt, minimum size=18pt, fill=black!10]

\node[punt] (s2a) [ right = of s1a,xshift=25pt] {$s_{2a}$};

\node[punt] (s2b) [ below = of s2a,yshift=-10pt] {$s_{2b}$};

\node[punt] (s3a) [ right = of s2a,xshift=25pt] {$s_{3a}$};

\node[punt] (s3b) [ below = of s3a,yshift=-10pt] {$s_{3b}$};

\node[punt2] (s4a) [ right = of s3a,xshift=25pt] {$s_{4a}$};

\node[punt2] (s4b) [ below = of s4a,yshift=-10pt] {$s_{4b}$};

\draw[postaction={decorate}] (s2a) -- (s3a);
\draw[postaction={decorate}] (s2a) -- (s3b);
\draw[postaction={decorate}] (s2b) -- (s3a);
\draw[postaction={decorate}] (s2b) -- (s3b);

\draw[postaction={decorate}] (s3a) -- (s4a);
\draw[postaction={decorate}] (s3a) -- (s4b);
\draw[postaction={decorate}] (s3b) -- (s4a);
\draw[postaction={decorate}] (s3b) -- (s4b);

\end{tikzpicture}
\vspace{8pt}
\caption{Simplified representation of the sub-network in Figure~\ref{fig:subnetworkdynamicnotdecomposable}}
\label{fig:subnetworkdynamicnotdecomposable:2}
\end{center}
\vspace{-2pt}
\end{figure}

As we will show, this problem can be solved recursively. Let $h_{m}\coloneqq h$. Then trivially, we have that $\underline{E}_{S_{2:m}}^\irr(h(X_{S_m}))
=\underline{E}_{S_{2:m}}^\irr(h_m(X_{S_m}))$. If we now apply Theorem~\ref{theo:marginalextension}, with $S=S_m$ and $T=S_{2:m-1}$, we find that
\begin{equation*}
\underline{E}_{S_{2:m}}^\irr(h_m(X_{S_m}))
=\underline{E}_{S_{2:m-1}}^\irr(
\underline{E}_{S_m\mid X_{S_{m-1}}}^\irr(
h_m(X_{S_m})
)
)
=\underline{E}_{S_{2:m-1}}^\irr(
h_{m-1}(X_{S_{m-1}})
),\vspace{-0pt}
\end{equation*}
with
\vspace{-3pt}
\begin{equation*}
h_{m-1}(x_{S_{m-1}})\coloneqq
\underline{E}_{S_m\mid x_{S_{m-1}}}^\irr(
h_{m}(X_{S_m})
)
\text{~~for all $x_{S_{m-1}}\in\values{S_{m-1}}$.}
\vspace{5pt}
\end{equation*}
In these expressions, $x_{S_{m-1}}$ is an instantiation of the value of the parents of $X_{S_m}$ in the network that is---for the case $m=4$---depicted in Figure~\ref{fig:subnetworkdynamicnotdecomposable:2}. However, it should not be forgotten that the local models of this network are derived from the network in Figure~\ref{fig:dynamicnotdecomposable}. In particular, as can for example be seen from Figure~\ref{fig:subnetworkdynamicnotdecomposable}, the sub-network that is used to compute $h_{m-1}$ does not only depend on $x_{S_{m-1}}$, but also on the fixed value $x_{t_{m-1}}$ of $X_{t_{m-1}}$ that corresponds to $\xval{\parents{S_{2:m}}}$. We suppress this value $x_{t_{m-1}}$ from the notation for the sake of clarity.

That being said, there are two important observations to be made here. First of all, the original problem has been reduced into a new yet smaller problem, where $m$ is now replaced by $m-1$. Secondly, in order to obtain this reduction, all we have to do is compute $h_{m-1}$, which requires us to solve a local inference problem in a network with two disconnected nodes. Due to the size of this local inference problem---that is, because the corresponding sub-graph consists of only two nodes---it can easily be solved by means of the linear programming methods in Section~\ref{sec:linearprogramming} and, if $X_{s_{ma}}$ and $X_{s_{mb}}$ are binary, by means of the specialized methods in Example~\ref{exmp:twobinarylinearprogram}.

By continuing in this way, the original problem can be made smaller and smaller, and we obtain the following recursive solution method. Let $h_m\coloneqq h$ as before and, for all $i\in\{3,\dots,m\}$, let $h_{i-1}\in\gambles{S_{i-1}}$ be defined by
\vspace{4pt}
\begin{equation*}
h_{i-1}(\xval{S_{i-1}})\coloneqq
\underline{E}_{S_i\mid x_{S_{i-1}}}^\irr(
h_i(X_{S_i})
)
\text{~~for all $\xval{S_{i-1}}\in\values{S_{i-1}}$}.
\vspace{7pt}
\end{equation*}
Again, the local models of the sub-network that is used to compute $h_{i-1}$ do not only depend on $x_{S_{i-1}}$, but also on the fixed value $x_{t_{i-1}}$ of $X_{t_{i-1}}$ that corresponds to $\xval{\parents{S_{2:m}}}$. The value of the final inference that we are after is now given by
\begin{equation*}
\underline{E}_{S_{2:m}}^\irr(h(X_{S_m}))
=
\underline{E}_{S_{2}}^\irr(h_2(X_{S_2})).
\vspace{7pt}
\end{equation*}
Once more, the local models of the sub-network that is used to compute $\underline{E}_{S_{2}}^\irr(h_2(X_{S_2}))$ depend on $\xval{\parents{S_{2:m}}}$. In this case, in particular, as can be seen from Figure~\ref{fig:subnetworkdynamicnotdecomposable}, these local models depend on the values of $X_{S_{i-1}}$ and $X_{t_{i-1}}$.

If the local computation of each of the functions $g_i$ can be performed in a reasonable---say constant---amount of time, the recursive procedure that is outlined above has a computational complexity that is linear in $m$.\hfill$\lozenge$
\end{example}

\subsection{Other Decomposition Properties, Including Factorisation and Additivity}

Besides the law of iterated expectation in Theorem~\ref{theo:marginalextension}, there are also other properties that are able to decompose an inference problem in a credal network under epistemic irrelevance into several similar yet smaller problems. The most general among them seems to be the following one.

\begin{theorem}\label{theo:sub-DAG:factorisationandadditivity:LP}
Let $K$ be a closed subset of $G$. Then for any $f\in\gambles{K}$, $\xval{\parents{K}}\in\values{\parents{K}}$, $h\in\gambles{\nondes{K}}$ and any non-negative $g\in\gambles{{\nonparnondes{K}}}$:
\vspace{-3pt}
\begin{multline*}
\underline{E}_G^\irr\big(h(\var{\nondes{K}})+g(\var{\nonparnondes{K}})\ind{\xval{\parents{K}}}(\var{\parents{K}})f(\var{K})\big)\\
=\underline{E}_{\nondes{K}}^\irr\big(h(\var{\nondes{K}})+g(\var{\nonparnondes{K}})\ind{\xval{\parents{K}}}(\var{\parents{K}})\underline{E}_{K\mid\xval{\parents{K}}}^\irr(f(\var{K}))\big).
\end{multline*}
\vspace{-4pt}
\end{theorem}
At first sight, this result might seem a bit complicated. However, upon closer inspection, it should become clear that it is in fact not. As in Theorem~\ref{theo:marginalextension}, the essential feature here is again that the left hand side is a lower expectation of a function that depends on all the variables $\{\var{s}\}_{s\in G}$, whereas the right hand side consists of two separate lower expectations, each of which depends on fewer variables. The following two corollaries of Theorem~\ref{theo:sub-DAG:factorisationandadditivity:LP} highlight this feature even more.

\begin{corollary}[factorisation]\label{corol:sub-DAG:factorisation:LP}
Let $K$ be a closed subset of $\nodes$. Then for any $f\in\gambles{K}$ and $\xval{\parents{K}}\in\values{\parents{K}}$ and any $g\in\gambles{{\nonparnondes{K}}}$ such that $g\geq0$:
\begin{multline*}
\underline{E}_G^\irr\big(g(\var{\nonparnondes{K}})\ind{\xval{\parents{K}}}(\var{\parents{K}})f(\var{K})\big)\\
=
\begin{cases}
\underline{E}_{K\mid\xval{\parents{K}}}^\irr\big(f(\var{K})\big)\,\underline{E}_{\nondes{K}}^\irr\big(g(\var{\nonparnondes{K}})\ind{\xval{\parents{K}}}(\var{\parents{K}})\big)
&\text{if\, $\underline{E}_{K\mid\xval{\parents{K}}}^\irr\big(f(\var{K})\big)\,\geq0$}\\
\underline{E}_{K\mid\xval{\parents{K}}}^\irr\big(f(\var{K})\big)\,\overline{E}_{\nondes{K}}^\irr\big(g(\var{\nonparnondes{K}})\ind{\xval{\parents{K}}}(\var{\parents{K}})\big)
&\text{if\, $\underline{E}_{K\mid\xval{\parents{K}}}^\irr\big(f(\var{K})\big)\,\leq0$}.
\end{cases}
\end{multline*}
\vspace{-2pt}
\end{corollary}

\begin{corollary}[external additivity]\label{corol:sub-DAG:additivity:LP}
Let $K$ be a closed subset of $\nodes$ such that $\parents{K}=\emptyset$. Then for any $f\in\gambles{K}$ and $h\in\gambles{\nonparnondes{K}}$:
\begin{equation*}
\underline{E}_G^\irr\big(h(\var{\nonparnondes{K}})+f(\var{K})\big)
=\underline{E}_{\nonparnondes{K}}^\irr\big(h(\var{\nonparnondes{K}})\big)+\underline{E}_K^\irr\big(f(\var{K})\big).\vspace{6pt}
\end{equation*}
\end{corollary}

Both of these corollaries are illustrated in our next example, where we apply them to a very simple network with two disconnected nodes.

\begin{example}\label{exmp:twodisconnectedwithoutlinearprogramming}
Consider again the network in Figure~\ref{fig:twodisconnectednodes}, which consists of two disconnected nodes. This network is identical to that of Examples~\ref{exmp:extensionoftwodisconnectednodes} and~\ref{exmp:twobinarylinearprogram}. This time though, we do not require the corresponding variables $\var{1}$ and $\var{2}$ to be binary.

In this network, for $K\coloneqq\{1\}$, we find that $\parents{K}=\emptyset$ and $\nondes{K}=\nonparnondes{K}=\{2\}$. Furthermore, for any $i\in\{1,2\}$ and any $f\in\gambles{i}$, we know from Equation~\eqref{eq:trivialsubnetwork} that $\underline{E}_i^\irr(f(X_i))=\underline{E}_i(f(X_i))$. Therefore, as a straightforward consequence of Corollary~\ref{corol:sub-DAG:additivity:LP}, we find that
\begin{equation*}
\underline{E}^\irr_G(f(X_1)+h(X_2))=\underline{E}_1(f(X_1))+\underline{E}_2(h(X_2))
\text{~~for all $f\in\gambles{1}$ and $h\in\gambles{2}$.}
\end{equation*}
Similarly, Corollary~\ref{corol:sub-DAG:additivity:LP} implies that
\vspace{-3pt}
\begin{equation*}
\underline{E}^\irr_G(f(X_1)g(X_2))
=
\begin{cases}
\underline{E}_1(f(X_1))\underline{E}_2(g(X_2))&\text{if\, $\underline{E}_1(f(X_1))\geq0$}\\
\underline{E}_1(f(X_1))\overline{E}_2(g(X_2))&\text{if\, $\underline{E}_1(f(X_1))\leq0$}
\end{cases}
\vspace{4pt}
\end{equation*}
for all $f\in\gambles{1}$ and $g\in\gambles{2}$ such that $g\geq0$. 

For a credal network under complete independence, similar expressions trivially hold, because they follow directly from Equation~\eqref{eq:factorisation}. This not the case for credal networks under epistemic irrelevance though, because for them, Equation~\eqref{eq:factorisation} no longer holds. Nevertheless, as we have just seen, the expressions above can still be obtained, by means of Corollaries~\ref{corol:sub-DAG:factorisation:LP} and~\ref{corol:sub-DAG:additivity:LP}.
\hfill$\lozenge$
\end{example}

Readers that are familiar with the notion of independent natural extension~\cite{deCooman:2011ey} will probably have recognized the expressions in this example. This is no coincidence. Indeed, as we already mentioned in Example~\ref{exmp:twobinarylinearprogram} for binary variables, the unconditional part of the irrelevant natural extension of a network with two disconnected nodes is equal to the independent natural extension of its two local models. This is also true for non-binary variables, as well as for networks that consist of multiple---possibly more than two---disconnected nodes~\cite{DeBock:2013wf}. Due to this connection, the equations in Example~\ref{exmp:twodisconnectedwithoutlinearprogramming} follow directly from the factorisation and additivity properties of the independent natural extension; see for example Reference~\cite[Theorem~22 and Proposition~27]{deCooman:2011ey}.

These properties of the independent natural extension have been very important to the development of algorithms for credal trees under epistemic irrelevance. In particular, the algorithms in Reference~\cite{deCooman:2010gd} are basically a combination of the factorisation property of the independent natural extension with a specific case of the law of iterated lower expectation in Theorem~\ref{theo:marginalextension}. In this sense, these algorithms can be regarded as a consequence of Theorems~\ref{theo:marginalextension} and~\ref{theo:sub-DAG:factorisationandadditivity:LP}. It should therefore not be surprising that these propositions---and their corollaries---can be used to develop inference algorithms for credal trees under epistemic irrelevance.
The following example illustrates that this is indeed possible, for the simple case of an imprecise hidden Markov model.

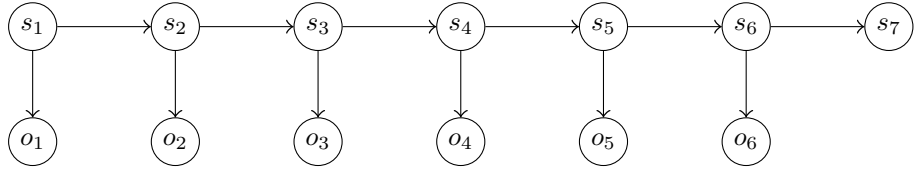
\begin{figure}
\begin{center}
\vspace{10pt}
\begin{tikzpicture}[node distance=15pt and 25pt
,>=triangle 45
,decoration={
      markings,
      mark=at position 0.999 with {\arrow[scale=1.5]{to}};
    }
]
\tikzstyle{punt} = [circle, draw, inner sep=0pt, minimum size=18pt]

\tikzstyle{punt2} = [circle, draw, line width=1.3pt, inner sep=0pt, minimum size=18pt]
\tikzstyle{punt3} = [circle, draw, densely dashed, inner sep=0pt, minimum size=18pt]
\tikzstyle{punt4} = [circle, draw, inner sep=0pt, minimum size=18pt, fill=black!10]

\node[punt] (s1) {$s_1$};

\node[punt] (s2) [ right = of s1, xshift=10pt] {$s_2$};

\node[punt] (s3) [ right = of s2, xshift=10pt] {$s_3$};

\node[punt] (s4) [ right = of s3, xshift=10pt] {$s_4$};

\node[punt] (s5) [ right = of s4, xshift=10pt] {$s_5$};

\node[punt] (s6) [ right = of s5, xshift=10pt] {$s_6$};

\node[punt] (s7) [ right = of s6, xshift=10pt] {$s_7$};

\node[punt] (o1) [ below = of s1,yshift=-10pt] {$o_1$};

\node[punt] (o2) [ below = of s2,yshift=-10pt] {$o_2$};

\node[punt] (o3) [ below = of s3,yshift=-10pt] {$o_3$};

\node[punt] (o4) [ below = of s4,yshift=-10pt] {$o_4$};

\node[punt] (o5) [ below = of s5,yshift=-10pt] {$o_5$};

\node[punt] (o6) [ below = of s6,yshift=-10pt] {$o_6$};

\draw[postaction={decorate}] (s1) -- (s2);
\draw[postaction={decorate}] (s2) -- (s3);
\draw[postaction={decorate}] (s3) -- (s4);
\draw[postaction={decorate}] (s4) -- (s5);
\draw[postaction={decorate}] (s5) -- (s6);
\draw[postaction={decorate}] (s6) -- (s7);

\draw[postaction={decorate}] (s1) -- (o1);
\draw[postaction={decorate}] (s2) -- (o2);
\draw[postaction={decorate}] (s3) -- (o3);
\draw[postaction={decorate}] (s4) -- (o4);
\draw[postaction={decorate}] (s5) -- (o5);
\draw[postaction={decorate}] (s6) -- (o6);
\end{tikzpicture}
\vspace{8pt}
\caption{The DAG of an imprecise hidden Markov model}
\label{fig:HMM}
\end{center}
\vspace{-2pt}
\end{figure}

\begin{example}\label{exmp:HMMforwardunconditional}
Consider a credal network that has $2n+1$ variables, with $n\in\mathbb{N}$. The state variables $X_{s_i}$, for $i\in\{1,\dots,n,n+1\}$, form a chain. The observation variables $X_{o_i}$, for $i\in\{1,\dots,n\}$, are connected to their corresponding state variable by means of a single arrow. Figure~\ref{fig:HMM} illustrates this graphical structure for the case $n=6$. In these networks, the observation variables are typically observed, and the inference task then consist in using these observations to infer something about the state variables. For this reason, the state variables are referred to as `hidden'. In the Bayesian network case, such a network is therefore called a hidden Markov model. In the case of a credal network, it is called an imprecise hidden Markov model.

The ultimate goal of this example is to illustrate how to compute conditional lower expectations of the form $\underline{E}_G^\irr(f(X_{s_{n+1}})\newmid\xval{o_1},\dots,\xval{o_n})$ in such a network, for some $f\in\gambles{s_{n+1}}$ and, for all $i\in\{1,\dots,n\}$, some $x_{o_i}\in\values{o_i}$. However, at the moment, we do not yet have the theoretical tools to tackle this problem, and we therefore postpone this task to Example~\ref{exmp:HMMforward} in Section~\ref{sec:conditioningversusupdating}. For now, we restrict ourselves to a simpler problem, which is to compute lower expectations of the form $\underline{E}_G^\irr\big(\ind{x_{o_1}}(X_{o_1})\cdots\ind{x_{o_n}}(X_{o_n})f(X_{s_{n+1}})\big)$. This may seem abstract, but it already includes as a particular case the computation of the lower probability $\smash{\underline{P}_G^\irr\big(x_{o_1},\cdots,x_{o_n}\big)}$ of the observations: as we know from Section~\ref{sec:lowerexpectations}, this correspond to choosing $f=1$. Similarly, we find that
\begin{equation*}
\overline{P}_G^\irr\big(x_{o_1},\cdots,x_{o_n}\big)
=
\overline{E}_G^\irr\big(\ind{x_{o_1}}(X_{o_1})\cdots\ind{x_{o_n}}(X_{o_n})\big)
=
-
\underline{E}_G^\irr\big(-\ind{x_{o_1}}(X_{o_1})\cdots\ind{x_{o_n}}(X_{o_n})\big),
\end{equation*}
\noindent
which can be computed by choosing $f=-1$. Furthermore, as we will see in Example~\ref{exmp:HMMforwardunconditional}, being able to solve this simpler problem serves as a crucial first step towards computing lower expectations of the form $\underline{E}^\irr_G(f(X_{s_{n+1}})\newmid\xval{o_1},\dots,\xval{o_n})$.

For all $k\in\{1,\dots,n+1\}$, we define the sets of nodes $S_k\coloneqq\{s_k,\dots,s_{n+1}\}$, $O_k\coloneqq\{o_k,\dots,o_n\}$ and $G_k\coloneqq S_k\cup O_k$, and we let $h_k$ be a real-valued function on $\values{\parents{s_k}}$, defined for all $\xval{\parents{s_k}}\in\values{\parents{s_k}}$ by
\begin{equation*}
h_k(\xval{\parents{s_k}})
\coloneqq\underline{E}^\irr_{G_k\mid\xval{\parents{s_k}}}\big(\ind{\xval{O_k}}(X_{O_k})f(X_{s_{n+1}})\big)
=
\underline{E}^\irr_{G_k\mid\xval{\parents{s_k}}}\big(\ind{\xval{o_k}}(X_{o_k})\cdots\ind{\xval{o_n}}(X_{o_n})f(X_{s_{n+1}})\big).
\end{equation*}
Since $G_1=G$, the lower expectation that we want to compute is then equal to $h_1(\xval{\parents{s_1}})$, where the value of $\xval{\parents{s_1}}$ is purely symbolic because $\parents{s_1}=\emptyset$. We will now compute $h_1(\xval{\parents{s_1}})$ recursively.

Consider any $k\in\{1,\dots,n\}$ and any $x_{\parents{s_k}}\in\values{s_k}$. Then on the one hand, by applying Theorem~\ref{theo:marginalextension} to the subnetwork that corresponds to $G_k$ and $\xval{\parents{s_k}}$, we find that
\begin{equation}\label{eq:HMM1}
h_k(\xval{\parents{s_k}})
=\underline{E}^\irr_{G_k\mid\xval{\parents{s_k}}}\big(\ind{\xval{O_k}}(X_{O_k})f(X_{s_{n+1}})\big)
=\underline{E}^\irr_{s_k\mid\xval{\parents{s_k}}}
\big(
\underline{E}^\irr_{G_{k+1}\cup\{o_k\}\mid X_{s_k}}
\big(\ind{\xval{O_k}}(X_{O_k})f(X_{s_{n+1}})\big)
\big).
\end{equation}
On the other hand, for any $\xval{s_k}\in\values{s_k}$, since $\parents{s_{k+1}}=s_k$ and $\ind{\xval{o_k}}\geq0$, applying Corollary~\ref{corol:sub-DAG:factorisation:LP} to the subnetwork that corresponds to $G_{k+1}\cup\{o_k\}$ and $\xval{s_k}$ implies that
\begin{align}
\underline{E}^\irr_{G_{k+1}\cup\{o_k\}\mid x_{s_k}}
\big(\ind{\xval{O_k}}(X_{O_k})f(X_{s_{n+1}})\big)
&=\underline{E}^\irr_{G_{k+1}\cup\{o_k\}\mid x_{s_k}}
\big(
\ind{\xval{o_k}}(X_{o_k})
\ind{\xval{O_{k+1}}}(X_{O_{k+1}})f(X_{s_{n+1}})
\big)\notag\\
&=
\begin{cases}
h_{k+1}(\xval{\parents{s_{k+1}}})
\underline{E}^\irr_{o_k\mid\xval{s_k}}\big(\ind{\xval{o_k}}(X_{o_k})\big)
&\text{if~~$h_{k+1}(\xval{\parents{s_{k+1}}})\geq0$}\\
h_{k+1}(\xval{\parents{s_{k+1}}})
\overline{E}^\irr_{o_k\mid\xval{s_k}}\big(\ind{\xval{o_k}}(X_{o_k})\big)
&\text{if~~$h_{k+1}(\xval{\parents{s_{k+1}}})\leq0$}.\label{eq:HMM2}
\end{cases}
\end{align}
Since we know from Equation~\eqref{eq:trivialsubnetwork} that
\begin{equation*}
\underline{E}^\irr_{o_k\mid\xval{s_k}}\big(\ind{\xval{o_k}}(X_{o_k})\big)=\underline{P}_{o_k\mid\xval{s_k}}(\xval{o_k})
\text{~~and~~}\overline{E}^\irr_{o_k\mid\xval{s_k}}\big(\ind{\xval{o_k}}(X_{o_k})\big)=\overline{P}_{o_k\mid\xval{s_k}}(\xval{o_k})
\end{equation*}
and that the unconditional part of $\smash{\underline{E}^\irr_{s_k\mid\xval{\parents{s_k}}}}$ is equal to the local lower expectation $\smash{\underline{E}_{s_k\mid\xval{\parents{s_k}}}}$,
Equations~\eqref{eq:HMM1} and~\eqref{eq:HMM2} provide a simple recursive method for computing $h_k$, which only requires local computations. In order to start the recursion, we use the fact that 
\begin{equation}\label{eq:startHMMrecursion}
h_{n+1}(\xval{\parents{s_{n+1}}})
=
\underline{E}^\irr_{s_{n+1}\mid\xval{\parents{s_{n+1}}}}\big(f(X_{s_{n+1}})\big)
=
\underline{E}_{s_{n+1}\mid\xval{\parents{s_{n+1}}}}(f),
\end{equation}
where we use Equation~\eqref{eq:trivialsubnetwork} for the last equality. Starting from this point, a simple backward propagation scheme will now yield the desired value of $h_1(\xval{\parents{s_1}})$.
\hfill$\lozenge$
\end{example}

Similar decomposition methods can also be developed for other types of inferences; see for example References~\cite{deCooman:2010gd,DeBock:2014ts}. They both consider the case of credal trees, and they both make use of the factorisation property of the independent natural extension, in combination with a specialized version of the law of iterated lower expectation in Theorem~\ref{theo:marginalextension}.
Therefore, in all these cases, the resulting algorithms can be regarded as a consequence of Theorems~\ref{theo:marginalextension} and~\ref{theo:sub-DAG:factorisationandadditivity:LP}.

What is important here though is that the generality of Theorems~\ref{theo:marginalextension} and~\ref{theo:sub-DAG:factorisationandadditivity:LP} also enables us to develop similar decomposition methods for credal networks whose graph is not a tree, thereby allowing us to move beyond the cases that are considered in References~\cite{deCooman:2010gd} and~\cite{DeBock:2014ts}. We justify this claim by means of two examples. Example~\ref{exmp:historyHMMforwardunconditional} is an extension of Example~\ref{exmp:HMMforward} to the case of a second order imprecise hidden Markov model, the graph of which is no longer a tree. Example~\ref{exmp:atomfactorisation} applies to arbitrary credal networks under epistemic irrelevance, regardless of their graphical structure.

\begin{figure}
\begin{center}
\vspace{10pt}
\begin{tikzpicture}[node distance=15pt and 25pt
,>=triangle 45
,decoration={
      markings,
      mark=at position 0.999 with {\arrow[scale=1.5]{to}};
    }
]
\tikzstyle{punt} = [circle, draw, inner sep=0pt, minimum size=18pt]

\tikzstyle{punt2} = [circle, draw, line width=1.3pt, inner sep=0pt, minimum size=18pt]
\tikzstyle{punt3} = [circle, draw, densely dashed, inner sep=0pt, minimum size=18pt]
\tikzstyle{punt4} = [circle, draw, inner sep=0pt, minimum size=18pt, fill=black!10]

\node[punt] (s1) {$s_1$};

\node[punt] (s2) [ right = of s1, xshift=10pt] {$s_2$};

\node[punt] (s3) [ right = of s2, xshift=10pt] {$s_3$};

\node[punt] (s4) [ right = of s3, xshift=10pt] {$s_4$};

\node[punt] (s5) [ right = of s4, xshift=10pt] {$s_5$};

\node[punt] (s6) [ right = of s5, xshift=10pt] {$s_6$};

\node[punt] (s7) [ right = of s6, xshift=10pt] {$s_7$};

\node[punt] (o1) [ below = of s1,yshift=-10pt] {$o_1$};

\node[punt] (o2) [ below = of s2,yshift=-10pt] {$o_2$};

\node[punt] (o3) [ below = of s3,yshift=-10pt] {$o_3$};

\node[punt] (o4) [ below = of s4,yshift=-10pt] {$o_4$};

\node[punt] (o5) [ below = of s5,yshift=-10pt] {$o_5$};

\node[punt] (o6) [ below = of s6,yshift=-10pt] {$o_6$};

\draw[postaction={decorate}] (s1) -- (s2);
\draw (s1) edge[out=45, in=135, postaction={decorate}] (s3);
\draw[postaction={decorate}] (s2) -- (s3);
\draw (s2) edge[out=45, in=135, postaction={decorate}] (s4);
\draw[postaction={decorate}] (s3) -- (s4);
\draw (s3) edge[out=45, in=135, postaction={decorate}] (s5);
\draw[postaction={decorate}] (s4) -- (s5);
\draw (s4) edge[out=45, in=135, postaction={decorate}] (s6);
\draw[postaction={decorate}] (s5) -- (s6);
\draw (s5) edge[out=45, in=135, postaction={decorate}] (s7);
\draw[postaction={decorate}] (s6) -- (s7);

\draw[postaction={decorate}] (s1) -- (o1);
\draw[postaction={decorate}] (s2) -- (o2);
\draw[postaction={decorate}] (s3) -- (o3);
\draw[postaction={decorate}] (s4) -- (o4);
\draw[postaction={decorate}] (s5) -- (o5);
\draw[postaction={decorate}] (s6) -- (o6);
\end{tikzpicture}
\vspace{8pt}
\caption{The DAG of a second order imprecise hidden Markov model}
\label{fig:secondorderHMM}
\end{center}
\vspace{-2pt}
\end{figure}
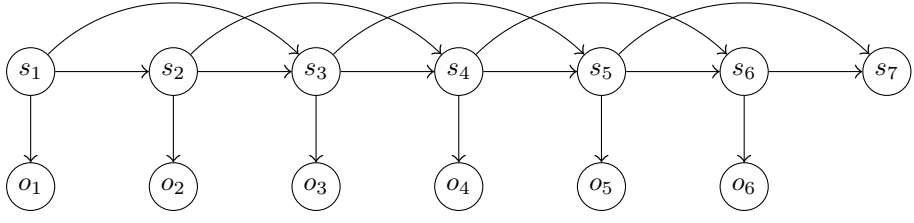

\begin{example}\label{exmp:historyHMMforwardunconditional}
Consider a similar set-up as in Example~\ref{exmp:HMMforwardunconditional}, the only difference being that the local model for the state variable $X_{s_i}$ now depends on the value of the two previous states instead of only that of the previous one. See Figure~\ref{fig:secondorderHMM} for an example with $n=6$. In the Bayesian network case, such a network is called a second order hidden Markov model. Similarly, in the case of a credal network, we call it a second order imprecise hidden Markov model.

Our aim here is again to compute a lower expectation of the form $\underline{E}_G^\irr\big(\ind{x_{o_1}}(X_{o_1})\cdots\ind{x_{o_n}}(X_{o_n})f(X_{s_{n+1}})\big)$. As we know from Example~\ref{exmp:HMMforwardunconditional}, this includes as particular cases the computation of $\smash{\underline{P}_G^\irr\big(x_{o_1},\cdots,x_{o_n}\big)}$ and $\smash{\overline{P}_G^\irr\big(x_{o_1},\cdots,x_{o_n}\big)}$. Furthermore, as we will see in Example~\ref{exmp:HMMforward}, it also allows us to compute inferences of the form $\underline{E}_G^\irr(f(X_{s_{n+1}})\newmid\xval{o_1},\dots,\xval{o_n})$.

For any $k\in\{1,\dots,n+1\}$, let $S_k$, $O_k$, $G_k$ and $h_k$ be defined as in Example~\ref{exmp:HMMforwardunconditional}. Then as in that example, the lower expectation that we want to compute is equal to $h_1(\xval{\parents{s_1}})$, with $\xval{\parents{s_1}}$ purely symbolic because $\parents{s_1}=\emptyset$. Similarly, $h_{n+1}(\xval{\parents{s_{n+1}}})$ is again given by Equation~\eqref{eq:startHMMrecursion}. Note though that in this case $s_{n+1}$ has two parents, whereas in Example~\ref{exmp:HMMforwardunconditional} it only had one.

The remaining step consists in developing a backward propagation scheme that allows us to easily compute $h_k$ based on $h_{k+1}$. For $k\geq2$, using an argument that is analogous to that in Example~\ref{exmp:HMMforwardunconditional}, we find that
\begin{equation*}
h_k(\xval{\parents{s_k}})
=\underline{E}^\irr_{G_k\mid\xval{\parents{s_k}}}\big(\ind{\xval{O_k}}(X_{O_k})f(X_{s_{n+1}})\big)
=\underline{E}_{s_k\mid\xval{\parents{s_k}}}
\big(
\underline{E}^\irr_{G_{k+1}\cup\{o_k\}\mid(X_{s_k},x_{s_{k-1}})}
\big(\ind{\xval{O_k}}(X_{O_k})f(X_{s_{n+1}})\big)
\big),
\end{equation*}
where, for all $x_{s_k}\in\values{s_k}$, we have that
\vspace{-4pt}
\begin{equation*}
\underline{E}^\irr_{G_{k+1}\cup\{o_k\}\mid(x_{s_k},x_{s_{k-1}})}
\big(\ind{\xval{O_k}}(X_{O_k})f(X_{s_{n+1}})\big)
=
\begin{cases}
h_{k+1}(\xval{\parents{s_{k+1}}})
\underline{P}_{o_k\mid\xval{s_k}}(\xval{o_k})
&\text{if~~$h_{k+1}(\xval{\parents{s_{k+1}}})\geq0$}\\
h_{k+1}(\xval{\parents{s_{k+1}}})
\overline{P}_{o_k\mid\xval{s_k}}(\xval{o_k})
&\text{if~~$h_{k+1}(\xval{\parents{s_{k+1}}})\leq0$}.
\end{cases}
\vspace{6pt}
\end{equation*}
For $k=1$, we can again use Equations~\eqref{eq:HMM1} and~\eqref{eq:HMM2}.
\hfill$\lozenge$
\end{example}

\begin{example}\label{exmp:atomfactorisation}
Consider an arbitrary credal network under epistemic irrelevance and let $G$ be its set of nodes. Consider any $\xval{G}\in\values{G}$. As we will show in this example, the lower and upper probability of $\xval{G}$ satisfy a global factorisation formula that is very similar to that of Bayesian networks; see Equation~\eqref{eq:factorisation}.

Let $\ell$ be an arbitrary leaf of the DAG of the network, that is, a node with no children (this is always possible, because every DAG has at least one leaf). Then \mbox{$\descendants{\ell}=\emptyset$} and $\nondes{\ell}=G\setminus\{\ell\}$, which implies that $\ind{\xval{G}}=\ind{\xval{\ell}}\ind{\xval{\parents{\ell}}}\ind{\xval{\nonparnondes{\ell}}}$.
Therefore, and because $\underline{E}^\irr_{s\rfloor\xval{\parents{s}}}(\ind{\xval{s}})\geq0$, we can apply Corollary~\ref{corol:sub-DAG:factorisation:LP} for $K=\{\ell\}$, $f=\ind{\xval{\ell}}$ and $g=\ind{\xval{\nonparnondes{\ell}}}$ to find that
\begin{equation*}
\underline{P}_G^\irr(\xval{G})
=\underline{E}_G^\irr(\ind{\xval{G}})
=\underline{E}^\irr_{G\setminus\{\ell\}}(\ind{\xval{G\setminus\{\ell\}}})
\underline{E}^\irr_{s\rfloor\xval{\parents{s}}}(\ind{\xval{s}})
=\underline{P}^\irr_{G\setminus\{\ell\}}(\xval{G\setminus\{\ell\}})
\underline{P}_{s\rfloor\xval{\parents{s}}}(\xval{s}),
\end{equation*}
where we have used Equation~\eqref{eq:trivialsubnetwork} for the last equality.
Similarly, since $\underline{E}^\irr_{s\rfloor\xval{\parents{s}}}(-\ind{\xval{s}})\leq0$, we find that
\begin{equation*}
\overline{P}_G^\irr(\xval{G})
=-\underline{E}_G^\irr(-\ind{\xval{G}})
=-\overline{E}^\irr_{G\setminus\{\ell\}}(\ind{\xval{G\setminus\{\ell\}}})
\underline{E}^\irr_{s\rfloor\xval{\parents{s}}}(-\ind{\xval{s}})
=\overline{P}^\irr_{G\setminus\{\ell\}}(\xval{G\setminus\{\ell\}})
\overline{P}_{s\rfloor\xval{\parents{s}}}(\xval{s}).
\end{equation*}
Exactly the same technique can now again be applied to the sub-network that corresponds to $G\setminus\{\ell\}$: it suffices to choose some arbitrary leaf of the corresponding sub-graph and factorise out the corresponding local lower or upper probability.
By continuing in this way, removing one node at the time, we eventually arrive at the following expressions:
\begin{equation}\label{eq:impreciselocalfactorisation}
\underline{P}_G^\irr(\xval{G})=\prod_{s\in G}\underline{P}_{s\rfloor\xval{\parents{s}}}(\xval{s})
\text{~and~}
\overline{P}_G^\irr(\xval{G})=\prod_{s\in G}\overline{P}_{s\rfloor\xval{\parents{s}}}(\xval{s}).
\end{equation}
Much like in Example~\ref{exmp:twodisconnectedwithoutlinearprogramming}, similar expressions trivially hold for credal networks under complete independence, because they then follow directly from Equation~\eqref{eq:factorisation}. 
\hfill$\lozenge$
\end{example}

\section{Epistemic h-irrelevance, AD-separation and Graphoid Axioms}\label{sec:separation}

As the reader may have noticed, the second equality in Theorem~\ref{theo:subnetwork} is redundant: it follows from the first equality by choosing $B_{\nonparnondes{K}}=\values{\nonparnondes{K}}$. The reason why we nevertheless state it explicitly, is because it illustrates that the irrelevant natural extension satisfies many more epistemic irrelevances than the basic ones that were used to define it in Equation~\eqref{eq:irrelevance:assessment}. In fact, it even satisfies statements of \emph{epistemic h-irrelevance}.

\begin{definition}\label{def:hirrelevance}\citep{Cozman:2013us}
For three pairwise disjoint sets $I,S,C\subseteq G$, we say that $\var{I}$ is (epistemically) h-irrelevant to $\var{S}$ conditional on $\var{C}$, and write $\HIR{I}{S}{C}$, if
\begin{align*}
\underline{E}_G^\irr\big(f(\var{S})\newmid B_S,\xval{C},B_{I}\big)
=
\underline{E}_G^\irr\big(f(\var{S})\newmid B_S,\xval{C}\big)
\end{align*}
for all $f\in\gambles{S}$, $B_S\in\mathcal{P}_\emptyset(\values{S})$, $\xval{C}\in\values{C}$ and $B_I\in\mathcal{P}_\emptyset(\values{I})$.
\end{definition}
\noindent
Indeed, 
Theorem~\ref{theo:subnetwork} clearly implies that for any closed subset $K$ of $G$, $\var{\nonparnondes{K}}$ is epistemically h-irrelevant to $\var{K}$ conditional on $\var{\parents{K}}$: $\HIR{\nonparnondes{K}}{K}{\parents{K}}$.

This statement of epistemic h-irrelevance is similar to the assessment of epistemic irrelevance that was imposed in Equation~\eqref{eq:irrelevance:assessment}---for $S=\{s\}$, $C=\parents{s}$ and $I=\nonparnondes{s}$---but differs on several levels. First of all: it is stated in terms of lower expectations, whereas Equation~\eqref{eq:irrelevance:assessment} was stated in terms of sets of probabilities. However, this is not so important; epistemic h-irrelevance can also be defined in terms of sets of probabilities~\citep{DeBock2015:thesis}. What really sets epistemic h-irrelevance apart from epistemic irrelevance is that it is far more powerful when it comes to conditional models. Unlike epistemic irrelevance, as can be seen from Definition~\ref{def:hirrelevance}, epistemic h-irrelevance requires \emph{all} information about the value of $\var{I}$---including \emph{partial} information---to be irrelevant to \emph{all} beliefs about $\var{S}$---conditional \emph{and} unconditional beliefs---conditional on the value of $\var{C}$.

As it turns out, the irrelevant natural extension satisfies many more statements of epistemic h-irrelevance than the ones that are implied by Theorem~\ref{theo:subnetwork}. Similarly to how for a Bayesian network, the well-known d-separation criterion implies stochastic independence, for the irrelevant natural extension, the so-called AD-separation criterion~\citep{DeBock2015:thesis,DeBock:2014bv} implies epistemic h-irrelevance; see Theorem~\ref{theo:ADimpliesHIR} below.

In order to define AD-separation, we require the concept of a blocked path. Consider any path $s_1,\dots,s_n$ in $\nodes$, with $n\geq1$. 
We say that this path is \emph{blocked} by a set of nodes $C\subseteq\nodes$ whenever at least one of the following four conditions holds:
\vspace{2pt}
\begin{enumerate}
[label=\upshape B\arabic*.,ref=\upshape B\arabic*,leftmargin=*]
\item\label{B1} $s_1\in C$;
\item\label{B2} there is a node $s_i$, with $1<i<n$, such that $s_i\rightarrow
  s_{i+1}$ and $s_i\in C$;
\item\label{B3} there is a node $s_i$, with $1<i<n$, such that $s_{i-1}\rightarrow
  s_i\leftarrow s_{i+1}$, $s_i\notin C$ and $\descendants{s_i}\cap C=\emptyset$;
\item\label{B4} $s_n\in C$.
\vspace{3pt}
\end{enumerate}

\noindent
Two sets of nodes are now said to be AD-separated by $C$ if every path between them is blocked by $C$.

\begin{definition}[AD-separation]\label{definition:ADseparation}
Consider (not necessarily pairwise disjoint) subsets $I$, $S$ and $C$ of $\nodes$. 
Then $I$ is \emph{AD-separated} from $S$ by $C$, denoted as $\AD{I}{S}{C}$, if every path \mbox{$i=s_1,\dots, s_n=s$}, $n\geq1$, from any node $i\in I$ to any node $s\in S$, is blocked by $C$.
\end{definition}

Readers that are familiar with d-separation will most likely recognise our definition for AD-separation; it is indeed very similar. The only difference is that d-separation adds an extra blocking condition:

\begin{enumerate}
[label=\upshape B2b.,ref=\upshape B2b,leftmargin=*]
\item\label{B2b} there is a node $s_i$, with $1<i<n$, such that $s_{i-1}\leftarrow
  s_{i}$ and $s_i\in C$.
\vspace{3pt}
\end{enumerate}

\noindent
The removal of condition~\ref{B2b} has two consequences for AD-separation. Firstly, it makes AD-separation asymmetric: $\AD{I}{S}{C}$ is not equivalent to $\AD{S}{I}{C}$. This also explains the name: the `A' in AD-separation is short for asymmetric. Secondly, it implies that AD-separation is a stronger notion of separation than d-separation, in the sense that is harder to satisfy. 

For our present purposes, it suffices to focus on the case where $I$, $S$ and $C$ are pairwise disjoint. In that case, our notion of AD-separation is equivalent to that of Moral~\cite{Moral:2005ce} and becomes a special case of Vantaggi's notion of L-separation~\cite{Vantaggi:2002va}, which also adds a logical component. More detailed information on the connection with these other notions of separation can be found in References~\cite{DeBock:2014bv,DeBock2015:thesis}, where it is also shown that AD-separation satisfies all asymmetric graphoid properties. We restrict ourselves here to providing the following alternative characterisation.

\begin{proposition}\label{prop:characteriseAD}
Consider any pairwise disjoint sets $I,S,C\subseteq G$. Then $\AD{I}{S}{C}$ if and only if there is some closed subset $K$ of $G$ such that $S\subseteq K$, $\parents{K}\subseteq C$, $I\subseteq\nonparnondes{K}$ and $\descendants{K}\cap C=\emptyset$.
\end{proposition}

\begin{example}\label{ex:ADseparation}
As a simple illustration of AD-separation, consider for example the sets $I=\{6\}$, $S=\{9\}$ and $C=\{3,4\}$. Then according to the graph in Figure~\ref{fig:DAGclosed}, as we already claimed at the end of Section~\ref{sec:subnetworks}, $I$ is AD-separated from $S$ by $C$. This claim can now be verified by applying Proposition~\ref{prop:characteriseAD} for $K=\{5,7,9\}$. In this particular case, the separation is symmetric, because $S$ is also AD-separated from $I$ by $C$---consider $K=\{6\}$ and apply Proposition~\ref{prop:characteriseAD}. However, this is not always the case. Consider for example the sets $I=\{1,6\}$, $S=\{5,7\}$ and $C=\{3,4,9\}$ in the same graph. Here too, applying Proposition~\ref{prop:characteriseAD} for $K=\{5,7,9\}$ leads us to conclude that $I$ is AD-separated from $S$ by $C$. However, in this case, $S$ is not AD-separated from $I$ by $C$ because the path $5\leftarrow3\leftarrow1$ is not blocked by $C$.
\hfill$\lozenge$
\end{example}

By combining Proposition~\ref{prop:characteriseAD} with Theorem~\ref{theo:subnetwork}, we obtain the following separation property for the irrelevant natural extension of a credal network.

\begin{theorem}\label{theo:ADimpliesHIR}
For any pairwise disjoint sets $I,S,C\subseteq G$ such that $\AD{I}{S}{C}$, the irrelevant natural extension $\underline{E}_G^{\irr}$ satisfies $\HIR{I}{S}{C}$.
\end{theorem}

The Bayesian network counterpart of this result---with AD-separation and epistemic h-irrelevance replaced by d-separation and stochastic independence---is proved by exploiting the fact that stochastic independence satisfies various graphoid properties (symmetry, redundancy, decomposition,weak union, contraction and intersection). Therefore, since epistemic irrelevance fails some of these graphoid properties~\citep{Cozman:2005fc}, it has long been thought that a similar result would not hold for credal networks under epistemic irrelevance. However, as Theorem~\ref{theo:ADimpliesHIR} shows, it is nevertheless possible to prove such a result. In order to do so, two steps were essential. The first step was to drop the symmetry of the separation criterion, by replacing d-separation with AD-separation; since epistemic irrelevance is asymmetric, symmetry is not to be expected anyway.
The second step was to not focus on graphoid properties, but to instead use other---more direct---methods to prove separation; for Theorems~\ref{theo:subnetwork} and~\ref{theo:ADimpliesHIR}---the proof of the latter is heavily based on the former---the main technical tool was a separating hyperplane result in terms of sets of desirable gambles; see Reference~\citep{DeBock:2014bv} for more information.

The fact that Theorem~\ref{theo:ADimpliesHIR} can be proved without the use of graphoid properties illustrates nicely that these properties are not essential, and that the fact that a notion of independence---such as epistemic irrelevance---fails some of them, should not be regarded as problematic. 
In fact, I think that the common practice of regarding these properties as axioms, and of comparing different notions of independence by means of the graphoid axioms that they satisfy, is flawed. Of course, when they are satisfied, graphoid properties are important and useful. However, one should be very careful in regarding them as axioms.  
For example, if we were to impose on epistemic h-irrelevance an asymmetric version of the `axiom' of contraction, it would require that
\begin{equation}\label{eq:examplecontraction}
\big(\HIR{I}{S}{C}
\text{ and }
\HIR{I}{W}{C\cup S}\big)
\Rightarrow
\HIR{I}{S\cup W}{C}.
\vspace{5pt}
\end{equation}
If we choose $C=\emptyset$ here, then basically, this property requires that if $\var{I}$ is irrelevant to our beliefs about $\var{S}$ and irrelevant to our conditional beliefs about $\var{W}$ given $\var{S}$, then $\var{I}$ should also be irrelevant to our joint beliefs about $\var{S\cup W}$.
I do not consider it reasonable to enforce this, because essentially, it requires that our beliefs about $\var{S\cup W}$ should be completely determined by our beliefs about $\var{S}$ and our conditional beliefs about $\var{W}$ given $\var{S}$. For probabilities, this is trivially true---under strict positivity assumptions---because it follows from Bayes's rule. However, for more general belief models, such as sets of probabilities, it is well known that this is not always the case. I think that this is perfectly normal, and that there is no fundamental reason why such a property should be enforced. 
For that reason, I consider it unreasonable to regard contraction as an axiom, at least not in general. A similar argument can be used to question the axiomatic status of the intersection property.

\section{Conditioning and Updating}\label{sec:conditioningversusupdating}

Conditioning on events with probability zero is traditionally problematic. Applying Bayes's rule then requires a division by zero, and therefore, the desired conditional probability or expectation is ill-defined. In the context of sets of probabilities, the issue is even more complex, because it can happen that the conditioning event has lower probability zero but positive upper probability.

Throughout this paper, so far, the use of full conditional probability measures has allowed us to stay clear of this issue. Indeed, conditioning on events with probability zero has been non-problematic in our approach. For example, regardless of the lower and/or upper probability of the event $B$, the lower conditional expectations $\underline{E}^{\mathrm{irr}}_G(f(\var{S})\newmid B)$ and $\underline{E}^{\mathrm{com}}_G(f(\var{S})\newmid B)$ are always well-defined. The reason why this is the case is because we do not regard conditioning as a two-step process, where the first step is to build an unconditional joint model and the second step is then to condition this unconditional model---using Bayes's rule---to obtain a conditional one. Instead, the use of full conditional probability measures has enabled us to define conditional models directly.

However, despite the advantages of this approach, there are still two issues left. First of all, the fact that we can define conditional models does not necessarily imply that we can compute them. As the reader may have noticed, most of the inferences that we have considered in Sections~\ref{sec:linearprogramming} and~\ref{sec:theory} were unconditional. Secondly, what we are usually interested in in practice is an updated model, that is, a model that takes into account some observation that we have made. Although these updated models are often identified with  conditional ones, there is no fundamental reason why this should be the case, even more so in the presence of probability zero. The computational issue will be addressed further on in Section~\ref{subsec:computingconditionals}. For now, we focus on the second issue, which is the updating problem.

\subsection{Updating by means of Regular Extension}\label{subsec:regular}

Consider a non-empty set $\mathcal{F}_G$ of full conditional probability measures on $\values{G}$, let $S$ be a subset of $G$, and consider some real-valued function $f\in\gambles{S}$ and an event $B\in\mathcal{P}_{\emptyset}(\values{G})$. Suppose now that we observe the event $B$, in the sense that we receive the information that the true value of $X_G$ belongs to $B$. Based on this observation, what should be our lower and upper expected value for $f(X_S)$? We will refer to this problem as the updating problem. At first sight, one might be inclined to think that we should of course use $\underline{E}_G(f(X_S)\newmid B)$ and $\overline{E}_G(f(X_S)\newmid B)$, which are the conditional lower and upper expectations that correspond to $\mathcal{F}_G$. However, as we are about to argue, this is not necessarily true. 

In order to simplify the problem, let us first consider the case of a single full conditional probability measure $P\in\mathcal{F}_G$. If we observe $B$, what then should be our expectation for $f(X_S)$? The most straightforward answer would be to use $\smash{E(f(X_S)\newmid B)}$. However, it is important to realise that by doing so, we are implicitly identifying conditional expectations with updated expectations, that is, we are interpreting them as expectations that should be adopted after observing an event. Despite the fact that it is often taken for granted, this identification is not trivial, nor is it necessary. Nevertheless, provided that the notion of `observing an event' is formalized very carefully, there are in fact some very good arguments for interpreting conditional expectations as updated ones---see for example the work of Shafer~\cite{shafer1985,shafer1982,shafer2003}---and we will therefore follow this approach here as well. 

However, there is one particular case where these arguments do not apply, which is when $P(B)=0$. In that case, after observing $B$, it seems reasonable to conclude that the full conditional probability measure $P$ is---at least partially---incorrect, because it assigns probability zero to an event that we have just observed. Should this lead us to discard $P$ altogether, or should we still adopt the corresponding conditional expectation $\smash{E(f(X_S)\newmid B)}$ as our updated expectation? For example, in the case of a Bayesian network, if we observe the value $\xval{\parents{s}}$ of the parents of a node $s\in G$, but the global model assigns probability zero to them, should we then still adopt the local expectation $E(f(X_s)\newmid\xval{\parents{s}})$ as our updated expectation for $f(X_s)$? Although we think that this indeed makes sense, it seems to us that this is a purely personal choice that is difficult to defend on objective grounds.

The situation becomes even more complicated in the imprecise case, where $B$ might have probability zero according to some $P\in\mathcal{F}_G$, but positive probability according to others. In that case, after observing $B$, one could argue that the models that assign positive probability to $B$ are better than the others, and that the others should therefore be discarded. The resulting updating rule is called \emph{regular extension}~\cite[Appendix~J2]{Walley:1991vk}. It corresponds to the use of the following updated lower expectation:
\begin{equation}\label{def:regular:unfinished}
\underline{R}_G(f(X_S)\newmid B)
\coloneqq
\inf\{E(f(X_S)\newmid B)\colon P\in\mathcal{F}_G, P(B)>0\}.
\end{equation}
If $\underline{P}_G(B)>0$, then $P(B)$ is strictly positive for all $P\in\mathcal{F}_G$, and therefore, in that case, it does not matter whether we use $\underline{R}_G(f(X_S)\newmid B)$ or $\underline{E}_G(f(X_S)\newmid B)$ as our updated lower expectation. However, if $\overline{P}_G(B)>\underline{P}_G(B)=0$, these two updating rules can lead to substantially different lower expectations; see for example~\cite[Appendix~J2]{Walley:1991vk} or the introduction of~\cite{DeBock2015IJAR2}.

If $\overline{P}_G(B)=0$, Equation~\eqref{def:regular:unfinished} is ill-defined because every $P\in\mathcal{F}_G$ then assigns probability zero to $B$. In that case, two options can be considered. The first option is to discard all the models in $\mathcal{F}_G$. No information about $E(f(X_S)\newmid B)$ is then available, and $\underline{R}_G(f(X_S)\newmid B)$ is then chosen to be the lowest possible bound on $E(f(X_S)\newmid B)$, which is equal to $\min\{f(x_S)\colon x_G\in B\}$; see for example References~\cite{Walley:1991vk,Miranda:2009cg}. Alternatively, since all the models in $\mathcal{F}_G$ are in this case `equally bad', we can choose to keep them all, leading us to adopt $\underline{R}_G(f(X_S)\newmid B)=\underline{E}_G(f(X_S)\newmid B)$ as our updated lower expectation. In the context of credal networks, this choice boils down to the following question: if we observe the value $\xval{\parents{s}}$ of the parents of a node $s\in G$, but the global model assigns upper probability zero to them, should we then use $\min f$ as our updated lower expectation for $f(X_s)$, or should we still adopt the local lower expectation $\underline{E}_{s\mid\xval{\parents{s}}}(f)$? We prefer the second option, and we therefore choose to adopt the following definition of regular extension:
\vspace{3pt}
\begin{equation}\label{def:regular}
\underline{R}_G(f(X_S)\newmid B)
\coloneqq
\begin{cases}
\underline{E}_G(f(X_S)\newmid B)
&\text{if $\underline{P}_G(B)>0$}\\
\inf\{E(f(X_S)\newmid B)\colon P\in\mathcal{F}_G, P(B)>0\}
&\text{if $\overline{P}_G(B)>\underline{P}_G(B)=0$}\\
\underline{E}_G(f(X_S)\newmid B)
&\text{if $\overline{P}_G(B)=0$}.\\
\end{cases}
\vspace{5pt}
\end{equation}

This definition can be applied to any non-empty set $\mathcal{F}_G$ of full conditional probability measures on $\values{G}$, and therefore, in particular, to $\mathcal{F}_G^\irr$ and $\mathcal{F}_G^{\mathrm{com}}$. We will denote the resulting conditional lower expectations by $\underline{R}^\irr_G(f(X_S)\newmid B)$ and $\underline{R}^\mathrm{com}_G(f(X_S)\newmid B)$, respectively. Within the context of this paper, we are mainly interested in the operator $\underline{R}^\irr_G$, which we call the \emph{irrelevant regular extension} of a credal network. It satisfies the following convenient result, which is basically a variation on Theorem~\ref{theo:subnetwork}.
\vspace{4pt}

\begin{theorem}\label{theo:subnetwork:regular}
Let $K$ be a closed subset of $G$. Then for any $f\in\gambles{K}$, $\xval{\parents{K}}\in\values{\parents{K}}$, $B_K\in\mathcal{P}_\emptyset(\values{K})$ and $B_{\nonparnondes{K}}\in\mathcal{P}_\emptyset(\values{\nonparnondes{K}})$, we find that
\begin{align*}
\underline{R}_G^\irr\big(f(\var{K})\newmid B_K,\xval{\parents{K}},B_{\nonparnondes{K}}\big)
=
\begin{cases}
\underline{E}_{K\vert\xval{\parents{K}}}^\irr\big(f(\var{K})\newmid B_K\big)
&\text{~~if $\overline{P}^\irr_{\nondes{K}}(\xval{\parents{K}},B_{\nonparnondes{K}})>0$}\\
\underline{R}_{K\vert\xval{\parents{K}}}^\irr\big(f(\var{K})\newmid B_K\big)
&\text{~~if $\overline{P}^\irr_{\nondes{K}}(\xval{\parents{K}},B_{\nonparnondes{K}})=0$}.\\[-8pt]
\end{cases}
\end{align*}
\vspace{4pt}
\end{theorem}

If the conditioning event $B_K$ is trivial, in the sense that it is equal to $\values{K}$ and can therefore be discarded, we obtain the following special case.
\vspace{3pt}

\begin{corollary}\label{corol:subnetwork:regular}
Let $K$ be a closed subset $G$. Then for any $\xval{\parents{K}}\in\values{\parents{K}}$ and $B_{\nonparnondes{K}}\in\mathcal{P}_\emptyset(\values{\nonparnondes{K}})$, we have that
\begin{equation*}
\underline{R}_G^\irr\big(f(\var{K})\newmid \xval{\parents{K}},B_{\nonparnondes{K}}\big)
=
\underline{E}_{K\vert\xval{\parents{K}}}^\irr\big(f(\var{K})\big)
\end{equation*}
for all $f\in\gambles{K}$.
\vspace{4pt}
\end{corollary}

This result also allows us to corroborate our earlier claim that Equation~\eqref{def:regular} preserves the local models of a credal network. This can be seen by choosing $K\coloneqq\{s\}$, with $B_{\nonparnondes{K}}=\values{\nonparnondes{K}}$ or $B_{\nonparnondes{K}}=\{\xval{\nonparnondes{K}}\}$. Indeed, for these two particular choices, we find that
\begin{equation}\label{eq:regularlocal}
\underline{R}^{\mathrm{irr}}_G(f(X_s)\newmid\xval{\nondes{s}})
=
\underline{R}^{\mathrm{irr}}_G(f(X_s)\newmid\xval{\parents{s}})
=\underline{E}^\irr_{s\vert\xval{\parents{s}}}(f(X_s))
=\underline{E}_{s\vert\xval{\parents{s}}}(f)
~~\text{for all $f\in\gambles{s},$}
\vspace{3pt}
\end{equation}
where the final equality follows from Equation~\eqref{eq:trivialsubnetwork}. 
By comparing this result with Equation~\eqref{eq:assessmentsusinglowerexps}, we see that the irrelevant regular extension of a credal network satisfies the same assessments as its irrelevant natural extension. 

Both extensions are however not equal. In general, the irrelevant regular extension of a credal network is more informative than the irrelevant natural extension, in the sense that
\begin{equation*}
\underline{R}^\irr_G(f(X_G)\newmid B)\geq\underline{E}^\irr_G(f(X_G)\newmid B)
\text{~~for all $f\in\gambles{G}$ and $B\in\mathcal{P}_\emptyset(\values{G})$},
\end{equation*}
and it therefore provides tighter bounds on the unknown expectation $E_G(f(X_G)\newmid B)$. Whether or not the extra assumptions that are required to obtain these tighter bounds are warranted is a matter of personal preference. We think that they are, and that updating by means of regular extension should therefore be preferred over the use of the natural extension. Furthermore, as we will see, the regular extension is also easier to compute. Nevertheless, in the remainder of this paper, we will consider both approaches.

With that, we leave this debate on the difference between updating and conditioning for what it is,  and move on to the more practical issue of how to compute updated lower expectations. More information on the interplay between updating, conditioning and lower probability zero can be found in References~\cite{Zaffalon:2013fk} and~\cite{DeBock2015IJAR2}; the latter also provides a rather extensive philosophical justification for the use of natural and regular extension as updating rules.

\subsection{Computing Conditional Lower Expectations}\label{subsec:computingconditionals}
So far, as the reader may have noticed, we have computed only very few conditional lower expectations: the results in Section~\ref{sec:linearprogramming} and the majority of the examples in Section~\ref{sec:theory} all consider the problem of computing an unconditional lower expectation. 
The only exceptions are the inference problems in Examples~\ref{exmp:unconditionalsubnetwork} and~\ref{exmp:cheesenatural}. However, these examples could be regarded as cheating because, in both cases, we used Theorem~\ref{theo:subnetwork} to reduce the conditional lower expectation of interest to an unconditional lower expectation in a sub-network. This is no coincidence. In fact, the methods that we have presented so far are not able to compute conditional lower expectations that are truly conditional, in the sense that it is not possible to reduce them to an unconditional inference problem in a sub-network. In order to compute such `truly' conditional lower expectations, we require some additional machinery.

As before, let $\mathcal{F}_G$ be a non-empty set of full conditional probability measures on $\values{G}$ and let $\underline{E}_G$ be the corresponding lower expectation operator. With any function $f\in\gambles{G}$ and any event $B\in\mathcal{P}_{\emptyset}(\values{G})$, we can then associate a real-valued function $\rho_{f,B}$, defined by
\begin{equation*}
\rho_{f,B}(\mu)
\coloneqq
\underline{E}_G\big(\ind{B}(X_G)[f(X_G)-\mu]\big)
\text{~~for all $\mu\in\reals$.}
\end{equation*}
In order to evaluate this function, we only need to know the unconditional part of $\underline{E}_G$. Nevertheless, rather remarkably, we can use this function to compute conditional inferences, in the following way.

The simplest case is when $\underline{P}_G(B)>0$. In that case, it can be shown that $\rho_{f,B}$ is a (Lipschitz) continuous strictly decreasing concave function of $\mu$, the unique root of which is equal to $\underline{E}_G(f(X_G)\newmid B)$~\cite{DeBock2015IJAR2}. Furthermore, since we know from Equation~\eqref{def:regular} that $\underline{R}_G(f(X_G)\newmid B)$ and $\underline{E}_G(f(X_G)\newmid B)$ coincide in this case, this unique root will then also be equal to $\underline{R}_G(f(X_G)\newmid B)$. Because $\rho_{f,B}$ is strictly decreasing, finding this root is relatively easy: since $\underline{E}_G(f(X_G)\newmid B)$ must belong to the interval $[\min f,\max f]$, we can use a simple root-finding procedure such as the bisection method. References~\cite[Section~6.3]{deCooman:2010gd} and \cite[page 18]{Walley:1996ht} provide more efficient methods, which take advantage of the specific properties of $\rho_{f,B}$. All of these approaches date back to the work of Lavine~\cite{Lavine1991}, and this method is therefore sometimes referred to as Lavine's bracketing algorithm.

If $\underline{P}_G(B)=0$, the situation becomes slightly more complicated. The function $\rho_{f,B}$ is again (Lipschitz) continuous and concave, but it is no longer strictly decreasing and does not have a unique root.
We distinguish between two cases: $\overline{P}_G(B)>0$ and $\overline{P}_G(B)=0$. 

If $\overline{P}_G(B)>0$, one can show that $\rho_{f,B}$ is equal to zero on the interval $(-\infty,\mu_{f,B}^*]$ and strictly decreasing on $[\mu_{f,B}^*,+\infty)$, where
\begin{equation*}
\mu_{f,B}^*\coloneqq\max\{\mu\in\reals
\colon
\rho_{f,B}(\mu)\geq0
\}.
\vspace{8pt}
\end{equation*}
The rightmost root $\mu_{f,B}^*$ now takes on the role of the unique root in the previous case, in the sense that it is equal to $\underline{R}_G(f(X_G)\newmid B)$. However, unlike in the previous case, it is not guaranteed to coincide with $\underline{E}_G(f(X_G)\newmid B)$, which can now no longer be inferred from $\rho_{f,B}$. Similarly to what happened in the previous case, finding the rightmost root $\mu_{f,B}^*$ is a matter of applying an appropriate bracketing algorithm. However, care should now be taken in order not to introduce numerical errors, especially if $\rho_{f,B}$ decreases slowly on $[\mu_{f,B}^*,+\infty)$; see Reference~\cite[Section 3.3]{DeBock2015IJAR2} for a more detailed discussion.

Finally, if $\overline{P}_G(B)=0$, then $\rho_{f,B}(\mu)=0$ for all $\mu\in\reals$, in which case $\rho_{f,B}$  cannot be used to infer information about $\underline{E}_G(f(X_G)\newmid B)$ or $\underline{R}_G(f(X_G)\newmid B)$.

In summary then, we conclude that if the conditional event $B$ has strictly positive upper probability $\overline{P}_G(B)$, we can use the function $\rho_{f,B}$ to compute $\underline{R}_G(f(X_G)\newmid B)$. Similarly, if $B$ has strictly positive lower probability $\underline{P}_G(B)$, we can use the function $\rho_{f,B}$ to compute $\underline{E}_G(f(X_G)\newmid B)$. What makes these computational techniques especially interesting is that they are all based on bisection or bracketing methods. Therefore, they allow us to compute $\underline{R}_G(f(X_G)\newmid B)$ and $\underline{E}_G(f(X_G)\newmid B)$ up to any desired accuracy $\epsilon$, using only a finite number of evaluations of $\rho_{f,B}(\mu)$.

Besides enabling us to execute the aforementioned methods for computing conditional inferences, the function $\rho_{f,B}$ also allows us to check whether these methods apply, in the sense that we can use it to decide whether or not $\underline{P}_G(B)$ or $\overline{P}_G(B)$ is strictly positive. On the one hand, for any $\mu<\min f$, we have that $\underline{P}_G(B)>0$ if and only if $\rho_{f,B}(\mu)>0$; on the other hand, for any $\mu>\max f$, we have that $\overline{P}_G(B)>0$ if and only if $\rho_{f,B}(\mu)<0$~\cite[Section~3.3]{DeBock2015IJAR2}.

The crucial aspect about the techniques that we have just discussed is that they allow us to turn any method for computing unconditional inferences into a method for computing conditional ones. In particular, any method that allows us to compute $\rho_{f,B}(\mu)=\underline{E}_G\big(\ind{B}(X_G)[f(X_G)-\mu]\big)$ immediately leads to a method for computing $\underline{R}_G(f(X_G)\newmid B)$ and $\underline{E}_G(f(X_G)\newmid B)$, provided of course that the lower or upper probability of $B$ is strictly positive, respectively.

In the context of credal networks under epistemic irrelevance, these findings imply that algorithmic efforts for computing conditional lower expectations can focus on devising an efficient method for evaluating $\rho_{f,B}^\irr$, defined by
\vspace{1pt}
\begin{equation}\label{eq:rhoirr}
\rho_{f,B}^\irr(\mu)\coloneqq\underline{E}_G^\irr\big(\ind{B}(X_G)[f(X_G)-\mu]\big)
\text{~~for all $\mu\in\reals$}.
\vspace{8pt}
\end{equation}
If we are able to come up with such a method, it suffices to combine it with the techniques described above to compute $\underline{E}^\irr_G(f(X_G)\newmid B)$ and $\underline{R}^\irr_G(f(X_G)\newmid B)$. For the former, it is necessary that $\underline{P}_G^\irr(B)>0$, whereas $\smash{\overline{P}_G^\irr(B)>0}$ is a sufficient condition for the latter. Similar conclusions apply to credal networks under complete independence.

For our present purposes, the most important remaining question is now how to compute $\smash{\rho_{f,B}^\irr(\mu)}$. The answer is surprisingly simple. Since we know from Equation~\eqref{eq:rhoirr} that this is an unconditional inference problem, we can use any of the methods that have been presented earlier on in this paper. First of all, for any $f\in\gambles{G}$ and any $B\in\mathcal{P}_{\emptyset}(\values{G})$, $\rho_{f,B}^\irr(\mu)$ can be computed using the linear programming methods of Section~\ref{sec:linearprogramming}. However, since these methods scale badly, this approach only works for networks that are sufficiently small. For larger networks, we therefore need to resort to recursive methods that are based on the decomposition results in Section~\ref{sec:theory}, possibly combined with a localised application of linear programming to small sub-networks. In the remainder of this section, we illustrate the feasibility of such an approach by means of several examples. 

Our first example considers reverse conditioning in a Markov chain, that is, the computation of the lower expectation of a function of the state at the current time point, conditional on the state value at some future time point.

\begin{example}\label{exmp:backwardMC}
Consider again the Markov chain of length ten that we considered in Example~\ref{exmp:forwardinferenceinMCs}. We are interested in computing $\underline{R}_G^\irr(h(X_1)\newmid x_{10})$, for some $h\in\gambles{1}$ and $x_{10}\in\values{10}$. If $\smash{\overline{P}_G^\irr(x_{10})}$ is strictly positive, then as we know from the discussion above, the only thing that is required to achieve this is an efficient method for computing
\vspace{4pt}
\begin{equation*}
\rho_{h,x_{10}}^\irr(\mu)
\coloneqq
\underline{E}_G^\irr\big(\ind{x_{10}}(X_{10})[h(X_1)-\mu]\big).
\vspace{9pt}
\end{equation*}
In the remainder of this example, we will derive such a method.

We start by introducing three sequences of functions. For $k=10$, we let $\underline{h}_{10}\coloneqq\overline{h}_{10}\coloneqq\ind{x_{10}}$. Next, for all $k\in\{1,\dots,10\}$, we let $\underline{h}_k\coloneqq\underline{T}_{k+1}\underline{h}_{k+1}$ and $\overline{h}_{k}\coloneqq-\underline{T}_{k+1}(-\underline{h}_{k+1})$, with $\underline{T}_{k+1}$ as in Example~\ref{exmp:forwardinferenceinMCs}. Finally, for all $k\in\{1,\dots,n\}$, we let $g_k(X_1,X_k)$ be defined by
\vspace{4pt}
\begin{equation}\label{eq:weirdgfunction}
g_k(x_1,x_k)
\coloneqq
\begin{cases}
\underline{h}_k(x_k)[h(x_1)-\mu]
&\text{~if\/ $h(x_1)\geq\mu$}\\
\overline{h}_k(x_k)[h(x_1)-\mu]
&\text{~if\/ $h(x_1)\leq\mu$}
\end{cases}
\text{~~~for all $x_1\in\values{1}$ and $x_k\in\values{k}$.}
\vspace{6pt}
\end{equation}
For any $k\in\{2,\dots,10\}$, 
$x_1\in\values{1}$ and $x_{k-1}\in\values{k-1}$, we then find that
\vspace{1pt}
\begin{align}
&\underline{E}_{k\mid x_{k-1}}\big(g_{k}(x_1,X_k)\big)
=
\begin{cases}
\underline{E}_{k\mid x_{k-1}}\big(\underline{h}_k(X_k)[h(x_1)-\mu]\big)
&\text{if\/ $h(x_1)\geq\mu$}\\
\underline{E}_{k\mid x_{k-1}}\big(\overline{h}_k(X_{k-1})[h(x_1)-\mu]\big)
&\text{if\/ $h(x_1)\leq\mu$}
\end{cases}\notag\\[4pt]
&\hspace{4cm}=
\begin{cases}
[h(x_1)-\mu]\underline{E}_{k\mid x_{k-1}}\big(\underline{h}_k(X_{k})\big)
&\text{if\/ $h(x_1)\geq\mu$}\\
-[h(x_1)-\mu]\underline{E}_{k\mid x_{k-1}}\big(-\overline{h}_k(X_k)\big)
&\text{if\/ $h(x_1)\leq\mu$}
\end{cases}
~=g_{k-1}(x_1,x_{k-1}).
\label{eq:fromgtog}
\\[-12pt]
\notag
\end{align}
Now let $f(X_G)\coloneqq g_{10}(X_1,X_{10})$ and adopt the notation from Example~\ref{exmp:forwardinferenceinMCs}. It then follows from Equation~\eqref{eq:MarkovGeneral} that
\begin{equation*}
\underline{E}_G^\irr(g_{10}(X_1,X_{10}))
=
\underline{E}_G^\irr(f(X_G))
=
\underline{E}_T^\irr(f_9(X_T)),
\vspace{-5pt}
\end{equation*}
where
\vspace{3pt}
\begin{equation*}
f_9(x_T)\coloneqq\underline{E}_{10\mid x_9}\big(f(x_T,X_{10})\big)=\underline{E}_{10\mid x_9}\big(g_{10}(x_1,X_{10})\big)=g_9(x_1,x_9)
\text{~~for all $x_T\in\values{T}$},
\vspace{4pt}
\end{equation*}
using Equations~\eqref{eq:fromgtog} and~\eqref{eq:weirdgfunction} for the final equality. Hence, we find that $\underline{E}_G^\irr(g_{10}(X_1,X_{10}))=\underline{E}_T^\irr(g_{9}(X_1,X_{9}))$ or equivalently, that
\begin{equation*}
\underline{E}_{\{1,\dots,10\}}^\irr(g_{10}(X_1,X_{10}))=\underline{E}_{\{1,\dots,9\}}^\irr(g_{9}(X_1,X_{9})).\vspace{6pt}
\end{equation*}
By repeating this argument over and over again, we eventually find that
\vspace{3pt}
\begin{equation*}
\underline{E}_{\{1,\dots,10\}}^\irr(g_{10}(X_1,X_{10}))
=
\underline{E}_{\{1,\dots,9\}}^\irr(g_{9}(X_1,X_{9}))
=
\underline{E}_{\{1,\dots,8\}}^\irr(g_{8}(X_1,X_{8}))
=
\cdots
=
\underline{E}_1^\irr(g_{1}(X_1,X_1)).
\vspace{4pt}
\end{equation*}
Hence, if we let $g(X_1)\coloneqq g_1(X_1,X_1)$, it follows that
\vspace{3pt}
\begin{equation*}
\rho_{h,x_{10}}^\irr(\mu)
=
\underline{E}_G^\irr\big(\ind{x_{10}}(X_{10})[g(X_1)-\mu]\big)
=
\underline{E}_{\{1,\dots,10\}}^\irr(g_{10}(X_1,X_{10}))
=
\underline{E}_1^\irr(g(X_1)),
\vspace{3pt}
\end{equation*}
which, when combined with Equation~\eqref{eq:trivialsubnetwork}, implies that
\vspace{-5pt}
\begin{equation}\label{eq:rhobackwards}
\rho_{h,x_{10}}^\irr(\mu)
=\underline{E}_1(g)
\text{ where, for all $x_1\in\values{1}$, }
g(x_1)\coloneqq
\begin{cases}
\underline{h}_1(x_1)[h(x_1)-\mu]
&\text{~if $h(x_1)\geq\mu$}\\
\overline{h}_1(x_1)[h(x_1)-\mu]
&\text{~if $h(x_1)\leq\mu$.}
\end{cases}
\vspace{4pt}
\end{equation}
What is important about this expression is that it implies that $\smash{\rho_{h,x_{10}}^\irr(\mu)}$ can be calculated very easily. We simply have to compute $\underline{h}_k$ and $\overline{h}_k$ recursively, starting from $k=10$ and moving backwards to $k=1$, and then, at the end, apply Equation~\eqref{eq:rhobackwards}. All of this requires only local computations, that is, evaluations of the local lower expectations of the model.

Using this recursive procedure, we can now compute $\smash{\rho_{h,x_{10}}^\irr(\mu)}$ for some $\mu>\max f$. If we find that $\smash{\rho_{h,x_{10}}^\irr(\mu)}$ is strictly negative, then as explained in the main text, we know that $\smash{\overline{P}_G^\irr(x_10)>0}$ and we can therefore use bracketing methods to compute $\underline{R}_G^\irr(h(X_1)\newmid x_{10})$ up to any desired accuracy, using only a finite---yet in some cases possible large---number of evaluations of $\smash{\rho_{h,x_{10}}^\irr(\mu)}$.
\hfill$\lozenge$
\end{example}

As a second example, we revisit the inference problems in Examples~\ref{exmp:HMMforwardunconditional} and~\ref{exmp:historyHMMforwardunconditional}. As the reader may recall, at that point, we were not yet able to completely solve these problems. At this point, however, using the results in this section, solving these problems does become feasible.

\begin{example}\label{exmp:HMMforward}
Consider again the inference problems in Examples~\ref{exmp:HMMforwardunconditional} and~\ref{exmp:historyHMMforwardunconditional}, which consisted in computing conditional lower expectations of the form $\underline{E}_G^\irr(f(X_{s_{n+1}})\newmid\xval{o_1},\dots,\xval{o_n})$. In Example~\ref{exmp:HMMforwardunconditional}, the network was an imprecise hidden Markov model, whereas in Example~\ref{exmp:historyHMMforwardunconditional}, we considered an imprecise second order hidden Markov model. In both of these examples, rather than solving this inference problem, we instead focussed on the simpler problem of computing inferences of the form $\underline{E}_G^\irr\big(\ind{x_{o_1}}(X_{o_1})\cdots\ind{x_{o_n}}(X_{o_n}) f(X_{s_{n+1}})\big)$ and claimed that this would  help us to compute the original inference problem. It should now be clear why this is indeed the case.

The clue is that in order to use the techniques in this section to compute $\smash{\underline{E}_G^\irr(f(X_{s_{n+1}})\newmid\xval{o_1},\dots,\xval{o_n})}$ or $\smash{\underline{R}_G^\irr(f(X_{s_{n+1}})\newmid\xval{o_1},\dots,\xval{o_n})}$, all that we require is a method for evaluating
\vspace{2pt}
\begin{align*}
\rho^\irr_{f,x_{o_1},\cdots,x_{o_n}}(\mu)
\coloneqq&
\underline{E}_G^\irr\big(\ind{x_{o_1}}(X_{o_1})\cdots\ind{x_{o_n}}(X_{o_n})\big[f(X_{s_{n+1}})-\mu\big]\big)
=\underline{E}_G^\irr\big(\ind{x_{o_1}}(X_{o_1})\cdots\ind{x_{o_n}}(X_{o_n}) f_\mu(X_{s_{n+1}})\big)\\[-13pt]
\end{align*}
for different values of $\mu\in\reals$, where we let $f_\mu\coloneqq f-\mu$. This inference problem is exactly of the type for which we provide a solution in Examples~\ref{exmp:HMMforwardunconditional} and~\ref{exmp:historyHMMforwardunconditional}; it suffices to replace $f$ by $f_\mu$. The methods that were presented there can therefore be used to compute $\smash{\rho^\irr_{f,x_{o_1},\cdots,x_{o_n}}}$, which can then in turn be used to compute the conditional lower expectations that we are after.

If $\underline{P}^\irr_G(\xval{o_1},\dots,\xval{o_n})$ is strictly positive, then as we know from Equation~\eqref{def:regular}, it does not matter whether we update by means of natural or regular extension. Furthermore, in that case, for the imprecise hidden Markov model, and for the particular inference problem that we are considering here, it also does not matter whether we consider a credal network under epistemic irrelevance or complete independence~\cite[Theorem~3]{Maua:2014ti}. For imprecise second-order hidden Markov models though, no such equivalence is known to hold.
\hfill$\lozenge$
\end{example}

The techniques that we have employed in Example~\ref{exmp:backwardMC} and~\ref{exmp:HMMforward} are very similar to the ones that were used in Reference~\cite{deCooman:2010gd} to develop algorithms for credal trees under epistemic irrelevance. Essentially, the main idea is to first use Equation~\eqref{eq:rhoirr} to turn the conditional inference problem into an unconditional one, and then to solve the remaining unconditional inference problem by a recursive decomposition scheme. In fact, the algorithms in Reference~\cite{deCooman:2010gd} can all be derived from our results. The converse is however not true, for the following three reasons.

Firstly, our decomposition properties do not require any positivity conditions, whereas the results in Reference~\cite{deCooman:2010gd} only apply to credal networks of which the local upper probabilities are strictly positive. Secondly, our methods are not restricted to credal networks of which the graphical structure is a tree: consider for example the inferences in Examples~\ref{exmp:unconditionalsubnetwork} or~\ref{exmp:twonodeswithcommonchild}, the dynamical network in Example~\ref{exmp:cheesenatural}, the second order hidden Markov model in Examples~\ref{exmp:historyHMMforwardunconditional} and~\ref{exmp:HMMforward}, the factorisation results in Example~\ref{exmp:atomfactorisation}, as well as the examples further on in this section.
Thirdly, for conditional lower expectations, the algorithms in Reference~\cite{deCooman:2010gd} are restricted to functions that depend on a single variable. As can be seen from Example~\ref{exmp:cheesenatural}, our methods do not require such a restriction. The following simple example provides a second illustration that this is indeed not required.

\begin{example}\label{exmp:twonodeswithcommonchild:conditional}
Let $G\coloneqq\{1,2,3\}$, consider a credal network of which the graph is depicted in Figure~\ref{fig:DAGreversetree}, and assume that we are interested in computing $\underline{R}_G^\irr(g(X_1,X_2)\newmid\xval{3})$ or $\underline{E}_G^\irr(g(X_1,X_2)\newmid\xval{3})$, for some $g\in\gambles{\{1,2\}}$ and $\xval{3}\in\values{3}$. In order to do that by means of the techniques in this section, all we have to do is come up with a method for computing
\begin{equation*}
\rho_{g,\xval{3}}^\irr(\mu)\coloneqq\underline{E}_G^\irr\big(\ind{\xval{3}}(X_3)[g(X_1,X_2)-\mu]\big).
\end{equation*}
However, we already know from Example~\ref{exmp:twonodeswithcommonchild} how to do this. If we let $f(X_G)\coloneqq\ind{\xval{3}}(X_3)[g(X_1,X_2)-\mu]$, it follows from the discussion in Example~\ref{exmp:twonodeswithcommonchild} that
\vspace{3pt}
\begin{equation*}
\rho_{g,\xval{3}}^\irr(\mu)
=\underline{E}_G^\irr\big(\ind{\xval{3}}(X_3)[g(X_1,X_2)-\mu]\big)
=
\underline{E}^\irr_{G}(f(X_G))
=
\underline{E}^\irr_{\{1,2\}}(
h_\mu(X_1,X_2)
),\vspace{5pt}
\end{equation*}
where, for all $x_1\in\values{1}$ and $x_2\in\values{2}$,
\vspace{-4pt}
\begin{equation*}
h_\mu(x_1,x_2)\coloneqq
\underline{E}^\irr_{3\mid(x_1,x_2)}(
f(x_1,x_2,X_3)
)
=
\begin{cases}
[g(x_1,x_2)-\mu]
\underline{P}_{3\mid(x_1,x_2)}(x_3)
&\text{if $g(x_1,x_2)\geq\mu$}\\
[g(x_1,x_2)-\mu]
\overline{P}_{3\mid(x_1,x_2)}(x_3)
&\text{if $g(x_1,x_2)\leq\mu$}.
\end{cases}
\vspace{5pt}
\end{equation*}
The computation of $h_\mu$ is trivial because it only requires the evaluation of local lower and upper probabilities. Therefore, the problem has been reduced to that of computing a lower expectation $\underline{E}^\irr_{\{1,2\}}(
h_\mu(X_1,X_2)$ in a credal network that consists of two disconnected nodes. This reduced problem can easily be solved by means of the linear programming techniques in Section~\ref{sec:linearprogramming}.
\hfill$\lozenge$
\end{example}

In this example, as well as in Example~\ref{exmp:cheesenatural}, the function for which we are computing the conditional lower expectation depends on only two variables. Since this is only a minor step beyond the case of a single variable, one could interpret this as a sign that these methods scale badly to situations where the function of interest depends on many variables. This is not the case though. Our reason for providing these rather simple examples is because they are relatively easy to explain, thereby allowing us to focus on conveying intuition. However, they are by no means representative for the frontier of the types of conditional lower expectations that can be computed. For example, for an imprecise hidden Markov model, similar techniques can also be used to estimate a sequence of hidden states based on the corresponding sequence of observation variables~\cite{DeBock:2014ts}, which requires computing conditional lower expectations of functions that depend on all state variables.

In any case, as illustrated in our examples so far, the function in Equation~\eqref{eq:rhoirr} is an extremely useful tool when it comes to the computation of conditional lower expectations. However, this should not be taken to mean that the first step in the computation of a conditional lower expectation should always be to apply Equation~\eqref{eq:rhoirr}. Indeed, it is often possible to first reduce the size of the problem---using Theorem~\ref{theo:subnetwork}, Theorem~\ref{theo:subnetwork:regular} or Corollary~\ref{corol:subnetwork:regular}---and to only afterwards apply Equation~\eqref{eq:rhoirr} to a sub-network.
In fact, in some cases, as we have seen in Examples~\ref{exmp:unconditionalsubnetwork} and~\ref{exmp:cheesenatural}, it is possible to turn a conditional inference problem into an unconditional inference problem in a sub-network, thereby dispensing with the need for Equation~\eqref{eq:rhoirr} altogether. Our next example illustrates why, in the case of Example~\ref{exmp:unconditionalsubnetwork}, such an approach is indeed to be preferred over a direct application of Equation~\eqref{eq:rhoirr}.

\begin{example}\label{exmp:regularequaltonatural}
Consider once more the inference problem of Example~\ref{exmp:unconditionalsubnetwork}, which consisted in computing a conditional lower expectation of the form $\underline{E}_G^\irr(h(X_9)\newmid x_3,x_4,x_6)$ for a credal network whose graph is depicted in Figure~\ref{fig:DAG}. By applying Theorem~\ref{theo:subnetwork}, for $K=\{5,7,9\}$, $B_K=\values{K}$ and $B_{N(K)}=(X_6=x_6)$, it was shown there that
\begin{equation}\label{eq:exmp:regularequaltonatural:1}
\underline{E}_G^\irr(h(X_9)\newmid x_3,x_4,x_6)
=
\underline{E}_{K\mid\xval{\parents{K}}}^\irr(h(X_9)),
\vspace{6pt}
\end{equation}
thereby transforming the problem into an unconditional inference problem in a smaller network that has only three nodes, which could then be computed using the recursive formulas that were derived in Example~\ref{exmp:forwardinferenceinMCs}.

Given the results in this section, we could of course also try to compute $\underline{E}_G^\irr(h(X_9)\newmid x_3,x_4,x_6)$ by applying a bracketing procedure to the function $\rho_{h,x_3,x_4,x_6}$, defined by
\vspace{1pt}
\begin{equation}\label{eq:rhoforregularnatural}
\rho_{h,x_3,x_4,x_6}(\mu)
\coloneqq
\underline{E}_G^\irr\big(\ind{x_3}(X_3)\ind{x_4}(X_4)\ind{x_6}(X_6)[h(X_9)-\mu]\big)
\text{~~for all $\mu\in\reals$.}
\vspace{2pt}
\end{equation}
However, this would not be as efficient as using Equation~\eqref{eq:exmp:regularequaltonatural:1}, for the following two reasons. First of all, it would require us to come up with a method for computing $\rho_{h,x_3,x_4,x_6}(\mu)$. This is far from easy. The linear programming methods in Section~\ref{sec:linearprogramming} would most likely run into scaling problems here, and although we might be able to partially decompose the problem by applying Corollary~\ref{corol:sub-DAG:factorisation:LP}, the resulting subproblems could still be non-trivial to compute. Secondly, in order for this approach to be applicable, the lower probability $\smash{\underline{P}_G^\irr(x_3,x_4,x_6)}$ must be strictly positive, which may not be the case. Both of these problems can be avoided by using Equation~\eqref{eq:exmp:regularequaltonatural:1} instead.

Similar observations can be made if we update by means up regular extension instead of natural extension. The only difference is that the role of Theorem~\ref{theo:subnetwork} is replaced by Corollary~\ref{corol:subnetwork:regular}. In particular, if we apply Corollary~\ref{corol:subnetwork:regular} for $K=\{5,7,9\}$, $B_K=\values{K}$ and $B_{N(K)}=(X_6=x_6)$, we find that
\vspace{1pt}
\begin{equation*}
\underline{R}_G^\irr(h(X_9)\newmid x_3,x_4,x_6)
=
\underline{E}_{K\mid\xval{\parents{K}}}^\irr(h(X_9)).
\vspace{2pt}
\end{equation*}
By comparing this result with Equation~\eqref{eq:exmp:regularequaltonatural:1}, we see that in this example, despite the fact that we are applying a different updating rule, the result of the inference nevertheless remains the same. Furthermore, as before, the direct use of Equation~\eqref{eq:exmp:regularequaltonatural:1} is to be preferred over an indirect method that is based on Equation~\eqref{eq:rhoforregularnatural}.
\hfill$\lozenge$
\end{example}

Completely analogous conclusions can be drawn for the inference problem in Example~\ref{exmp:cheesenatural}. Again, it makes no difference whether we use the natural or the regular extension, and the use of Theorem~\ref{theo:subnetwork} or Corollary~\ref{corol:subnetwork:regular} is be preferred over an indirect approach that is based on Equation~\eqref{eq:rhoforregularnatural}.

We end with an example where the two approaches are combined, in the sense that we first reduce the size of the problem---using Theorem~\ref{theo:subnetwork} or Corollary~\ref{corol:subnetwork:regular}---and then tackle the remaining problem by means of Equation~\eqref{eq:rhoirr}. The specific problem that we consider is that of computing the lower expectation of a function that depends on a single variable, conditional on the value of all the other variables in the network. Remarkably, our solution applies to credal networks with arbitrary graphs.

\begin{example}\label{exmp:completeevidence}
Consider an arbitrary credal network under epistemic irrelevance, choose any node $q\in G$ and let $E\coloneqq G\setminus\{q\}$ be the set of remaining nodes. We consider the problem of computing a conditional lower expectation of the form $\smash{\underline{E}_G^{\irr}(f(X_q)\newmid\xval{E})}$ or $\underline{R}_G^{\irr}(f(X_q)\newmid\xval{E})$, with $f\in\gambles{q}$ and $\xval{E}\in\values{E}$.

Solving this inference problem is trivial if $q$ is a leaf node. We then have that $\nondes{q}=E$ and it therefore follows directly from Equations~\eqref{eq:assessmentsusinglowerexps} and~\eqref{eq:regularlocal} that
\vspace{2pt}
\begin{equation*}
\underline{E}_G^{\irr}(f(X_q)\rfloor\xval{E})=\underline{R}_G^{\irr}(f(X_q)\rfloor\xval{E})
=\underline{E}_{q\mid\xval{\parents{q}}}(f).
\vspace{2pt}
\end{equation*}
When $q$ is not a leaf node, the problem can be solved as follows.

Let $K\coloneqq\{q\}\cup\descendants{q}$. By applying Theorems~\ref{theo:subnetwork} and~\ref{theo:subnetwork:regular} for  $B_K=\{\xval{\descendants{q}}\}$ and $B_{\nonparnondes{K}}=\{\xval{\nonparnondes{K}}\}$, we find that
\begin{equation*}
\underline{E}_G^{\irr}(f(X_q)\newmid\xval{E})
=\underline{E}_{K\mid\xval{\parents{K}}}^{\irr}(f(X_q)\newmid\xval{\descendants{q}})
\end{equation*}
and
\begin{equation*}
\underline{R}_G^{\irr}(f(X_q)\newmid\xval{E})
=
\begin{cases}
\underline{R}_{K\mid\xval{\parents{K}}}^{\irr}(f(X_q)\newmid\xval{\descendants{q}})
&\text{~~if\, $\overline{P}^\irr_{\nondes{q}}(\xval{\nondes{q}})>0$}\\
\underline{E}_{K\mid\xval{\parents{K}}}^{\irr}(f(X_q)\newmid\xval{\descendants{q}})
&\text{~~if\, $\overline{P}^\irr_{\nondes{q}}(\xval{\nondes{q}})=0$},
\end{cases}
\vspace{2pt}
\end{equation*}
where
\vspace{1pt}
\begin{equation*}
\overline{P}_{\nondes{q}}^\irr(\xval{\nondes{q}})=\prod_{s\in \nondes{q}}\overline{P}_{s\rfloor\xval{\parents{s}}}(\xval{s})
\vspace{6pt}
\end{equation*}
because of Equation~\eqref{eq:impreciselocalfactorisation}.
In this way, we have reduced the original inference problem to a similar but smaller sized inference problem in the sub-network that corresponds to $K$ and $\smash{\xval{\parents{K}}}$. The remaining task is now to compute $\underline{E}_{K\mid\xval{\parents{K}}}^{\irr}(f(X_q)\newmid\xval{\descendants{q}})$ or $\underline{R}_{K\mid\xval{\parents{K}}}^{\irr}(f(X_q)\newmid\xval{\descendants{q}})$.

If the lower probability $\underline{P}^\irr_{K\mid\xval{\parents{K}}}(x_{\descendants{q}})$ or upper probability $\overline{P}^\irr_{K\mid\xval{\parents{K}}}(x_{\descendants{q}})$ is strictly positive, these computations can be performed by applying the techniques in this section to the subnetwork that corresponds to $K$ and $\xval{\parents{K}}$. All that we require is an efficient method for evaluating the real-valued function $\rho_{f,\,\xval{K}\mid\xval{\parents{K}}}^\irr$, defined by
\begin{equation*}
\rho_{f,\,\xval{K}\mid\xval{\parents{K}}}^\irr(\mu)
\coloneqq
\underline{E}_{K\rfloor\xval{\parents{K}}}^{\irr}\big(\ind{\{\xval{\descendants{q}}\}}(X_{\descendants{q}})[f(X_q)-\mu]\big)
\text{~~for all $\mu\in\reals$.}
\vspace{7pt}
\end{equation*}
So let us fix some $\mu\in\reals$. Since $q$ is not a leaf node of the network, we know that $\descendants{q}\neq\emptyset$.
Therefore, if we apply Theorem~\ref{theo:marginalextension} to the sub-network that corresponds to $K$ and $\xval{\parents{K}}$, for $S=\descendants{q}$ and $T=\{q\}$, we find that
\begin{align}
\rho_{f,\,\xval{K}\mid\xval{\parents{K}}}^\irr(\mu)
&=\underline{E}_{K\mid\xval{\parents{K}}}^{\irr}\big(\ind{\{\xval{S}\}}(X_{S})[f(X_q)-\mu]\big)\notag\\
&=
\underline{E}_{T\mid\xval{\parents{T}}}^\irr(g_\mu(X_T))
=
\underline{E}_{q\mid\xval{\parents{q}}}^\irr(g_\mu(X_q))
=\underline{E}_{q\mid\xval{\parents{q}}}(g_\mu),\label{eq:rhoforcompleteevidence}\\[-7pt]\notag
\end{align}
where the final equality follows from Equation~\eqref{eq:trivialsubnetwork} and where $g_\mu$ is a function on $\values{q}$, defined for all $x_q\in\values{q}$ by
\vspace{-8pt}
\begin{align*}
g_\mu(\xval{q})
\coloneqq
\underline{E}_{S\mid\xval{\parents{S}}}^{\irr}\big(\ind{\{\xval{S}\}}(X_{S})[f(x_q)-\mu]\big)
=
\begin{cases}
[f(x_q)-\mu]\underline{P}_{S\mid\xval{\parents{S}}}^{\irr}(\xval{S})
&\text{if $f(x_q)\geq\mu$}\\
[f(x_q)-\mu]\overline{P}_{S\mid\xval{\parents{S}}}^{\irr}(\xval{S})
&\text{if $f(x_q)\leq\mu$.}
\end{cases}\\[-10pt]
\end{align*}
The reason why this function only depends on $x_q$ is because for all other $s\in\parents{S}$, $x_s$ is uniquely determined by $x_E$. Finally, we can simplify the expression for $g_\mu$ even further, by applying Equation~\eqref{eq:impreciselocalfactorisation} to the sub-network that corresponds to $S$ and $\xval{\parents{S}}$. In this way we find that
\begin{equation*}
g_\mu(\xval{q})
=
\begin{cases}
&\\[-10pt]
[f(\xval{q})-\mu]
\displaystyle\prod_{s\in S}
\underline{P}_{s\rfloor\xval{\parents{s}}}(\xval{s})
&\text{~if $f(\xval{q})\geq\mu$}\\
[f(\xval{q})-\mu]
\displaystyle\prod_{s\in S}
\overline{P}_{s\rfloor\xval{\parents{s}}}(\xval{s})
&\text{~if $f(\xval{q})\leq\mu$}
\end{cases}
\text{~~~~for all $\xval{q}\in\values{q}$.}
\vspace{2pt}
\end{equation*}
This expression implies that the computational complexity of evaluating Equation~\eqref{eq:rhoforcompleteevidence} is linear in $\vert S\vert$, or equivalently, in the number of descendants of $q$. For this reason, as explained in this section, a simple bracketing method will allow us to compute $\underline{E}_{K\mid\xval{\parents{K}}}^{\irr}(f(X_q)\newmid\xval{\descendants{q}})$ or $\smash{\underline{R}_{K\mid\xval{\parents{K}}}^{\irr}(f(X_q)\newmid\xval{\descendants{q}})}$, and therefore also $\smash{\underline{E}_G^{\irr}(f(X_q)\newmid\xval{E})}$ or $\smash{\underline{R}_G^{\irr}(f(X_q)\newmid\xval{E})}$.
\hfill$\lozenge$
\end{example}

\section{Conclusions and a Brief Look Beyond the Horizon}\label{sec:conclusion}

The main conclusion of this paper is that credal networks under epistemic irrelevance satisfy surprisingly many powerful theoretical properties, and that these properties can be exploited to develop efficient computational methods, for various types of inference problems that were previously presumed to be intractable. Since many of these inference problems are NP-hard in credal networks under complete or strong independence, these results turn credal networks under epistemic irrelevance into a serious, practically feasible alternative that should enable practitioners to solve real-life problems for which the corresponding necessary inferences were hitherto regarded as intractable.

On the theoretical side, we would like to highlight the following contributions. First of all, we have shown that full conditional probability measures can be used to develop a rigorous yet simple definition for a credal network under epistemic irrelevance and its irrelevant natural and regular extension, without having to impose strict positivity assumptions. Secondly, these extensions were shown to satisfy various properties that can be used to decompose large inference problems into smaller ones, including marginalisation properties for sub-networks, a law of iterated lower expectation, and specific types of factorisation and external additivity. Thirdly, similarly to how d-separation implies independence in a Bayesian network, we have seen that in the irrelevant natural extension of a credal network, AD-separation implies epistemic irrelevance.

From a computational point of view, our most important contribution is perhaps the realisation that inferences in credal networks under epistemic irrelevance can be computed by means of several techniques, and  that all of these techniques can be fruitfully combined with one another. On the one hand, we have shown that the recursive techniques of De Cooman et al.~\cite{deCooman:2010gd} can be extended in two ways: they are not limited to credal trees, nor ar they limited to inferences about a single variable. On the other hand, we have seen that the linear programming methods of Cozman~\cite{Cozman:2000ug} do not require any positivity conditions. Furthermore, by applying these linear programming methods to sub-networks, they can be succesfully combined with the recursive techniques that were mentioned before. Finally, through the use of Lavine's bracketing algorithm, all of these techniques can be extended to the conditional case.

As far as future work is concerned, a first important task is to use the techniques in this paper to develop new efficient algorithms for credal networks epistemic irrelevance, for various types of inference problems.
So far, most existing algorithms are restricted to networks whose graph is a tree, and are fairly limited in the type of inference problems that they consider~\cite{deCooman:2010gd,DeBock:2014ts,Lopatatzidis2015,Benavoli:2011ei}. The only exception seems to be Reference~\citep[Chapter~7]{DeBock2015:thesis}, which considers credal networks with a recursively decomposable graph---a generalisation of a tree.
Furthermore, all existing algorithms are based on decomposition techniques, and none of them apply linear programming methods to sub-networks. 
In contrast, the examples in this paper clearly illustrate that these restrictions are by no means necessary.
Therefore, it seems to us that the search for efficient inference algorithms in credal networks under epistemic irrelevance is wide open, and that significant improvements upon the state of the art should still be possible. Furthermore, since existing algorithms all focus on exact computations, the design of approximate algorithms should definitely be considered as well.

Another important future line of research would be to apply credal networks under epistemic irrelevance to real applications, in situations where the defining assumptions of a Bayesian network---exactly specified probabilities and exact independence assessments---are unrealistic. In principle, this is already feasible now: the examples in this paper and the algorithms in References~\citep{DeBock2015:thesis,deCooman:2010gd} should already allow practitioners to solve large classes of problems that are relevant to their applications. 
However, in practice, two additional steps are needed. First of all, it is necessary to implement existing and/or new algorithms, and to develop user-friendly software to compute with them; no such software currently exists. Secondly, it should be thoroughly tested whether the bounds that are computed by these algorithms are informative enough to be useful in practice. Indeed, since epistemic irrelevance imposes less stringent constraints than complete or strong independence, the bounds of a credal network under epistemic irrelevance can be more conservative than those that correspond to other types of credal networks, and it may therefore happen that they are too conservative to be of practical use. Although this type of behaviour did not occur in the proofs of concept in References~\citep{DeBock:2014ts,Benavoli:2011ei}, it remains to be seen whether this will be the case in other applications as well.

\section*{Acknowledgements}

I am a Postdoctoral Fellow of the Research Foundation - Flanders (FWO) and wish to acknowledge its financial support. I also thank the anonymous reviewers of this paper and of the two conference papers on which it is based~\cite{DeBock2016PGM,DeBock:2013wf}, for their detailed reading and constructive feedback.
The current extended version was written during a research visit of mine to the Imprecise Probability Group of IDSIA, the members of which I would like to thank for their hospitality. 
Finally, I would like to express my great gratitude to Gert de Cooman, my former PhD supervisor, for teaching me the art of mathematics.




\appendix

\section*{References}

\bibliographystyle{plain}


\begin{thebibliography}{10}

\bibitem{Antonucci:2009jm}
Alessandro Antonucci, Ralph Br{\"u}hlmann, Alberto Piatti, and Marco Zaffalon.
\newblock {Credal networks for military identification problems}.
\newblock {\em International Journal of Approximate Reasoning}, 50(4):666--679,
  April 2009.

\bibitem{Antonucci:2014ty}
Alessandro Antonucci, Cassio~P. de~Campos, and Marco Zaffalon.
\newblock {Probabilistic graphical models}.
\newblock In {\em Introduction to Imprecise Probabilities}, pages 207--229.
  Wiley, 2014.

\bibitem{Antonucci:2007wv}
Alessandro Antonucci, Andrea Salvetti, and Marco Zaffalon.
\newblock {Credal networks for hazard assessment of debris flows}.
\newblock In J{\"u}rgen~P. Kropp and J{\"u}rgen Scheffran, editors, {\em
  Advanced Methods for Decision Making and Risk Management in Sustainability
  Science}, pages 237--256. NOVA Science Publishers, New York, 2007.

\bibitem{Baudrit:2010cc}
C{\'e}dric Baudrit, Mariette Sicard, Pierre-Henri Wuillemin, and Nathalie
  Perrot.
\newblock {Towards a global modelling of the Camembert-type cheese ripening
  process by coupling heterogeneous knowledge with dynamic Bayesian networks}.
\newblock {\em Journal of Food Engineering}, 98(3):283--293, June 2010.

\bibitem{Benavoli:2011ei}
Alessio Benavoli, Marco Zaffalon, and Enrique Miranda.
\newblock {Robust filtering through coherent lower previsions}.
\newblock {\em IEEE Transactions on Automatic Control}, 56(7):1567--1581, 2011.

\bibitem{berger1985}
{James O.} Berger.
\newblock {\em Statistical decision theory and Bayesian analysis}.
\newblock Springer series in statistics. Springer, New York, NY, 1985.

\bibitem{Bernard:2005hd}
Jean-Marc Bernard.
\newblock {An introduction to the imprecise Dirichlet model for multinomial
  data}.
\newblock {\em International Journal of Approximate Reasoning},
  39(2-3):123--150, June 2005.

\bibitem{DeBock2015IJAR2}
Jasper~De Bock and Gert de~Cooman.
\newblock Conditioning, updating and lower probability zero.
\newblock {\em International Journal of Approximate Reasoning}, 67:1--36,
  2015.

\bibitem{Couso:1999wh}
In{\'e}s Couso, Seraf{\'\i}n Moral, and Peter Walley.
\newblock {Examples of Independence for Imprecise Probabilities}.
\newblock In Gert de~Cooman, Fabio~G. Cozman, Seraf{\'\i}n Moral, and Peter
  Walley, editors, {\em ISIPTA '99: Proceedings of the First International
  Symposium on Imprecise Probabilities and their Applications}, pages 121--130.
  Ghent, 1999.

\bibitem{Cozman:1998:IIR:2074094.2074105}
Fabio~G. Cozman.
\newblock Irrelevance and independence relations in quasi-bayesian networks.
\newblock In {\em Proceedings of UAI `98}, pages 89--96, San Francisco, 1998.

\bibitem{Cozman:2000ug}
Fabio~G. Cozman.
\newblock {Credal networks}.
\newblock {\em Artificial Intelligence}, 120(2):199--233, 2000.

\bibitem{Cozman:2005ct}
Fabio~G. Cozman.
\newblock {Graphical models for imprecise probabilities}.
\newblock {\em International Journal of Approximate Reasoning}, 39(2):167--184,
  2005.

\bibitem{Cozman:2012fc}
Fabio~G. Cozman.
\newblock {Sets of probability distributions, independence, and convexity}.
\newblock {\em Synthese}, 186(2):577--600, May 2012.

\bibitem{Cozman:2013us}
Fabio~G. Cozman.
\newblock {Independence for sets of full conditional probabilities, sets of
  lexicographic probabilities, and sets of desirable gambles}.
\newblock In {\em Proceedings of ISIPTA `13}, pages 87--97. Compi{\`e}gne,
  2013.

\bibitem{Cozman:2005fc}
Fabio~G. Cozman and Peter Walley.
\newblock {Graphoid properties of epistemic irrelevance and independence}.
\newblock {\em Annals of Mathematics and Artificial Intelligence},
  45(1-2):173--195, 2005.

\bibitem{Dawid:1997je}
A.~Philip Dawid.
\newblock {Conditional independence}.
\newblock In {\em Encyclopedia of Statistical Sciences}, pages 146--155.
  Wiley-Interscience, 1997.

\bibitem{DeBock2015:thesis}
Jasper De~Bock.
\newblock {\em Credal networks under epistemic irrelevance: theory and
  algorithms}.
\newblock PhD thesis, 2015.

\bibitem{DeBock2016PGM}
Jasper De~Bock.
\newblock Reintroducing credal networks under epistemic irrelevance.
\newblock In {\em PGM}, volume~52 of {\em JMLR: Workshop and Conference
  Proceedings}, pages 123--135, 2016.

\bibitem{DeBock:2013wf}
Jasper De~Bock and Gert de~Cooman.
\newblock {Allowing for probability zero in credal networks under epistemic
  irrelevance}.
\newblock In Fabio~G. Cozman, Thierry Den{\oe}ux, S{\'e}bastien Destercke, and
  Teddy Seidenfeld, editors, {\em ISIPTA '13: Proceedings of the Eighth
  International Symposium on Imprecise Probability: Theory and Applications},
  pages 109--118. SIPTA, Compi{\`e}gne, 2013.

\bibitem{DeBock:2013tt}
Jasper De~Bock and Gert de~Cooman.
\newblock {Credal networks under epistemic irrelevance using sets of desirable
  gambles}.
\newblock In {\em Proceedings of ISIPTA `13}, pages 99--108. SIPTA,
  Compi{\`e}gne, 2013.

\bibitem{DeBock:2014ts}
Jasper De~Bock and Gert de~Cooman.
\newblock {An efficient algorithm for estimating state sequences in imprecise
  hidden Markov models}.
\newblock {\em Journal of Artificial Intelligence Research}, 50:189--233, 2014.

\bibitem{DeBock:2014bv}
Jasper De~Bock and Gert de~Cooman.
\newblock {Credal networks under epistemic irrelevance: the sets of desirable
  gambles approach}.
\newblock {\em International Journal of Approximate Reasoning}, 56:178--207,
  2015.

\bibitem{deCampos:2007kg}
Cassio~P. de~Campos and Fabio~G. Cozman.
\newblock {Computing lower and upper expectations under epistemic
  independence}.
\newblock {\em International Journal of Approximate Reasoning}, 44(3):244--260,
  2007.

\bibitem{deCooman:2010gd}
Gert de~Cooman, Filip Hermans, Alessandro Antonucci, and Marco Zaffalon.
\newblock {Epistemic irrelevance in credal nets: the case of imprecise Markov
  trees}.
\newblock {\em International Journal of Approximate Reasoning},
  51(9):1029--1052, 2010.

\bibitem{deCooman:2009jz}
Gert de~Cooman, Filip Hermans, and Erik Quaeghebeur.
\newblock {Imprecise Markov chains and their limit behaviour}.
\newblock {\em Probability in the Engineering and Informational Sciences},
  23(4):597--635, October 2009.

\bibitem{deCooman:2011ey}
Gert de~Cooman, Enrique Miranda, and Marco Zaffalon.
\newblock {Independent natural extension}.
\newblock {\em Artificial Intelligence}, 175(12):1911--1950, 2011.

\bibitem{Dubins:1975ej}
Lester~E. Dubins.
\newblock {Finitely additive conditional probabilities, conglomerability and
  disintegrations}.
\newblock {\em The Annals of Probability}, 3(1):89--99, 1975.

\bibitem{Fagiuoli:1998ft}
Enrico Fagiuoli and Marco Zaffalon.
\newblock {2U: an exact interval propagation algorithm for polytrees with
  binary variables}.
\newblock {\em Artificial Intelligence}, 106(1):77--107, 1998.

\bibitem{Grunbaum:2003vq}
Branko Gr{\"u}nbaum.
\newblock {\em {Convex polytopes}}.
\newblock Springer Science {\&} Business Media, second edition, May 2003.

\bibitem{Hourbracq:2013un}
Matthieu Hourbracq, C{\'e}dric Baudrit, Pierre-Henri Wuillemin, and
  S{\'e}bastien Destercke.
\newblock {Dynamic credal networks: introduction and use in robustness
  analysis}.
\newblock In Fabio~G. Cozman, Thierry Den{\oe}ux, S{\'e}bastien Destercke, and
  Teddy Seidenfeld, editors, {\em ISIPTA '13: Proceedings of the Eighth
  International Symposium on Imprecise Probability: Theory and Applications},
  pages 159--169. SIPTA, Compi{\`e}gne, July 2013.

\bibitem{Koller:2009ty}
Daphne Koller and Nir Friedman.
\newblock {\em {Probabilistic graphical models: principles and techniques}}.
\newblock MIT Press, 2009.

\bibitem{Lavine1991}
Michael Lavine, Larry Wasserman, and Robert~L. Wolpert.
\newblock Bayesian inference with specified prior marginals.
\newblock {\em Journal of the American Statistical Association},
  86(416):964--971, 1991.

\bibitem{Lopatatzidis2015}
Stavros Lopatatzidis, Jasper Bock, Gert Cooman, Stijn Vuyst, and Joris
  Walraevens.
\newblock Robust queueing theory: an initial study using imprecise
  probabilities.
\newblock {\em Queueing Systems}, 82(1):75--101, 2015.

\bibitem{Maua:2014ti}
Denis~D. Mau{\'a}, Cassio~P. de~Campos, Alessio Benavoli, and Alessandro
  Antonucci.
\newblock Probabilistic inference in credal networks: new complexity results.
\newblock {\em Journal of Artificial Intelligence Research}, 50:603--637, 2014.

\bibitem{Miranda:2008fo}
Enrique Miranda.
\newblock {A survey of the theory of coherent lower previsions}.
\newblock {\em International Journal of Approximate Reasoning}, 48(2):628--658,
  2008.

\bibitem{Miranda:2009cg}
Enrique Miranda.
\newblock {Updating coherent previsions on finite spaces}.
\newblock {\em Fuzzy Sets and Systems}, 160(9):1286--1307, May 2009.

\bibitem{Moral:2005ce}
Seraf{\'\i}n Moral.
\newblock {Epistemic irrelevance on sets of desirable gambles}.
\newblock {\em Annals of Mathematics and Artificial Intelligence},
  45(1-2):197--214, 2005.

\bibitem{Pearl:1988wk}
Judea Pearl.
\newblock {\em {Probabilistic reasoning in intelligent systems: networks of
  plausible inference}}.
\newblock Morgan Kaufmann, San Mateo, 1988.

\bibitem{Piatti:2010vd}
Alberto Piatti, Alessandro Antonucci, and Marco Zaffalon.
\newblock {Building knowledge-based systems by credal networks: a tutorial}.
\newblock In Albert~R. Baswell, editor, {\em Advances in Mathematics Research,
  Volume 11}, pages 227--279. Nova Science Publishers, 2010.

\bibitem{Quaeghebeur:2014tjb}
Erik Quaeghebeur.
\newblock {Desirability}.
\newblock In Thomas Augustin, Frank P.~A. Coolen, Gert de~Cooman, and Matthias
  C.~M. Troffaes, editors, {\em Introduction to Imprecise Probabilities}, pages
  1--27. John Wiley {\&} Sons, Chichester, May 2014.

\bibitem{seidenfeld2007}
Teddy Seidenfeld.
\newblock Conditional independence, imprecise probabilities, null-events and
  graph-theoretic models. seimour geisser distinguished lecture, school of
  statistics, university of minnesota, minneapolis, {MN}, 2007.

\bibitem{shafer1982}
Glenn Shafer.
\newblock Bayes's two arguments for the {R}ule of {C}onditioning.
\newblock {\em The Annals of Statistics}, 10:1075--1089, 1982.

\bibitem{shafer1985}
Glenn Shafer.
\newblock Conditional probability.
\newblock {\em International Statistical Review}, 53:261--277, 1985.

\bibitem{Shafer:1997uk}
Glenn Shafer.
\newblock {Advances in the understanding and use of conditional independence}.
\newblock {\em Annals of Mathematics and Artificial Intelligence}, 21(1):1--11,
  1997.

\bibitem{shafer2003}
Glenn Shafer, Peter~R. Gillett, and Richard~B. Scherl.
\newblock A new understanding of subjective probability and its generalization
  to lower and upper prevision.
\newblock {\em International Journal of Approximate Reasoning}, 33:1--49, 2003.

\bibitem{Skulj:2011db}
Damjan {\v{S}}kulj and Robert Hable.
\newblock {Coefficients of ergodicity for Markov chains with uncertain
  parameters}.
\newblock {\em Metrika}, 76(1):107--133, December 2011.

\bibitem{Troffaes:2014tl}
Matthias C.~M. Troffaes and Gert de~Cooman.
\newblock {\em {Lower previsions}}.
\newblock John Wiley {\&} Sons, March 2014.

\bibitem{Vantaggi:2002va}
Barbara Vantaggi.
\newblock {The L-separation criterion for description of cs-independence
  models}.
\newblock {\em International Journal of Approximate Reasoning}, 29(3):291--316,
  2002.

\bibitem{Walley:1991vk}
Peter Walley.
\newblock {\em Statistical reasoning with imprecise probabilities}.
\newblock Chapman and Hall, London, 1991.

\bibitem{Walley:1996vt}
Peter Walley.
\newblock {Inferences from multinomial data: learning about a bag of marbles}.
\newblock {\em Journal of the Royal Statistical Society, Series B}, 58:3--57,
  1996.

\bibitem{Walley:1996ht}
Peter Walley.
\newblock {Measures of uncertainty in expert systems}.
\newblock {\em Artificial Intelligence}, 83(1):1--58, May 1996.

\bibitem{Walley:2000ef}
Peter Walley.
\newblock {Towards a unified theory of imprecise probability}.
\newblock {\em International Journal of Approximate Reasoning},
  24(2-3):125--148, 2000.

\bibitem{Williams:2007eu}
Peter~M. Williams.
\newblock {Notes on conditional previsions}.
\newblock {\em International Journal of Approximate Reasoning}, 44(3):366--383,
  2007.

\bibitem{Zaffalon:2013fk}
Marco Zaffalon and Enrique Miranda.
\newblock {Probability and time}.
\newblock {\em Artificial Intelligence}, 198:1--51, 2013.

\end{thebibliography}




\section{Pointers to the proofs of the results}\label{sec:app}

The results in this paper have already been proved in Reference~\cite{DeBock2015:thesis}. However, many of these proofs rely heavily on the use of sets of desirable gambles~\cite{Quaeghebeur:2014tjb,Walley:2000ef} or coherent lower previsions~\cite{Miranda:2008fo,Troffaes:2014tl,Walley:1991vk}, two types of uncertainty models that fall beyond the scope of our present contribution. Repeating these proofs here would require a rather massive appendix, filled with technical lemmas that add little intuition. For example, including the lemmas that precede it, our proof for Theorem~\ref{theo:sub-DAG:factorisationandadditivity:LP} spans about twenty pages. For this reason, we prefer to provide pointers to the proofs of our results rather than to repeat them explicitly. The interested reader is cordially invited to follow these pointers, and to acquaint him- or herself with the desirability- and coherence-based development of credal networks under epistemic irrelevance that is presented in Reference~\cite{DeBock2015:thesis}.

The linear programming results in Section~\ref{sec:linearprogramming} were already reported on in Reference~\cite{DeBock:2013wf} without proof. Proofs can be found in Reference~\cite{DeBock2015:thesis}; Proposition~\ref{prop:linearprogram} corresponds to~\cite[Corollary~87]{DeBock2015:thesis}, whereas Theorem~\ref{theo:linearprogramenkelunitary} corresponds to~\cite[Theorem~88]{DeBock2015:thesis}. Under strict positivity conditions, similar properties have also been published in Reference~\cite{Cozman:2000ug}.

Simplified versions of most of the decomposition properties in Section~\ref{sec:theory} have been published with proof in Reference~\cite{DeBock:2014bv}, but these were presented solely in terms of sets of desirable gambles. Extended versions of these desirable gambles results are available in~\cite{DeBock2015:thesis}. They serve as a basis for the lower expectation versions that were presented here, the proofs of which can also be found in~\cite{DeBock2015:thesis}. Theorem~\ref{theo:subnetwork} corresponds to the second case in~\cite[Corollary~60]{DeBock2015:thesis}. 
Theorem~\ref{theo:marginalextension} corresponds to the first case in~\cite[Proposition~71]{DeBock2015:thesis}.
Theorem~\ref{theo:sub-DAG:factorisationandadditivity:LP} corresponds to~\cite[Proposition~56]{DeBock2015:thesis}, whereas Corollaries~\ref{corol:sub-DAG:factorisation:LP} and~\ref{corol:sub-DAG:additivity:LP} correspond to~\cite[Corollary~57]{DeBock2015:thesis} and~\cite[Corollary~59]{DeBock2015:thesis}, respectively.

The proofs for the results in Section~\ref{sec:irrelevance} are also given in Reference~\cite{DeBock2015:thesis}. Proposition~\ref{prop:characteriseAD} corresponds to~\cite[Theorem~64]{DeBock2015:thesis}, whereas Theorem~\ref{theo:ADimpliesHIR} corresponds to the second case in~\cite[Corollary~66]{DeBock2015:thesis}.

Finally, for the results in Section~\ref{sec:conditioningversusupdating}, again, the proofs are given in Reference~\cite{DeBock2015:thesis}. Theorem~\ref{theo:subnetwork:regular} corresponds to~\cite[Theorem~73]{DeBock2015:thesis} and Corollary~\ref{corol:subnetwork:regular} corresponds to~\cite[Corollary~74]{DeBock2015:thesis}.

\end{document}